\documentclass[manuscript, screen]{acmart}
\AtBeginDocument{%
	\providecommand\BibTeX{{%
			\normalfont B\kern-0.5em{\scshape i\kern-0.25em b}\kern-0.8em\TeX}}}
		
\setcopyright{none}

\usepackage[english]{babel}
\usepackage[utf8]{inputenc}
\usepackage{amsmath}
\usepackage{graphicx}
\usepackage{amsfonts}
\usepackage[colorinlistoftodos]{todonotes}

\usepackage{multirow}
\usepackage{hyperref}
\usepackage[flushleft]{threeparttable}
\usepackage{natbib}

\usepackage{multicol}

\usepackage{caption}
\usepackage{subcaption}

\usepackage{booktabs}
\newcommand{\ra}[1]{\renewcommand{\arraystretch}{#1}}

\usepackage{pifont}
\usepackage{longtable}

\usepackage{adjustbox}
\usepackage{array}

\usepackage{soul}
\usepackage{natbib}

\newcolumntype{R}[2]{%
    >{\adjustbox{angle=#1,lap=\width-(#2)}\bgroup}%
    l%
    <{\egroup}%
}
\begin{document}

% Page heads
\markboth{A. Bl\'azquez-Garc\'ia et al.}{A review on outlier/anomaly detection in time series data}

% Title portion
\title{A review on outlier/anomaly detection in time series data}

\author{Ane Bl\'azquez-Garc\'ia}
\email{ablazquez@ikerlan.es}
\author{Angel Conde}
\email{aconde@ikerlan.es}
\affiliation{
  \institution{Ikerlan Technology Research Centre, Basque Research and Technology Alliance (BRTA)}
  \streetaddress{Pº J.M. Arizmendiarrieta, 2}
  \city{Arrasate/Mondrag\'on}
  \postcode{20500}
  \country{Spain}}
\author{Usue Mori}
\email{usue.mori@ehu.es}
\affiliation{
  \institution{Intelligent Systems Group (ISG), Department of Computer Science and Artificial Intelligence, University of the Basque Country (UPV/EHU)}
  \streetaddress{Manuel de Lardizabal, 1}
  \postcode{20018}
  \city{Donostia/San Sebasti\'an}
  \country{Spain}}
\author{Jose A. Lozano}
\email{ja.lozano@ehu.es}
\affiliation{
  \institution{Intelligent Systems Group (ISG), Department of Computer Science and Artificial Intelligence, University of the Basque Country (UPV/EHU)}
  \streetaddress{Manuel de Lardizabal, 1}
  \postcode{20018}
  \city{Donostia/San Sebasti\'an}
  \country{Spain}}
\affiliation{
  \institution{Basque Center for Applied Mathematics (BCAM)}
  \streetaddress{Mazarredo Zumarkalea, 14}
  \postcode{48009}
  \city{Bilbao}
  \country{Spain}}

\begin{abstract}

Recent advances in technology have brought major breakthroughs in data collection, enabling a large amount of data to be gathered over time and thus generating time series. Mining this data has become an important task for researchers and practitioners in the past few years, including the detection of outliers or anomalies that may represent errors or events of interest. This review aims to provide a structured and comprehensive state-of-the-art on outlier detection techniques in the context of time series. To this end, a taxonomy is presented based on the main aspects that characterize an outlier detection technique.

\end{abstract}

\keywords{Outlier detection, anomaly detection, time series, data mining, taxonomy, software}

\maketitle

\section{Introduction}
\label{sec:introduction}

Recent advances in technology allow us to collect a large amount of data over time in diverse research areas. Observations that have been recorded in an orderly fashion and which are correlated in time constitute a time series. Time series data mining aims to extract all meaningful knowledge from this data, and several mining tasks (e.g., classification, clustering, forecasting, and outlier detection) have been considered in the literature \citep{Ratanamahatana2010, Fu2011, Esling2012}.

Outlier detection has become a field of interest for many researchers and practitioners and is now one of the main tasks of time series data mining. Outlier detection has been studied in a variety of application domains such as credit card fraud detection, intrusion detection in cybersecurity, or fault diagnosis in industry. In particular, the analysis of outliers in time series data examines anomalous behaviors across time \citep{Gupta2014}. In the first study on this topic, which was conducted by \citet{Fox1972}, two types of outliers in univariate time series were defined: type I, which affects a single observation; and type II, which affects both a particular observation and the subsequent observations. This work was first extended to four outlier types \citep{Tsay1988}, and then to the case of multivariate time series \citep{Tsay2000}. Since then, many definitions of the term \textit{outlier} and numerous detection methods have been proposed in the literature. However, to this day, there is still no consensus on the terms used \citep{Carreno2019}; for example, outlier observations are often referred to as anomalies, discordant observations, discords, exceptions, aberrations, surprises, peculiarities or contaminants. 

From a classical point of view, a widely used definition for the concept \textit{outlier} has been provided by \citet{Hawkins1980}: 
\begin{quote}
``An observation which deviates so much from other observations as to arouse suspicions that it was generated by a different mechanism."
\end{quote}
Therefore, outliers can be thought of as observations that do not follow the expected behavior. 

\begin{figure}[htb!]
    \centering
    \includegraphics[width=9cm]{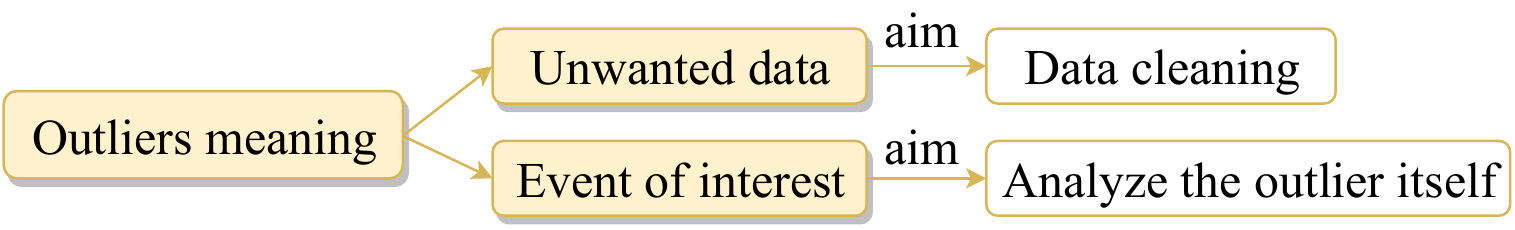}
    \caption{Meaning of the outliers in time series data depending on the aim of the analyst.}
    \label{fig:meaning}
\end{figure}

As shown in Fig. \ref{fig:meaning}, outliers in time series can have two different meanings, and the semantic distinction between them is mainly based on the interest of the analyst or the particular scenario considered. These observations have been related to noise, erroneous, or unwanted data, which by themselves are not interesting to the analyst \citep{Aggarwal2016}. In these cases, outliers should be deleted or corrected to improve the data quality and generate a cleaner dataset that can be used by other data mining algorithms. For example, sensor transmission errors are eliminated to obtain more accurate predictions because the principal aim is to make predictions. Nevertheless, in recent years and, especially in the area of time series data, many researchers have aimed to detect and analyze unusual but interesting phenomena. Fraud detection is an example of this because the main objective is to detect and analyze the outlier itself. These observations are often referred to as anomalies \citep{Aggarwal2016}. 

The purpose of this review is to present a structured and comprehensive state-of-the-art on outlier detection techniques in time series data and attempt to extract the essence of the concept \textit{outlier}, focusing on the detection algorithms given by different authors. Despite the broad terminology that is used to refer to outliers, this review focuses on the identification of outliers in the unsupervised framework, regardless of the term used in the original papers. 

Although a number of surveys on outlier detection methods have been presented in the literature \citep{Hodge2004,Chandola2007,Chandola2009,Aguinis2013,Aggarwal2016,Xu2019}, very few focus on temporal data, including time series \citep{Gupta2013a}. In fact, to the best of our knowledge, there is no review in the literature that deals exclusively with time series data and provides a taxonomy for the classification of outlier detection techniques according to their main characteristics. In addition to grouping the techniques and providing a global understanding of the outliers and their detection in time series, this taxonomy also helps the readers to choose the type of technique that best adapts to the problem that they intend to solve. This review will also report some of the publicly available software. 

The rest of this paper is organized as follows. In Section \ref{sec:taxonomy}, a taxonomy for the classification of outlier detection techniques in time series data is proposed. Section \ref{sec:pointout}, Section \ref{sec:subsequenceout} and Section \ref{sec:ts} present different techniques used for point, subsequence, and time series outlier detection, respectively. The techniques are classified according to the taxonomy proposed in Section \ref{sec:taxonomy}, and the intuition of the concept \textit{outlier} on which the methods are based is provided. In Section \ref{sec:4}, the publicly available software for some of the considered outlier detection methods is presented. Finally, Section \ref{sec:6} contains the concluding remarks and outlines some areas for further research. 
\section{A taxonomy of outlier detection techniques in the time series context}
\label{sec:taxonomy}

\begin{figure}[htb!]
    \centering
    \includegraphics[width=10.7cm]{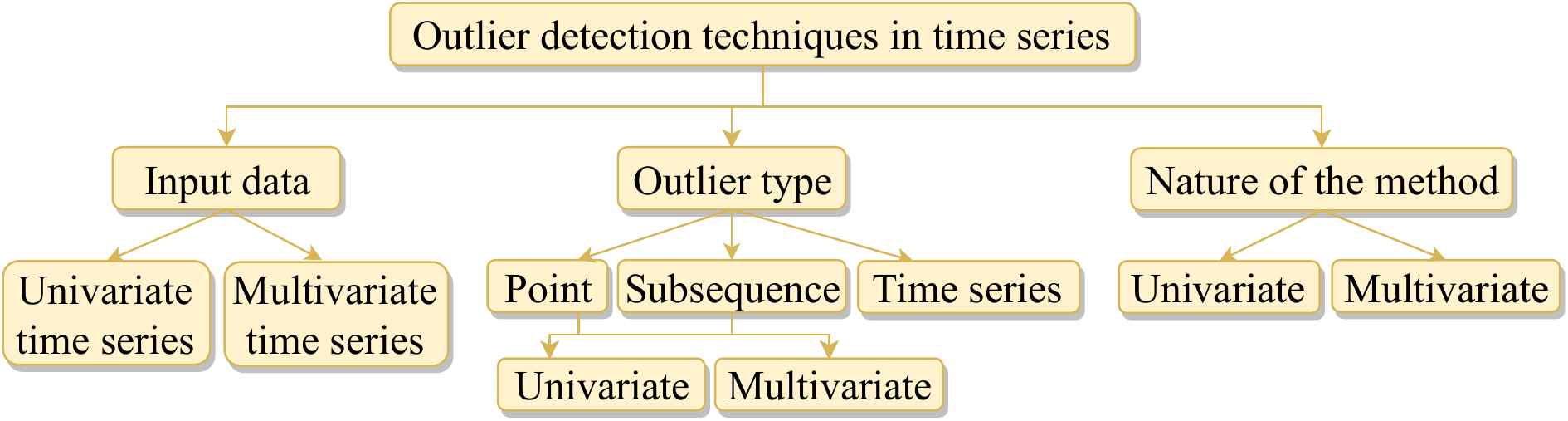}
    \caption{Proposed taxonomy of outlier detection techniques in time series data.}
    \label{fig:esquema}
\end{figure}

Outlier detection techniques in time series data vary depending on the input data type, the outlier type, and the nature of the method. Therefore, a comprehensive taxonomy that encompasses these three aspects is proposed in this section. Fig. \ref{fig:esquema} depicts an overview of the resulting taxonomy, and each axis is described in detail below.

\subsection{Input data}

The first axis represents the type of input data that the detection method is able to deal with (i.e., a \textit{univariate} or a \textit{multivariate} time series).

\begin{definition}{(\textit{Univariate time series})}
    A \textit{univariate time series} $X=\{x_t\}_{t\in T }$ is an ordered set of real-valued observations, where each observation is recorded at a specific time $t\in T\subseteq \mathbb{Z^{+}}$. 
\end{definition}

Then, $x_t$ is the \textit{point} or observation collected at time $t$ and $S = x_p,x_{p+1},...,x_{p+n-1}$ the \textit{subsequence} of length $n\leq |T|$ starting at position $p$ of the time series $X$, for $p,t\in T$ and $p\leq |T|-n+1$. It is assumed that each observation $x_t$ is a realized value of a certain random variable $X_t$. Some examples of univariate time series can be found in Fig. \ref{fig:pointuni} and Fig. \ref{fig:subsequni}.

\begin{definition}{(\textit{Multivariate time series})}
    A \textit{multivariate time series} $\boldsymbol{X}=\{\boldsymbol{x}_t\}_{t\in T }$ is defined as an ordered set of $k$-dimensional vectors, each of which is recorded at a specific time $t\in T\subseteq \mathbb{Z^+}$ and consists of $k$ real-valued observations, $\boldsymbol{x}_t=(x_{1t},...,x_{kt})$ \footnote{Note that this definition could include irregularly sampled time series or dimensions (variables) with different temporal granularities, assuming that some of the $x_{it}$ values might be missing. However, the techniques discussed in this paper only consider regularly sampled time series with the same temporal granularity in all dimensions.}. 
\end{definition}

Then, $\boldsymbol{x}_t$ is said to be a point and $\boldsymbol{S}=\boldsymbol{x}_p,\boldsymbol{x}_{p+1},...,\boldsymbol{x}_{p+n-1}$ a subsequence of length $n\leq |T|$ of the multivariate time series $\boldsymbol{X}$, for $p,t\in T$ and $p\leq |T|-n+1$. For each dimension $j\in\{1,...,k\}$, $X_j=\{x_{jt}\}_{t\in T}$ is a univariate time series and each observation $x_{jt}$ in the vector $\boldsymbol{x}_t$ is a realized value of a random time-dependent variable $X_{jt}$ in $\boldsymbol{X}_t=(X_{1t},...,X_{kt})$. In this case, each variable could depend not only on its past values but also on the other time-dependent variables. Some examples of multivariate time series can be seen in Fig. \ref{fig:pointmulti} and Fig. \ref{fig:subseqmulti}.

\subsection{Outlier type}
The second axis describes the outlier type that the method aims to detect (i.e., a point, a subsequence, or a time series). 

 \begin{itemize}
        \item \textit{Point outliers}. A point outlier is a datum that behaves unusually in a specific time instant when compared either to the other values in the time series (global outlier) or to its neighboring points (local outlier). Point outliers can be univariate or multivariate depending on whether they affect one or more time-dependent variables, respectively. For example, Fig. \ref{fig:pointuni} contains two univariate point outliers, O1 and O2, whereas the multivariate time series composed of three variables in Fig. \ref{fig:pointmulti} has both univariate (O3) and multivariate (O1 and O2) point outliers. 
    
        \begin{figure}[htb!]
             \centering
             \begin{subfigure}[b]{0.45\textwidth}
                 \centering
                \includegraphics[width=\textwidth]{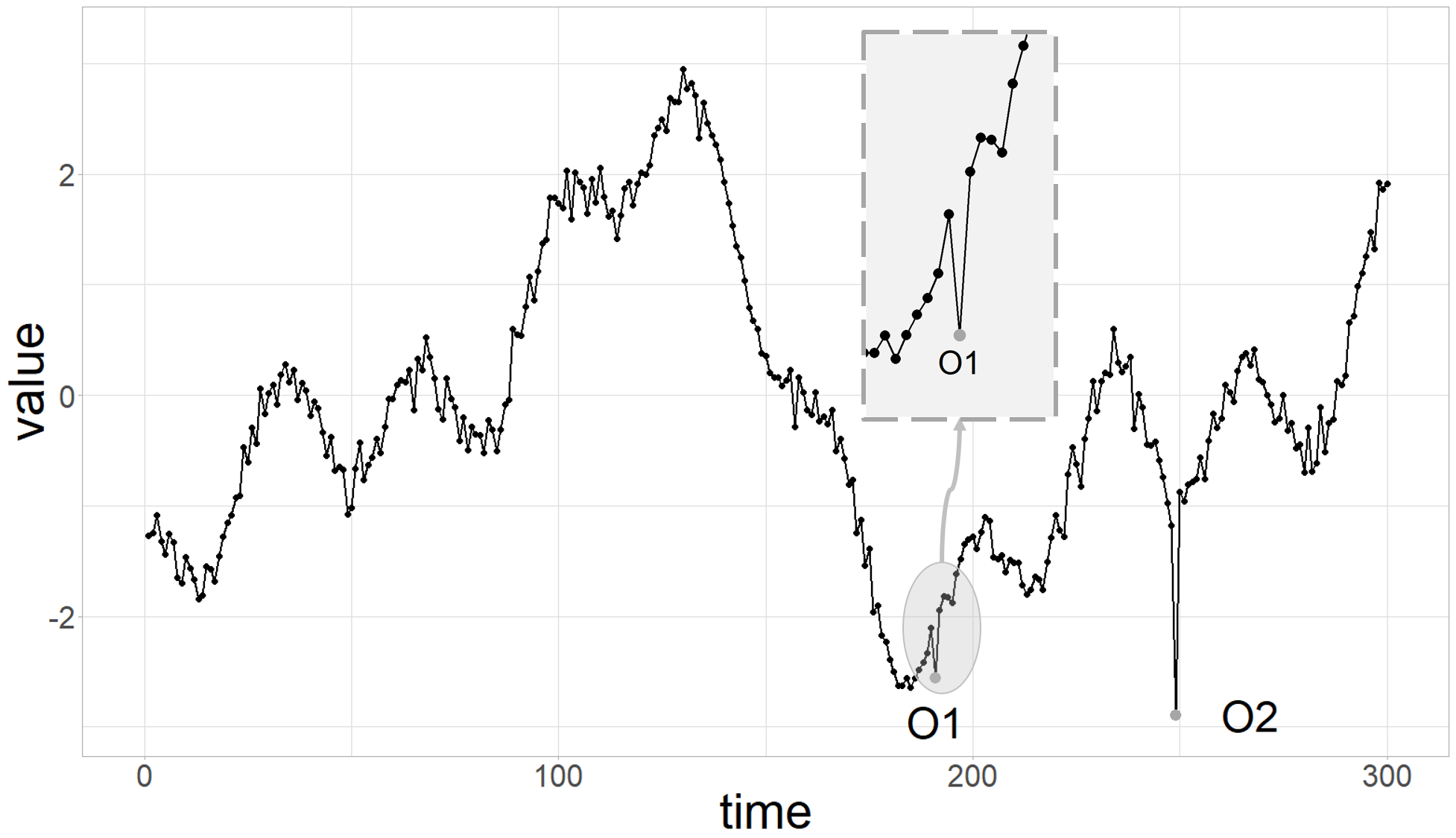}
                 \caption{Univariate time series.}
                 \label{fig:pointuni}
             \end{subfigure}
             \hspace{0.45cm}
             \begin{subfigure}[b]{0.31\textwidth}
                 \centering
                 \includegraphics[width=\textwidth]{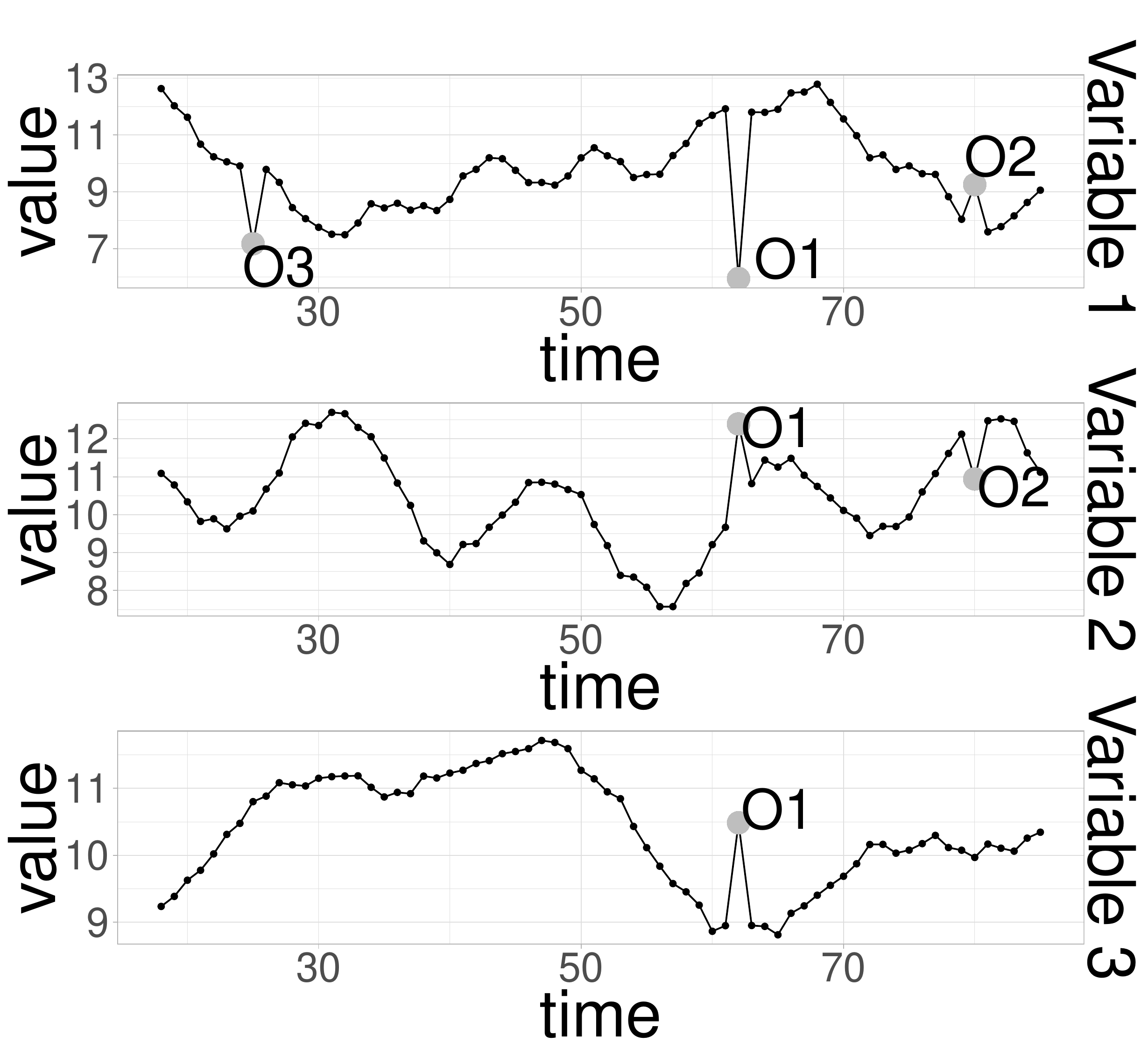}
                 \caption{Multivariate time series.}
                 \label{fig:pointmulti}
             \end{subfigure}
                \caption{Point outliers in time series data.}
                \label{fig:pointoutliers}
        \end{figure}
    
        \item \textit{Subsequence outliers}. This term refers to consecutive points in time whose joint behavior is unusual, although each observation individually is not necessarily a point outlier. Subsequence outliers can also be global or local and can affect one (univariate subsequence outlier) or more (multivariate subsequence outlier) time-dependent variables. Fig. \ref{fig:subseqoutliers} provides an example of univariate (O1 and O2 in Fig. \ref{fig:subsequni}, and O3 in Fig. \ref{fig:subseqmulti}) and multivariate (O1 and O2 in Fig. \ref{fig:subseqmulti}) subsequence outliers. Note that the latter do not necessarily affect all the variables (e.g., O2 in Fig. \ref{fig:subseqmulti}).
        
        \begin{figure}[htb!]
             \centering
             \begin{subfigure}[b]{0.425\textwidth}
                 \centering
                 \includegraphics[width=\textwidth]{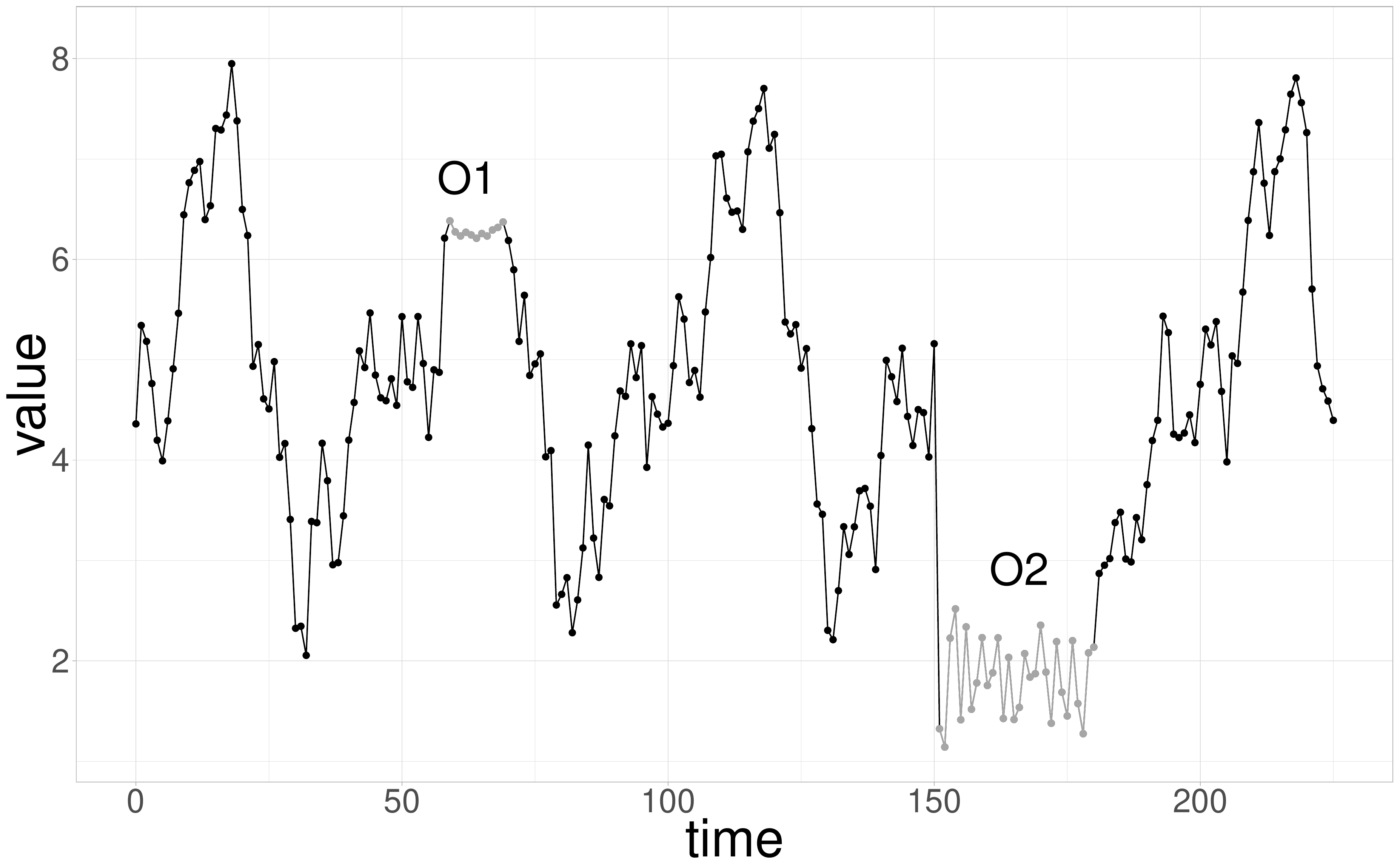}
                 \caption{Univariate time series.}
                 \label{fig:subsequni}
             \end{subfigure}
             \hspace{0.45cm}
             \begin{subfigure}[b]{0.31\textwidth}
                 \centering
                 \includegraphics[width=\textwidth]{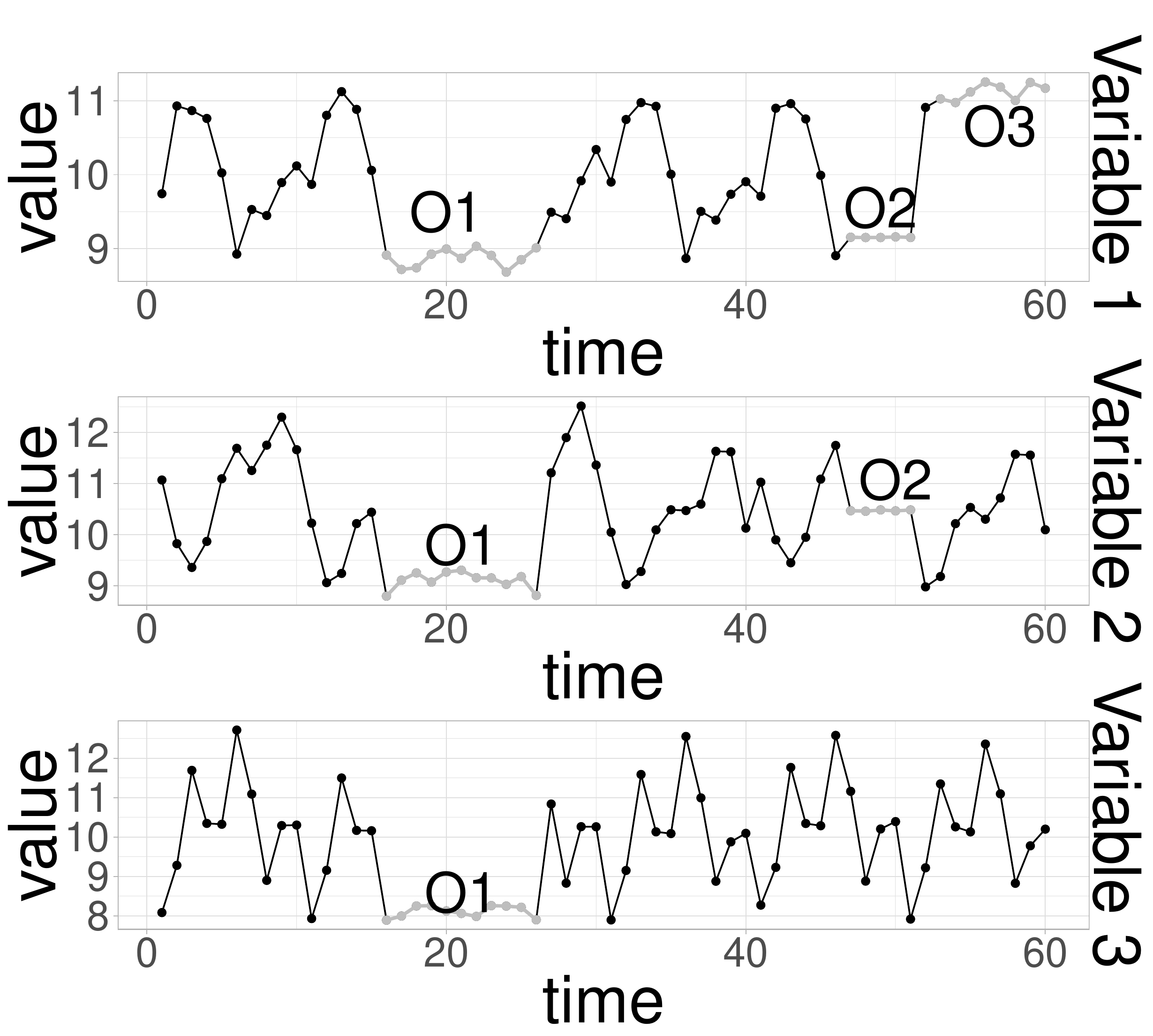}
                 \caption{Multivariate time series.}
                 \label{fig:subseqmulti}
             \end{subfigure}
                \caption{Subsequence outliers in time series data.}
                \label{fig:subseqoutliers}
        \end{figure}

        \item \textit{Outlier time series}. Entire time series can also be outliers, but they can only be detected when the input data is a multivariate time series. Fig. \ref{fig:wholets} depicts an example of an outlier time series that corresponds to \textit{Variable 4} in a multivariate time series composed of four variables. The behavior of \textit{Variable 4} significantly differs from the rest. 
        
       \begin{figure}[htb!]
            \centering
            \includegraphics[width=5.6cm]{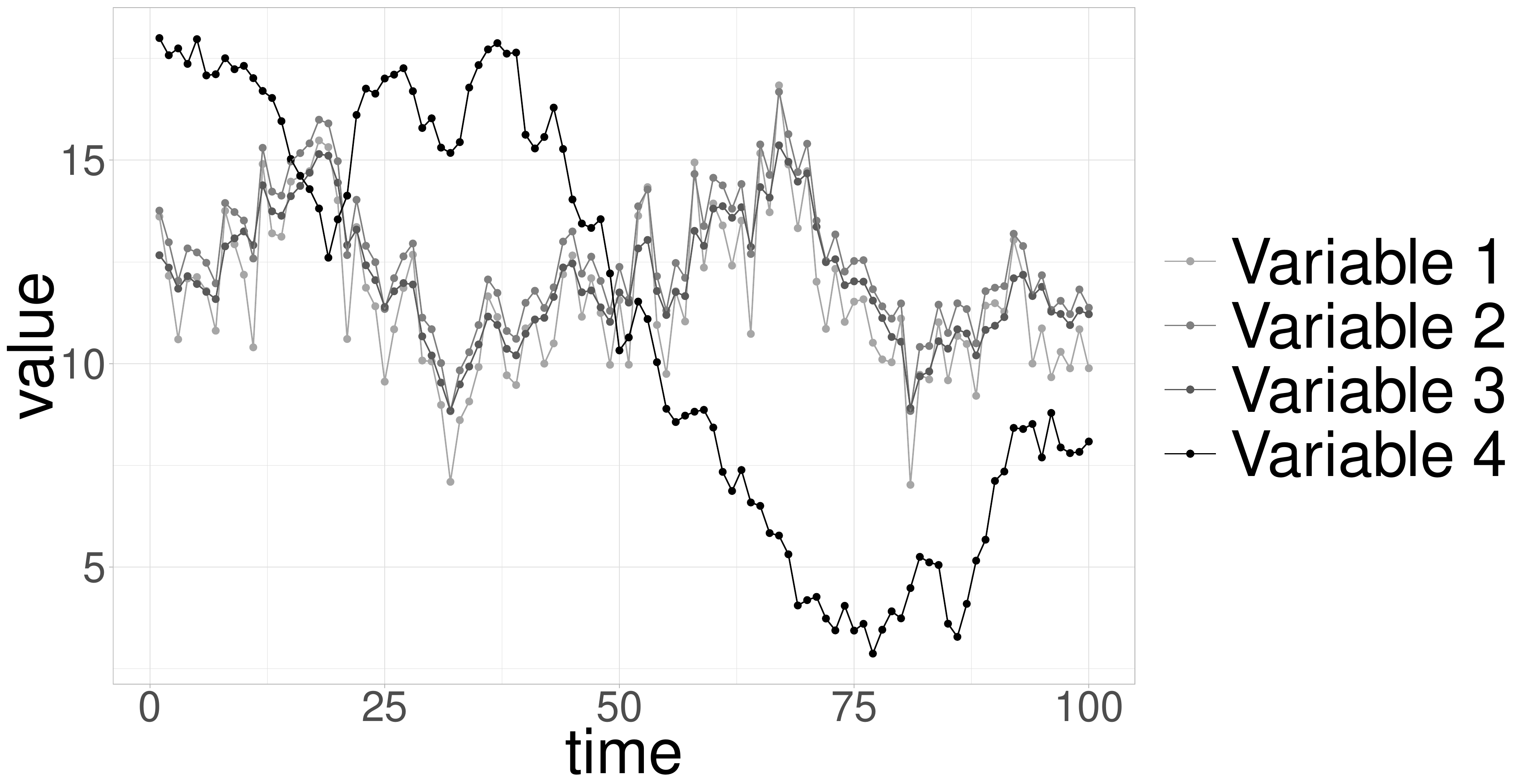}
            \caption{Outlier time series (\textit{Variable 4}) in a multivariate time series.}
            \label{fig:wholets}
        \end{figure}
        
    \end{itemize}
    
    Observe that this axis is closely related to the input data type. If the method only allows univariate time series as input, then no multivariate point or subsequence outliers can be identified. In addition, outlier time series can only be found in multivariate time series. Finally, it should be noted that the outliers depend on the context. Thus, if the detection method uses the entire time series as contextual information, then the detected outliers are global. Otherwise, if the method only uses a segment of the series (a time window), then the detected outliers are local because they are outliers within their neighborhood. Global outliers are also local, but not all local outliers are global. In other words, some local outliers may seem normal if all of the time series is observed but may be anomalous if we focus only on their neighborhood (e.g., O1 in Fig. \ref{fig:pointuni}).

\subsection{Nature of the method} 
The third axis analyzes the nature of the detection method employed (i.e., if the detection method is \textit{univariate} or \textit{multivariate}). A univariate detection method only considers a single time-dependent variable, whereas a multivariate detection method is able to simultaneously work with more than one time-dependent variable. Note that the detection method can be univariate, even if the input data is a multivariate time series, because an individual analysis can be performed on each time-dependent variable without considering the dependencies that may exist between the variables. In contrast, a multivariate technique cannot be used if the input data is a univariate time series. Consequently, this axis will only be mentioned for multivariate time series data.
\section{Point outliers}
\label{sec:pointout}

Point outlier detection is the most common outlier detection task in the area of time series. This section presents the techniques used to detect this type of outlier, in both univariate (Section \ref{sec:univariatets}) and multivariate (Section \ref{sec:multivariatets}) time series data. 

\begin{figure}[htb!]
    \centering
    \includegraphics[width=6.5cm]{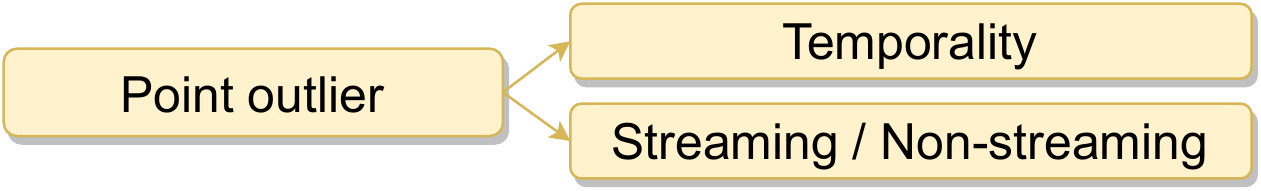}
    \caption{Characteristics related to point outlier detection problems.}
    \label{fig:characpoint}
\end{figure}

Specifically, as shown in Fig. \ref{fig:characpoint}, two key characteristics of these methods will be highlighted throughout their presentation. Concerning the first characteristic, or the treatment of temporality, some methods consider the inherent temporal order of the observations, while others completely ignore this information. The main difference between the methods that include temporal information and those that do not is that the latter methods produce the same results, even if they are applied to a shuffled version of the series. Within the methods that use temporality, a subgroup of methods use time windows. Consequently, the same results are obtained when shuffling the observations within the window, but not when shuffling the whole time series.

In relation to the second characteristic (see Fig. \ref{fig:characpoint}), some techniques are able to detect outliers in streaming time series by determining whether or not a new incoming datum is an outlier as soon as it arrives, without having to wait for more data. Within this group, some methods use a fixed model throughout the stream evolution, whereas others update the models used for detection with the new information received---either by retraining the whole model or by learning in an incremental manner. We consider that a technique does not apply to a streaming time series (i.e., non-streaming) if it is unable to make a decision at the arrival of the new datum.

Most of the analyzed point outlier detection techniques can be applied in a streaming context and they take the temporality of the data into account, either by considering the full time series as an ordered sequence or with the use of time windows. Therefore, we will only make reference to this axis for methods that cannot be applied in a streaming environment or which completely ignore the temporal information in the data. Finally, even though many techniques can theoretically deal with streaming time series, very few are able to adapt incrementally to the evolution of the stream. Consequently, we will also highlight these techniques.

\subsection{Univariate time series} \label{sec:univariatets}

\begin{figure}[htb!]
    \centering
    \includegraphics[width=9cm]{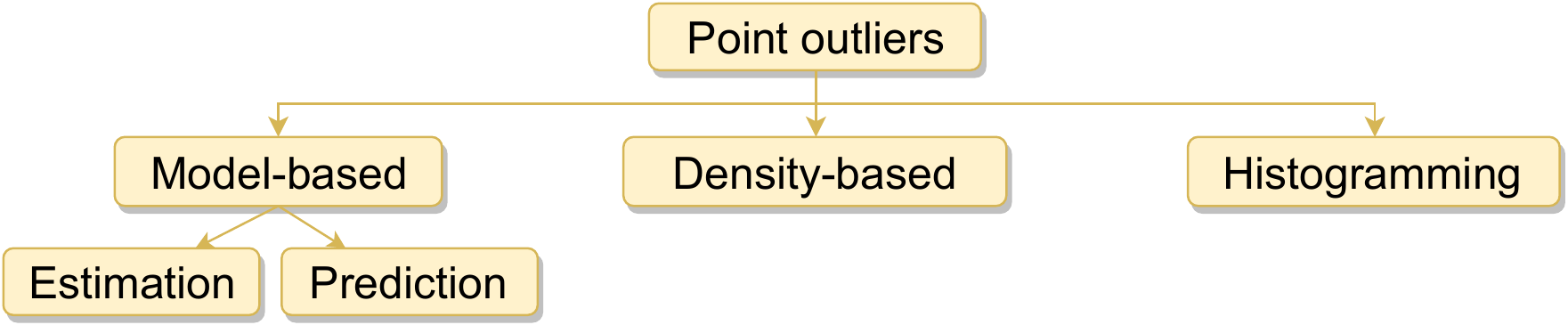}
    \caption{Types of methods for detecting point outliers in univariate time series.}
    \label{fig:esqpoint}
\end{figure}

The techniques that will be discussed in this section intend to detect point outliers in a univariate time series and are organized based on the diagram shown in Fig. \ref{fig:esqpoint}. Since a single time-dependent variable is considered, recall that these outliers are univariate points and that only univariate detection techniques can be used for their detection.

\begin{figure}[htb!]
    \centering
    \resizebox{.6\textwidth}{!}{
        \includegraphics[width=6.8cm]{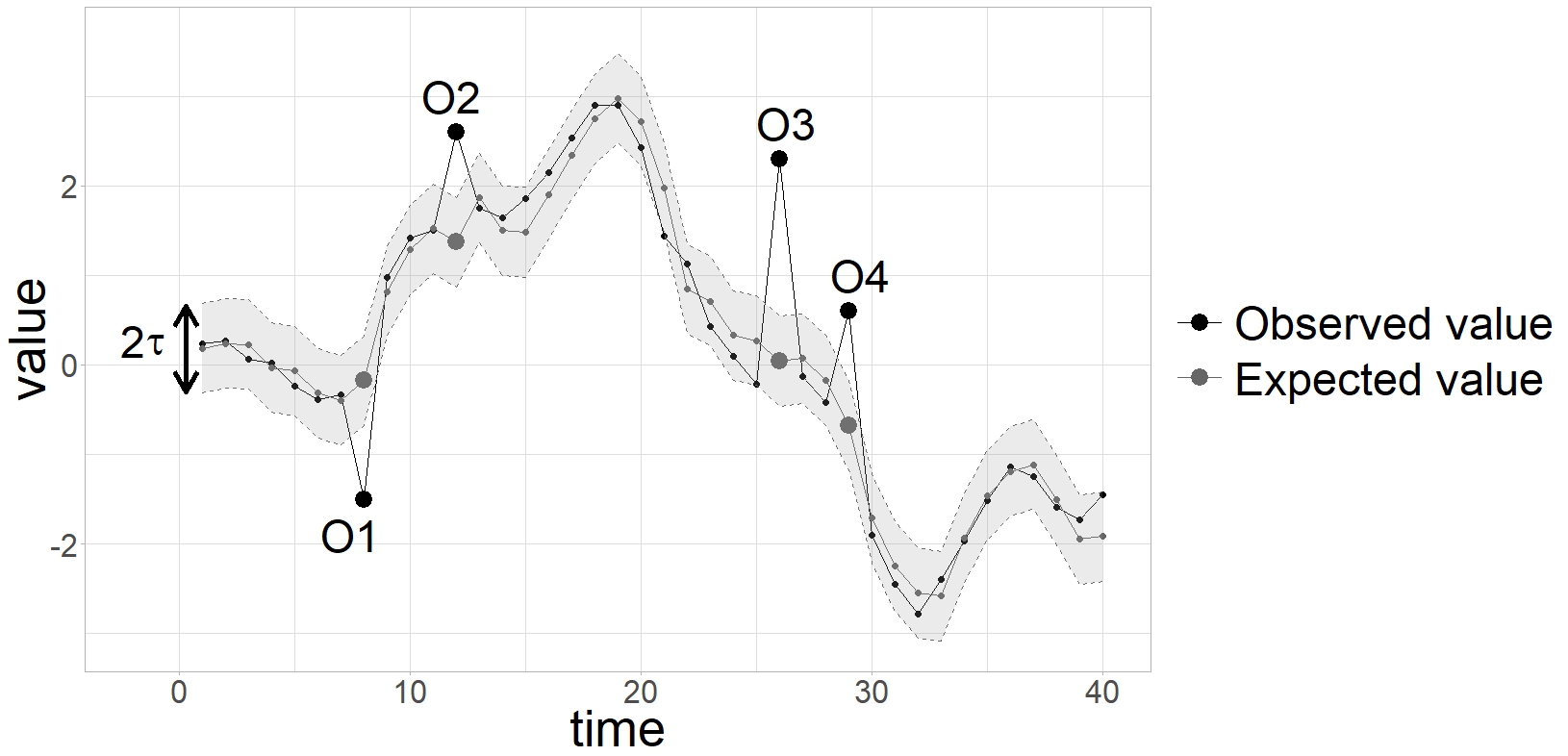}
    }
    \caption{Point outlier detection in univariate time series based on the comparison of expected and observed values.}
    \label{fig:prediuni}
\end{figure}

The most popular and intuitive definition for the concept of \textit{point outlier} is a point that significantly deviates from its expected value. Therefore, given a univariate time series, a point at time $t$ can be declared an outlier if the distance to its expected value is higher than a predefined threshold $\tau$:
\begin{equation}
    |x_t-\hat{x}_t|>\tau
    \label{eq:pred}
\end{equation}
where $x_t$ is the observed data point, and $\hat{x}_t$ is its expected value. This problem is graphically depicted in Fig. \ref{fig:prediuni}, where the observed values within the shadowed area are at most at distance $\tau$ from their expected values. Clearly, O3 is the point that differs the most from its expected value, although O1, O2, and O4 are also farther than distance $\tau$ (outside the band), so all four are declared outliers. 

The outlier detection methods based on the strategy described in equation~\eqref{eq:pred} are denominated \textit{model-based} methods in this paper and are the most common approaches in the literature. Even though each technique computes the expected value $\hat{x}_t$ and the threshold $\tau$ differently, they are all based on fitting a model (either explicitly or implicitly). As shown in Table \ref{tab:estiprediuni}, if $\hat{x}_t$ is obtained using previous and subsequent observations to $x_t$ (past, current, and future data), then the technique is within the \textit{estimation model-based} methods. In contrast, if $\hat{x}_t$ is obtained relying only on previous observations to $x_t$ (past data), then the technique is within the \textit{prediction model-based} methods. In practice, the main difference between using estimation or prediction methods is that techniques within this latter category can be used in streaming time series because they can determine whether or not a new datum is an outlier as soon as it arrives. In the case of estimation methods, this can only be done if, in addition to some points preceding $x_t$, only the current point $x_t$ is used to compute the estimated value ($k_2=0$).

\begin{table}[htb!]
    \caption{Data used in model-based techniques in univariate time series, for $k\geq 1$, and $k_1,k_2\geq 0$ such that $k_1+k_2>0$.} 
    \label{tab:estiprediuni} 
    \resizebox{.7\textwidth}{!}{
        \centering
        \ra{1.3}
        \begin{tabular}{@{}cccccc@{}} \toprule
             & Data used & $\longrightarrow$ & Expected value & $\longrightarrow$ & Point outliers\\ \midrule
            Estimation models & $\{x_{t-k_1},...,x_{t},...,x_{t+k_2} \}$& $\longrightarrow$  & $\hat{x}_{t}$ & \multirow{2}{*}{$\longrightarrow$} &  \multirow{2}{*}{$|x_t-\hat{x}_t|>\tau$}  \\ 
            Prediction models & $\{x_{t-k},...,x_{t-1} \}$ & $\longrightarrow$ & $\hat{x}_{t}$   \\ 
            \bottomrule
        \end{tabular}
        }
\end{table}

The most simple \textit{estimation models} are based on constant or piecewise constant models, where basic statistics such as the median \citep{Basu2007} or the Median Absolute Deviation (MAD) \citep{Mehrang2015} are used to obtain the expected value $\hat{x}_t$. These statistics are calculated using the full series or by grouping the data in equal-length segments and cannot be applied in streaming when future data is needed ($k_2>0$). A more sophisticated approach is to utilize unequal-length segments obtained with a segmentation technique. \citet{Dani2015} use the mean of each segment to determine the expected value of the points within that segment. They also use an adaptive threshold $\tau_i=\alpha \sigma_i$, where $\alpha$ is a fixed value and $\sigma_i$ the standard deviation of segment $i$. 

Other estimation-based techniques intend to identify data points that are unlikely if a certain fitted model or distribution is assumed to have generated the data. For instance, some authors model the structure of the data using smoothing methods such as B-splines or kernels \citep{Chen2010} or variants of the Exponentially Weighted Moving Average (EWMA) method \citep{Carter2012}, \citet{Song2015} and \citet{Zhang2016} use slope constraints, \citet{Mehrang2015} assume that the data are approximately normal if the outliers are not taken into account, and \citet{Reddy2017} use Gaussian Mixture Models (GMM). 

Once a model or distribution is assumed and fitted, \citet{Chen2010, Carter2012} and \citet{Reddy2017} use equation~\eqref{eq:pred} directly to decide whether a point is an outlier or not. Similarly, \citet{Song2015} establish a maximum and a minimum possible slope between consecutive values, whereas \citet{Zhang2016} model the change in slopes before and after time $t$, assuming that the slopes should not change significantly at a time point. Finally, in \citet{Mehrang2015}, the Extreme Studentized Deviate (ESD) test is employed to make the decision: the null hypothesis considered is that there are no outliers, whereas the alternative is that there are up to $k$. Regardless of the temporal correlation, the algorithm computes $k$ test statistics iteratively to detect $k$ point outliers. At each iteration, it removes the most outlying observation (i.e., the furthest from the mean value).

Some other univariate outlier detection methods have analyzed all of the residuals obtained from different models to identify the outliers. For example, \citet{Hochenbaum2017} use the STL decomposition, and \citet{Akouemo2014, Akouemo2016, Akouemo2017} use ARIMA models with exogenous inputs, linear regression or Artificial Neural Networks (ANNs). Although most of these models can also be used in prediction, in this case, the outliers are detected in the residual set using past and future data. Specifically, once the selected model is learned, hypothesis testing is applied over the residuals to detect the outliers. In \citet{Akouemo2014, Akouemo2016, Akouemo2017}, assuming that the underlying distribution of the residuals is known, the minimum and maximum values are examined simultaneously at each iteration of the algorithm. The hypothesis to be tested is whether an extremum is an outlier (alternative hypothesis) or not (null hypothesis). The detected outliers are corrected, and the process is repeated until no more outliers are detected. In contrast, in \citet{Hochenbaum2017}, the ESD test explained in the previous paragraph is applied but the median and MAD are used for robustness instead of the mean and standard deviation.

In contrast to estimation models, techniques based on \textit{prediction models} fit a model to the time series and obtain $\hat{x}_t$ using only past data; that is, without using the current point $x_t$ or any posterior observations. Points that are very different from their predicted values are identified as outliers. Recall that all of the techniques within this category can deal with streaming time series. 

Within the prediction-based methods, some use a fixed model and thus are not able to adapt to the changes that occur in the data over time. For example, the DeepAnT algorithm that was presented by \citet{Munir2018} is a deep learning-based outlier detection approach that applies a fixed Convolutional Neural Networks (CNNs) to predict values in the future. Other methods use an autoregressive model \citep{Hill2010} or an ARIMA model \citep{Zhang2012}, which obtain confidence intervals for the predictions instead of only point estimates. Consequently, these methods implicitly define the value of $\tau$.

Other techniques adapt to the evolution of the time series by retraining the model. As the most basic approach, \citet{Basu2007} describe a method that predicts the value $\hat{x}_t$ with the median of its past data. More elaborately, \citet{Zhou2018a} fit an ARIMA model within a sliding window to compute the prediction interval, so the parameters are refitted each time that the window moves a step forward.

Extreme value theory is another prediction-based approach that has been employed to detect point outliers in streaming univariate series, using past data and retraining. Given a fixed risk $q$, this theory allows us to obtain a threshold value $z_{q,t}$ that adapts itself to the evolution of the data such that $P(X_t>z_{q,t})<q$, for any $t\geq0$, assuming that the extreme values follow a Generalized Pareto Distribution (GPD). Incoming data is used to both detect anomalies ($X_t>z_{q,t}$) and refine $z_{q,t}$. Even if this notation is valid for upper thresholding, the idea is symmetrical when getting lower-bound thresholds or both upper- and lower-bound thresholds. \citet{Siffer2017} propose two algorithms based on this theory: SPOT, for data following any stationary distribution; and DSPOT, for data that can be subject to concept drift \citep{Tsymbal2004,Gama2014}.

Some of these prediction-based methods retrain the underlying model periodically, or each time a new point arrives. Therefore, they can adapt to the evolution of the data. However, none of them applies incremental model learning approaches, where the model is not rebuilt from scratch each time but is updated incrementally using only the new information received. This avoids the cost associated with training the models more than once and permits a more gradual adaptation to changes that can occur in the data, which is of special interest in streaming contexts \citep{Losing2018}. In this sense, and in contrast to the previous approaches, \citet{Xu2016,Xu2017} suggest modeling a univariate time series stream in an incremental fashion. This method uses Student-t processes to compute the prediction interval and updates the covariance matrix with the newly arrived data point. \citet{Ahmad2017} use the Hierarchical Temporal Memory (HTM) network, which is also a prediction model-based technique that updates incrementally as new observations arrive.

All of these techniques are based on equation \eqref{eq:pred}. However, not all the existing point outlier detection methods rely on that idea, such as the \textit{density-based} methods, which belong to the second category of methods depicted in Fig. \ref{fig:esqpoint}. Techniques within this group consider that points with less than $\tau$ neighbors are outliers; that is, when less than $\tau$ objects lie within distance $R$ from those points. This could be denoted as
\begin{equation}
    x_t \text{ is an outlier} \Longleftrightarrow |\{x \in X |  d(x,x_t)\leq R\}|< \tau
    \label{eq:density-based}
\end{equation} where $d$ is most commonly the Euclidean distance, $x_t$ is the data point at time stamp $t$ to be analyzed, $X$ is the set of data points, and $R\in \mathbb{R}^{+}$. Thus, a point is an outlier if $\tau_p+\tau_s<\tau$, where $\tau_p$ and $\tau_s$ are the number of preceding and succeeding neighbors (points that appear before and after $x_t$) at distance lower or equal than $R$, respectively. 

The detection of density-based outliers has been widely handled in non-temporal data, but the concept of neighborhood is more complex in time series because the data are ordered. To take temporality into account, \citet{Fassetti, Angiulli2010} and \citet{Ishimtsev2017a} apply this method within a sliding window, which allows us to wheter or not a new value of a streaming time series is an outlier upon arrival.

\begin{figure}[htb!]
     \centering
     \begin{subfigure}[b]{0.37\textwidth}
         \centering
         \includegraphics[width=\textwidth]{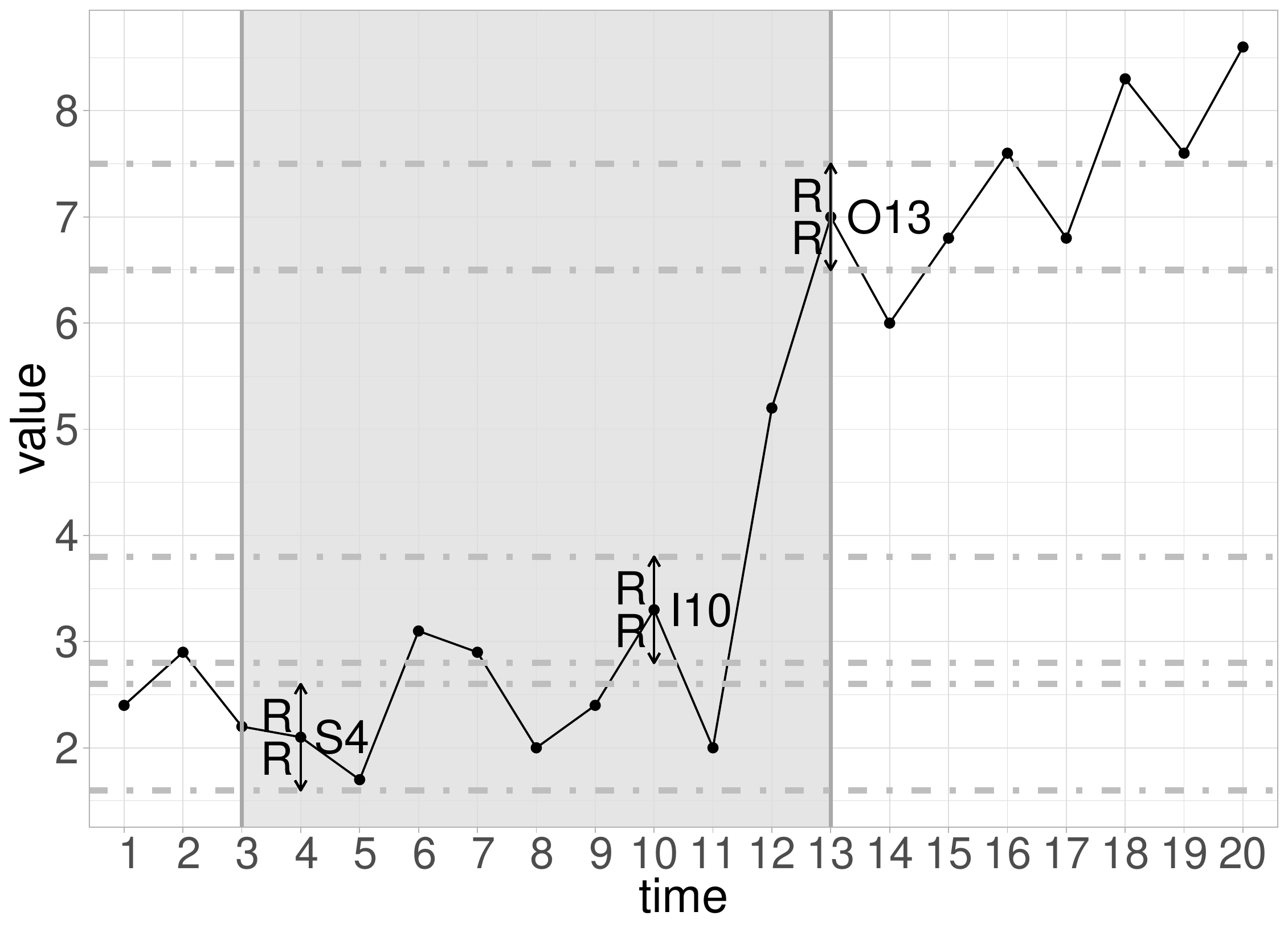}
         \caption{$t=13$}
         \label{fig:distance1}
     \end{subfigure}
     \hspace*{0.7cm}
     \begin{subfigure}[b]{0.37\textwidth}
         \centering
         \includegraphics[width=\textwidth]{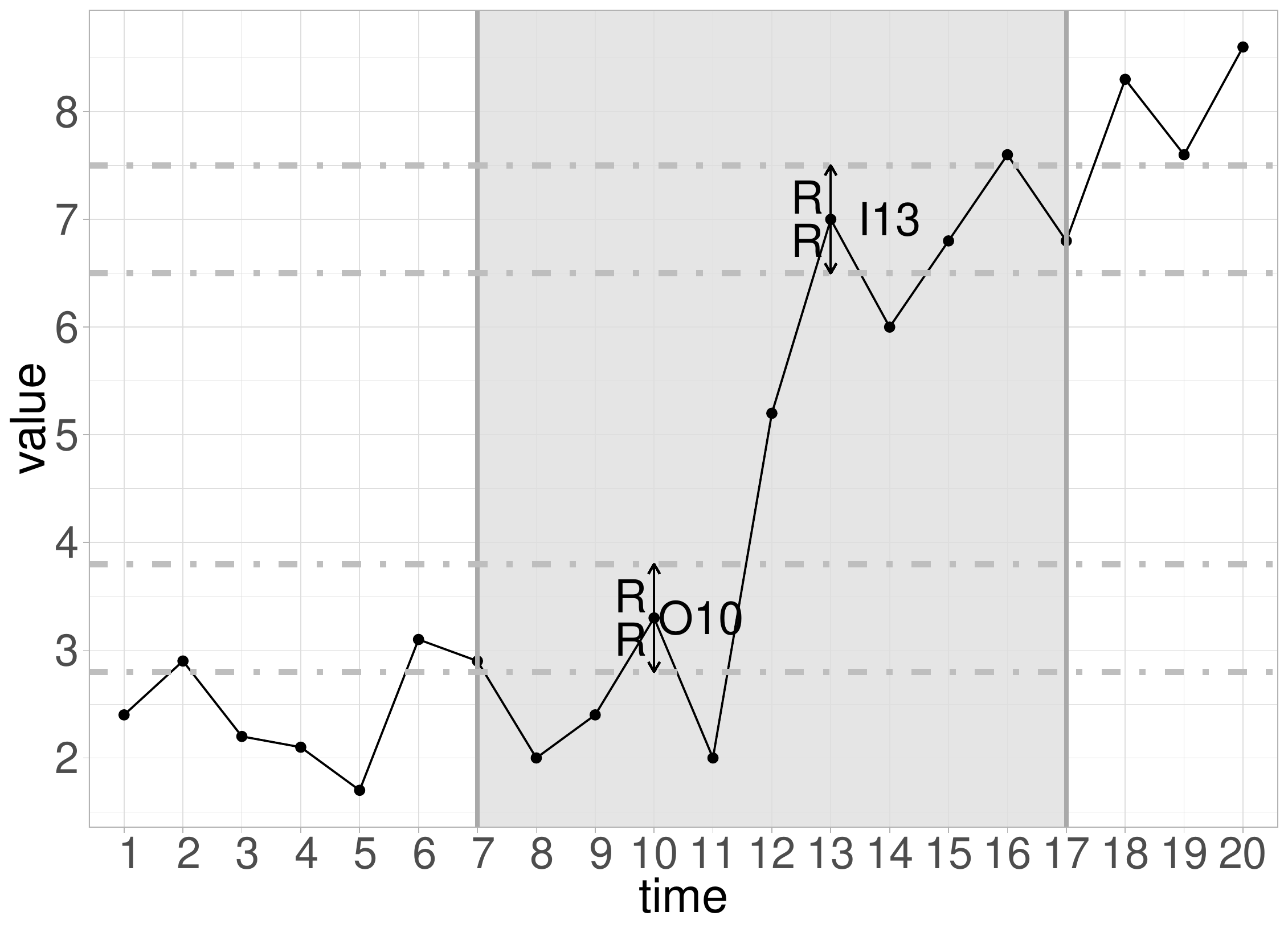}
         \caption{$t=17$}
         \label{fig:distance2}
     \end{subfigure}
        \caption{Density-based outliers within a sliding window of length 11 at time step $t$.}
        \label{fig:density}
\end{figure}

An illustration of this density-based outlier concept is provided in Fig. \ref{fig:density} at two different time steps with $R=0.5$, $\tau=3$, and a sliding window of length 11. When using sliding windows, a point can be an outlier for a window (e.g., O13 at $t=13$) but not for another (e.g., I13 at $t=17$). However, if a data point has at least $\tau$ succeeding neighbors within a window, then it cannot be an outlier for any future evolution (e.g., S4 at $t=13$). 

\begin{figure}[htb!]
     \centering
     \begin{subfigure}[b]{0.38\textwidth}
         \centering
         \includegraphics[width=\textwidth]{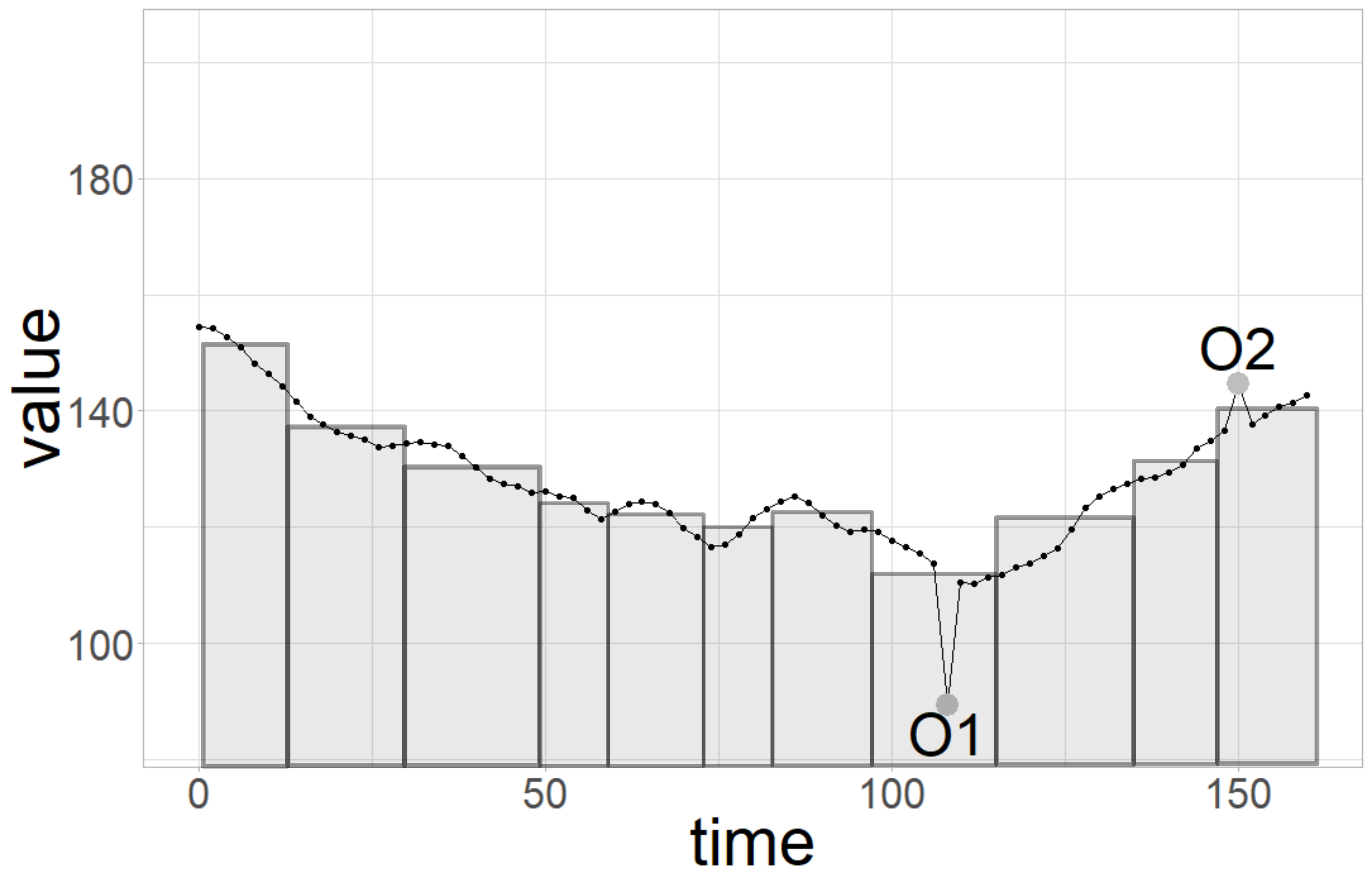}
         \caption{Optimal histogram with eleven buckets.}
         \label{fig:histog}
     \end{subfigure}
     \hspace*{0.3cm}
     \begin{subfigure}[b]{0.38\textwidth}
         \centering
         \includegraphics[width=\textwidth]{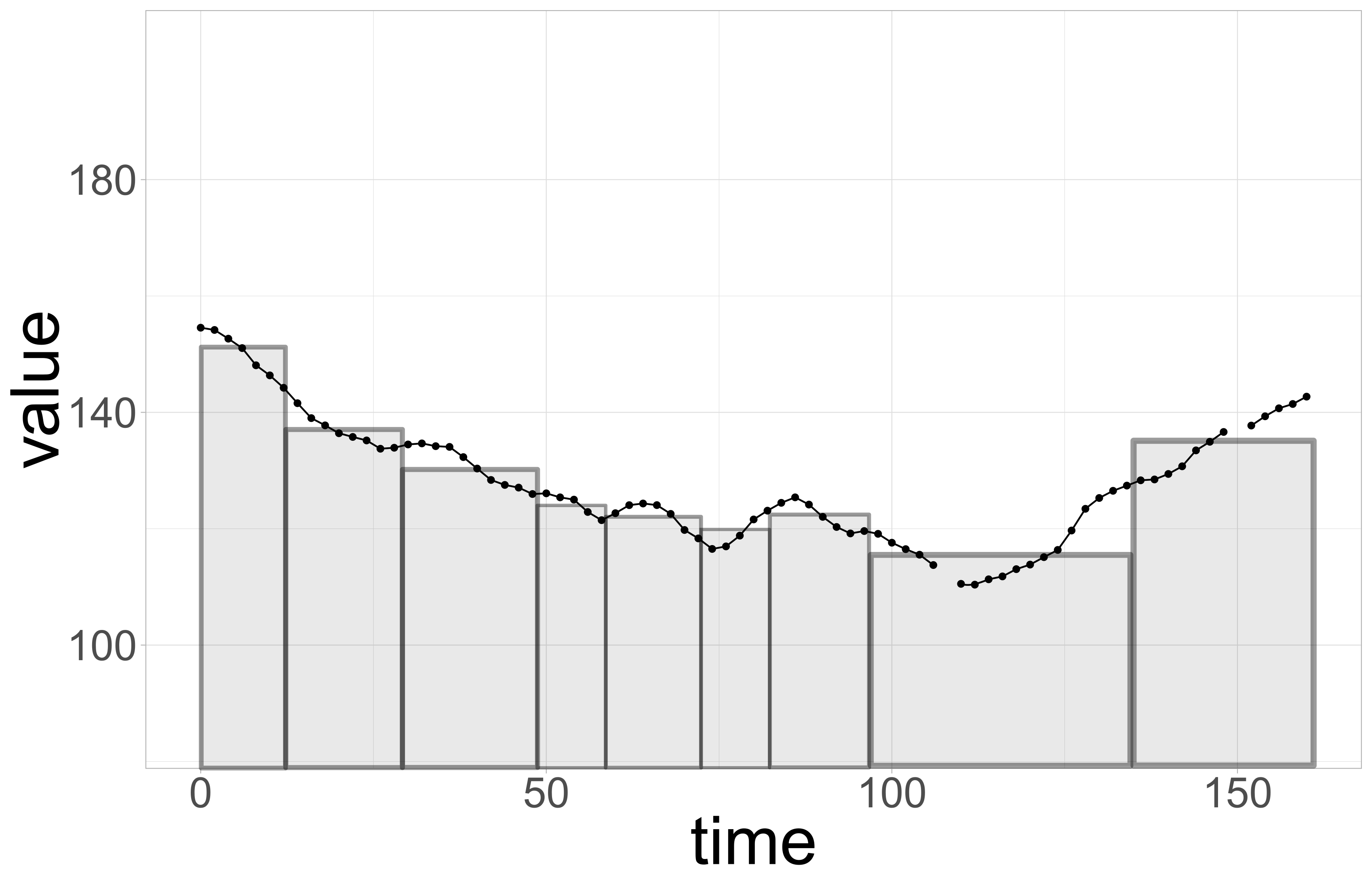}
         \caption{Optimal histogram with nine buckets and O1 and O2 removed.}
         \label{fig:histogsin}
     \end{subfigure}
        \caption{Example of a deviant set $D=\{O1, O2\}$ in a univariate time series.}
        \label{fig:histogramming}
\end{figure}

\textit{Histogramming} is the last group analyzed in this section. This type of method is based on detecting the points whose removal from the univariate time series results in a histogram representation with lower error than the original, even after the number of buckets has been reduced to account for the separate storage of these points (see Fig. \ref{fig:histogramming}). The histogram is built by computing the average of the values within each bucket, in which the order of the observations is preserved. Then, given a sequence of points $X$ and a number of buckets $B$, $D\subset X$ is a deviant set if
\begin{equation}
    E_X(H_B^*)>E_{X-D}(H_{B-|D|}^*)
    \label{eq:histogram}
\end{equation}
where $H_B^*$ is the optimal histogram (histogram with lowest approximation error) on $X$ with $B$ buckets, $E_X()$ is the total error in the approximation and $H_{B-|D|}^*$ is the optimal histogram on $X-D$ with $B-|D|$ buckets. \citet{Dalgleish1999} introduced the term \textit{deviant} to refer to these point outliers. They proposed a dynamic programming mechanism to produce a histogram consisting of $B-|D|$ buckets and $|D|$ deviants, minimizing its total error. Some years later \citet{Muthukrishnan2004} observed that for any bucket, the optimal set of $k=|D|$ deviants always consists of the $l$ highest and remaining $k-l$ lowest values within the bucket, for some $l\leq k$. Moreover, they presented not only an optimal algorithm for non-streaming time series but also a closely approximate algorithm and a heuristic approach for the streaming case. 

To conclude our review of point outlier detection in univariate time series, a summary of the analyzed techniques and their characteristics is provided in Table \ref{tab:univ}. This table includes the inherent characteristics of the point outliers in time series data mentioned in Fig. \ref{fig:characpoint}, whether the algorithm is iterative or not, the meaning the point outlier has for the authors, and the term they use to name it. We consider that a method is iterative if it is repeatedly applied on the same set to find the outliers. In this case, the few iterative methods are related to detecting unwanted data and improving the quality of the time series.

\begin{table}[htb!]
    \caption{Summary of the techniques used in point outlier detection in univariate time series.}
    \label{tab:univ}
    \centering
    \ra{1.3}
    \resizebox{.7\textwidth}{!}{
    \begin{threeparttable}
    \begin{tabular}{llcccccp{0.5cm}}
        \toprule
        \multirow{2}{*}{Paper} & \multirow{2}{*}{Technique}& \multirow{2}{*}{Iterative}& \multirow{2}{*}{Temporality} & \multirow{2}{*}{Streaming} & \multicolumn{2}{c}{Meaning}  & \multirow{2}{*}{Term}\\
        \cmidrule{6-7}
        & & & & &I & U &  \\ \midrule
        \citet{Dalgleish1999} & Histogram &\ding{55} & \checkmark& \checkmark& \checkmark &  & D \\ \hline
        \citet{Muthukrishnan2004} & Histogram &\ding{55} & \checkmark&\checkmark\tnote{*}& \checkmark &  & D \\ \hline
         \multirow{2}{*}{\citet{Basu2007}} & Estimation & \ding{55} & W & \ding{55}&  & \checkmark & O \\ \cmidrule{2-8}
        & Prediction  & \ding{55} & W &\checkmark&& \checkmark & O \\ \hline
        \citet{Fassetti, Angiulli2010} & Density&\ding{55}& W&\checkmark\tnote{*} &\checkmark & & O \\ \hline
        \citet{Chen2010}& Estimation  &\ding{55}& \checkmark&\checkmark&&\checkmark&O/C \\\hline
        \citet{Hill2010} & Prediction  &\ding{55}&\checkmark &\checkmark&  & \checkmark & A \\ \hline
        \citet{Zhang2012} & Prediction  &\ding{55}&\checkmark&\checkmark & \checkmark & \checkmark & O \\ \hline
        \citet{Carter2012} & Estimation  &\ding{55}&\checkmark&\checkmark\tnote{*} & \checkmark & & A \\ \hline
        \citet{Akouemo2014} & Estimation &\checkmark&\checkmark &\checkmark& & \checkmark& O \\ \hline
        \multirow{2}{*}{\citet{Mehrang2015}} & Estimation\tnote{1} & \ding{55} & W &\ding{55}& & \checkmark & O \\  %, model-based
        \cmidrule{2-8}
        & Estimation \tnote{2} &\checkmark &\ding{55}&\checkmark&&\checkmark& O\\\hline
        \citet{Dani2015} & Estimation &\ding{55}&\checkmark &\checkmark& \checkmark &  & O/A \\ \hline
        \citet{Song2015} & Estimation &\ding{55}&\checkmark &\checkmark&  & \checkmark & Dirty \\ \hline
        \citet{Zhang2016} & Estimation &\ding{55}&\checkmark &\checkmark&  & \checkmark & Dirty \\ \hline
        \citet{Akouemo2016} & Estimation &\checkmark&\checkmark &\checkmark& & \checkmark & O/A \\ \hline
        \citet{Xu2016,Xu2017} & Prediction  &\ding{55}& \checkmark &\checkmark\tnote{*}& \checkmark &  & O/A \\ \hline
        \citet{Ishimtsev2017a} & Density  &\ding{55}& \checkmark &\checkmark& \checkmark &  & A \\ \hline
        \citet{Akouemo2017} & Estimation &\checkmark&\checkmark & \checkmark&& \checkmark & O/A \\ \hline
        \citet{Reddy2017} & Estimation &\ding{55}&\checkmark&\checkmark &\checkmark &  & O \\ \hline
        \citet{Siffer2017} & Prediction &\ding{55}&\ding{55} &\checkmark& \checkmark &  & O/A \\ \hline
        \citet{Ahmad2017} & Prediction &\ding{55}&\checkmark &\checkmark\tnote{*}& \checkmark &  & A \\ \hline
        \citet{Hochenbaum2017} & Estimation  &\checkmark& \ding{55} &\checkmark& \ding{55} &  & O/A \\ \hline
        \citet{Zhou2018a} & Prediction &\ding{55}&\checkmark&\checkmark & & \checkmark & O \\ \hline
        \citet{Munir2018}& Prediction  &\ding{55}&\checkmark&\checkmark & \checkmark & & O/A  \\ 
        \bottomrule
    \end{tabular}
    \begin{tablenotes}
        \item I: Event of interest; U: Unwanted data // W: Window // D: Deviant; O: Outlier; A: Anomaly; C: Corrupted.
        \item[1] MAD. 
        \item[2] Hypothesis testing.
        \item[*] Incremental updating.
     \end{tablenotes}
  \end{threeparttable}
  }
\end{table}

Caution must be taken with the estimation methods that can theoretically be applied in a streaming time series; that is, those that do not use subsequent observations to the last arrived data point $x_t$ ($k_2=0$ in Table \ref{tab:estiprediuni}). Although, it may in theory be possible to apply these techniques in streaming contexts, these methods use the last observation received ($x_t$) and some other past information to calculate its expected value ($\hat{x}_t$) and they then decide whether or not it is an outlier. Consequently, they must perform some calculations after the new point has arrived. The cost of this computation depends on the complexity of each method (which has not been analyzed in this paper or in every original works) but may not always guarantee a fast enough response. Thus, in practice, some of these techniques may not be applicable in streaming contexts and prediction methods are more recommendable for these situations.

%%%%%%%%%%%%%%%%%%%%%%%%%%%%%%%%%%%%%%%%%%%%%%%%
%%%%%%%%%%%%%%%%%%%%%%%%%%%%%%%%%%%%%%%%%%%%%%%%
% MULTIVARIATE TIME SERIES
%%%%%%%%%%%%%%%%%%%%%%%%%%%%%%%%%%%%%%%%%%%%%%%%
%%%%%%%%%%%%%%%%%%%%%%%%%%%%%%%%%%%%%%%%%%%%%%%%

\subsection{Multivariate time series} \label{sec:multivariatets}

The input time series is sometimes a multivariate time series with possibly correlated variables rather than a univariate time series. As opposed to the univariate time series case, the detection method used to identify point outliers in multivariate time series can deal not only with a single variable (Section \ref{sec:univ_tech_multi}) but also with more than one variable simultaneously (Section \ref{sec:multitech}). Additionally, a point outlier in a multivariate time series can affect one (univariate point) or more than one (multivariate point, a vector at time $t$) variables (see Fig. \ref{fig:pointmulti}). As will be seen in the following sections, some multivariate techniques are able to detect univariate point outliers and (similarly) some univariate techniques can be used to detect multivariate point outliers. The characteristics mentioned in Fig. \ref{fig:characpoint} will also be highlighted.

\subsubsection{Univariate techniques} \label{sec:univ_tech_multi}

Given that a multivariate time series is composed of more than one time-dependent variable, a univariate analysis can be performed for each variable to detect univariate point outliers, without considering dependencies that may exist between the variables. Although the literature barely provides examples of this type of approach, in essence, all of the univariate techniques discussed in Section \ref{sec:univariatets} could be applied to each time-dependent variable of the input multivariate time series. As one of the few examples, \citet{Hundman2018} propose using the Long Short-Term Memory (LSTM) deep learning technique (which is a prediction model-based method) to predict spacecraft telemetry and find point outliers within each variable in a multivariate time series, following the idea of equation \eqref{eq:pred}. Additionally, the authors present a dynamic thresholding approach in which some smoothed residuals of the model obtained from past data are used to determine the threshold at each time step. 

\begin{figure}[htb!]
    \centering
    \includegraphics[width=12cm]{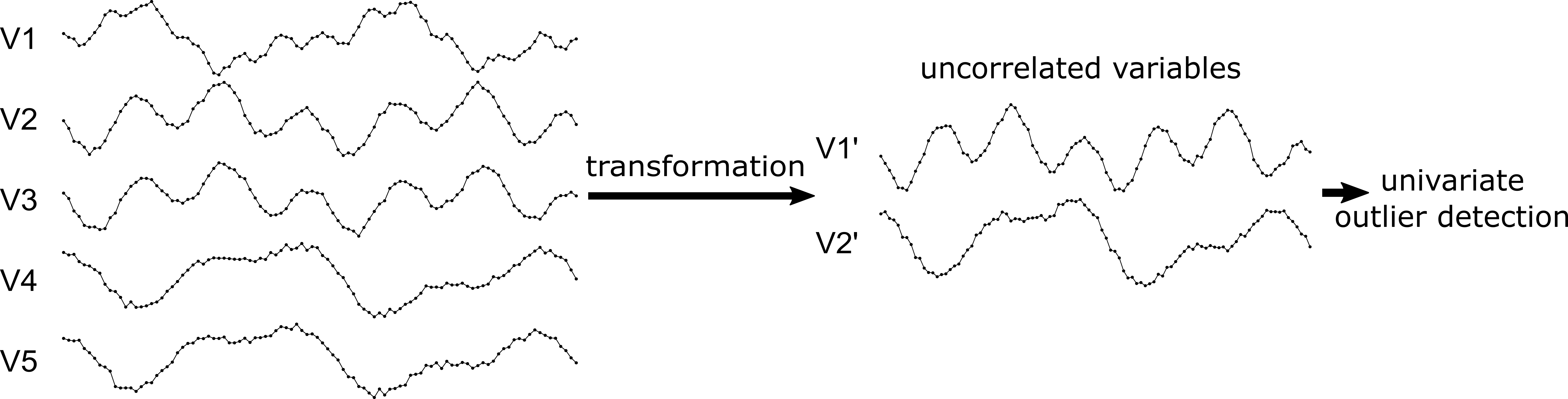}
    \caption{Simplification of point outlier detection in multivariate time series.}
    \label{fig:unimulti}
\end{figure}

Correlation dependencies between the variables are not considered when applying univariate techniques to each time-dependent variable, leading to a loss of information. To overcome this problem, and at the same time to leverage that univariate detection techniques are highly developed, some researchers apply a preprocessing method to the multivariate time series to find a new set of uncorrelated variables where univariate techniques can be applied. These methods are based on \textit{dimensionality reduction} techniques, and as depicted in Fig. \ref{fig:unimulti}, the multivariate series is simplified into a representation of lower dimension before applying univariate detection techniques. Since the new series are combinations of the initial input variables, the identified outliers are multivariate; that is, they affect more than one variable.

Some of those dimensionality reduction techniques are based on finding the new set of uncorrelated variables by calculating linear combinations of the initial variables. For example, \citet{Papadimitriou2005} propose an incremental Principal Component Analysis (PCA) algorithm to determine the new independent variables. In this case, the posterior univariate point outlier detection technique that is applied is based on the autoregressive prediction model (AR). Alternatively, \citet{Galeano2006} suggest reducing the dimensionality with projection pursuit, which aims to find the best projections to identify outliers. The authors mathematically prove that the optimal directions are those that maximize or minimize the kurtosis coefficient of the projected time series. Univariate statistical tests \citep{Fox1972, Chen1993} are then applied iteratively to each projected univariate time series for multivariate point outlier detection. Similarly, \citet{Baragona2007} propose using Independent Component Analysis (ICA) to obtain a set of unobservable independent nonGaussian variables, assuming that the observed data has been generated by a linear combination of those variables. Outliers are identified in each new series independently if equation \eqref{eq:pred} is satisfied for $\hat{x}_{it}=\mu_i$ and $\tau_i=4.47\sigma_i$, where $\mu_i$ is the mean and $\sigma_i$ the standard error of the $i$th new variable.

Other techniques reduce the input multivariate time series into a single time-dependent variable rather than into a set of uncorrelated variables. \citet{Lu2018} define the transformed univariate series using the cross-correlation function between adjacent vectors in time; that is, $\boldsymbol{x_{t-1}}$ and $\boldsymbol{x_{t}}$. Point outliers are iteratively identified in this new series as those that have a low correlation with their adjacent multivariate points. The threshold value $\tau$ is determined at each iteration by the multilevel Otsu's method. \citet{Shahriar2016} also transformed the multivariate time series into a univariate series using a transformation specifically designed for the application domain considered. Point outliers are identified using equation \eqref{eq:pred} and the 3-sigma rule.

The main characteristics of the univariate techniques analyzed for point outlier detection in multivariate time series are described in Table \ref{tab:multipoint_uni} in chronological order. Most of them are non-iterative. In addition, point outliers are events of interest for almost all the researchers. Particularly, note that the transformation methods proposed by \citet{Galeano2006} and \citet{Lu2018} are specific for outlier detection and not general dimensionality reduction techniques. 

\begin{table}[htb!]
    \caption{Summary of the univariate techniques used in multivariate time series for point outlier detection. }
    \label{tab:multipoint_uni}
    \centering
    \ra{1.3}
    \resizebox{.7\textwidth}{!}{
    \begin{threeparttable}
    \begin{tabular}{@{}llccccl@{}}\toprule
   \multirow{2}{*}{Paper} & \multirow{2}{2.5cm}{Technique}& \multirow{2}{*}{Iterative}& \multirow{2}{*}{Temporality} &                              \multicolumn{2}{c}{Meaning}  & \multirow{2}{*}{Term}\\
    \cmidrule{5-6} 
    & & & & I & U &  \\ \midrule
        \citet{Papadimitriou2005} & Dim. reduction & \ding{55} & \checkmark &          \checkmark &  & O \\ \hline
        \citet{Galeano2006} & Dim. reduction &\checkmark& \checkmark &   & \checkmark & O \\ \hline
        \citet{Baragona2007} & Dim. reduction &\ding{55} & \ding{55} &          \checkmark &  & O \\ \hline
        \citet{Shahriar2016} & Dim. reduction &\ding{55}& \checkmark&  \checkmark&  & O/E \\\hline
        \citet{Lu2018} & Dim. reduction &\checkmark& \checkmark&\checkmark&  & O \\ \hline
        \citet{Hundman2018} & Prediction &\ding{55}& \checkmark & \checkmark&  & A \\ 
    \bottomrule
    \end{tabular}
    \begin{tablenotes}
            \item I: Event of interest; U: Unwanted data // O: Outlier; A: Anomaly; E: Event.
         \end{tablenotes}
  \end{threeparttable}
  }
\end{table}

As far as the characteristics mentioned in Fig. \ref{fig:characpoint} are concerned, in dimensionality reduction techniques, the temporality depends on both the transformation and the outlier detection method applied. If at least one of these considers temporality, then so does the complete method because the same results are not obtained when the observations of the input multivariate time series are shuffled. In the case of PCA, projection pursuit, and ICA, these transformation methods do not consider temporality, so the temporality depends on the univariate detection method. Conversely, \citet{Shahriar2016} and \citet{Lu2018} use methods that include temporality in the transformation phase. In the other approach where the dimensionality is not reduced, the temporality directly depends on the applied detection method.

Finally, all of the methods that we have reviewed in this section can theoretically detect point outliers in a streaming context using sliding windows because none needs future data to provide an output. In this context, the prediction-based approach proposed by \citet{Hundman2018} would work the best because, in contrast to techniques based on dimensionality reduction, the newly arrived point to be analyzed is not used on the model construction. However, no incremental versions have been proposed.

\subsubsection{Multivariate techniques} \label{sec:multitech}  

\begin{figure}[htb!]
    \centering
    \includegraphics[width=8.5cm]{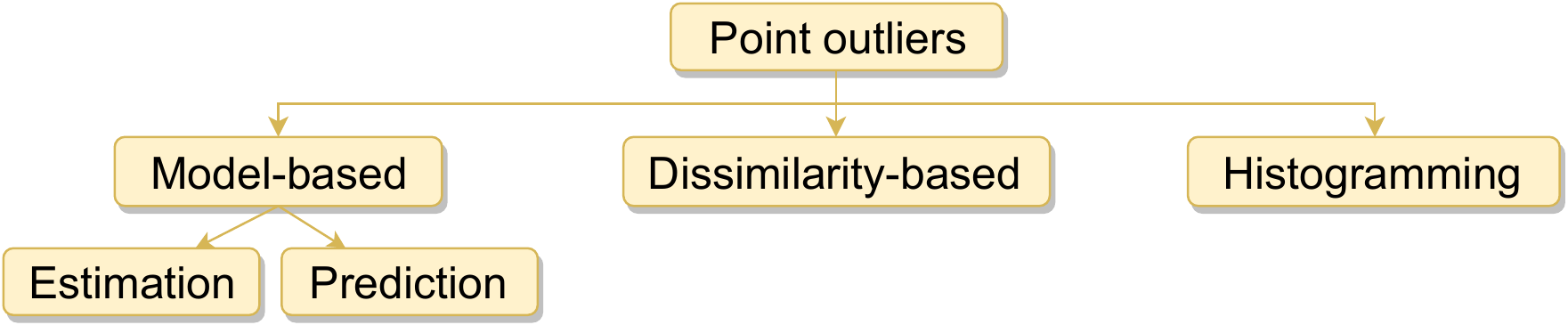}
    \caption{Types of methods for detecting point outliers in multivariate time series.}
    \label{fig:esqpointMULTI}
\end{figure}

In contrast to the univariate techniques discussed previously, this section analyzes the multivariate methods that deal simultaneously with multiple time-dependent variables, without applying previous transformations. These approaches perform the detection directly using all the original input data variables and can be divided into three groups, as described in Fig. \ref{fig:esqpointMULTI}. 

\begin{figure}[htb!]
             \centering
             \begin{subfigure}[b]{0.39\textwidth}
                 \centering
                \includegraphics[width=\textwidth]{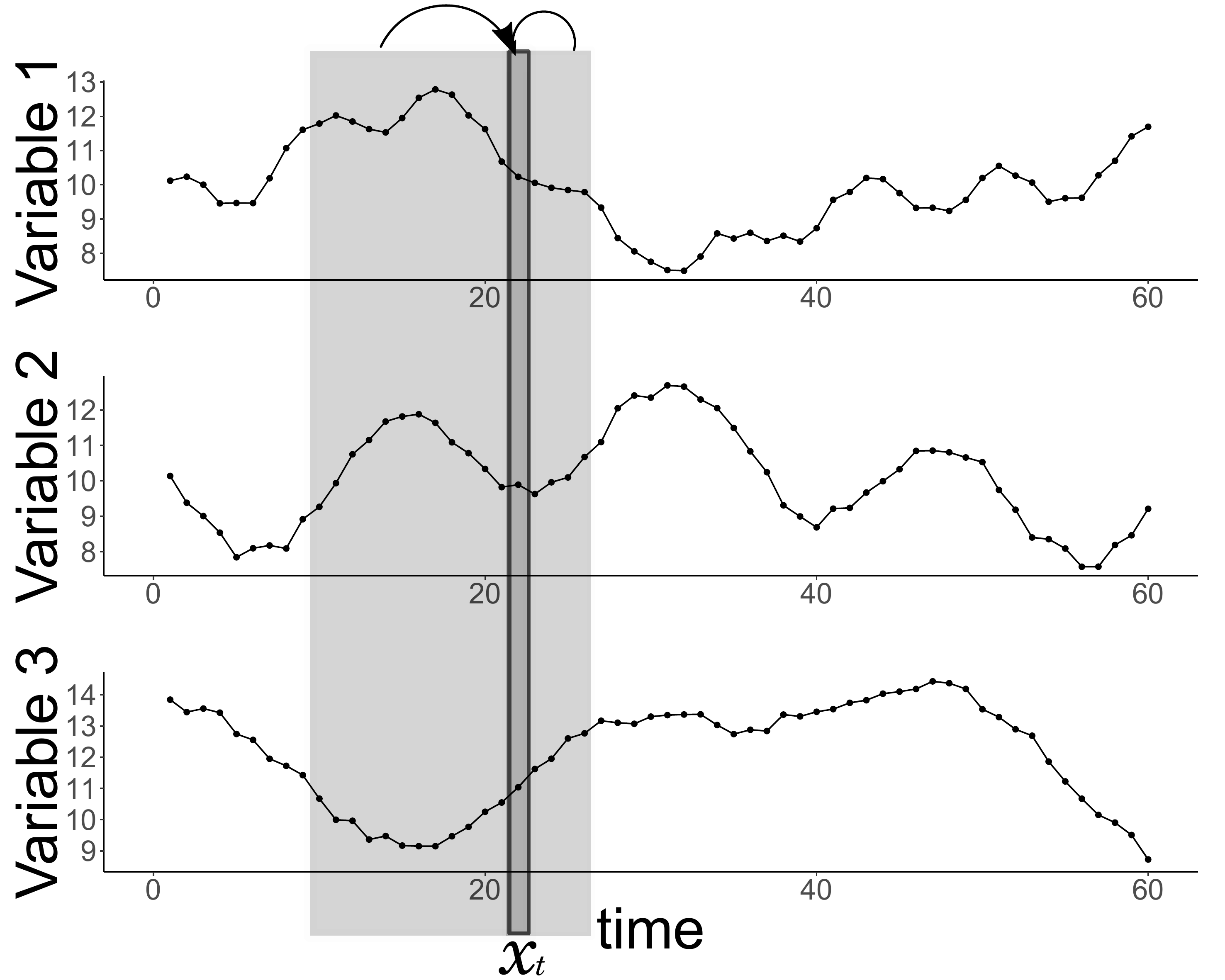}
                 \caption{Estimation models.}
             \end{subfigure}
             \hspace*{0.5cm}
             \begin{subfigure}[b]{0.39\textwidth}
                 \centering
                \includegraphics[width=\textwidth]{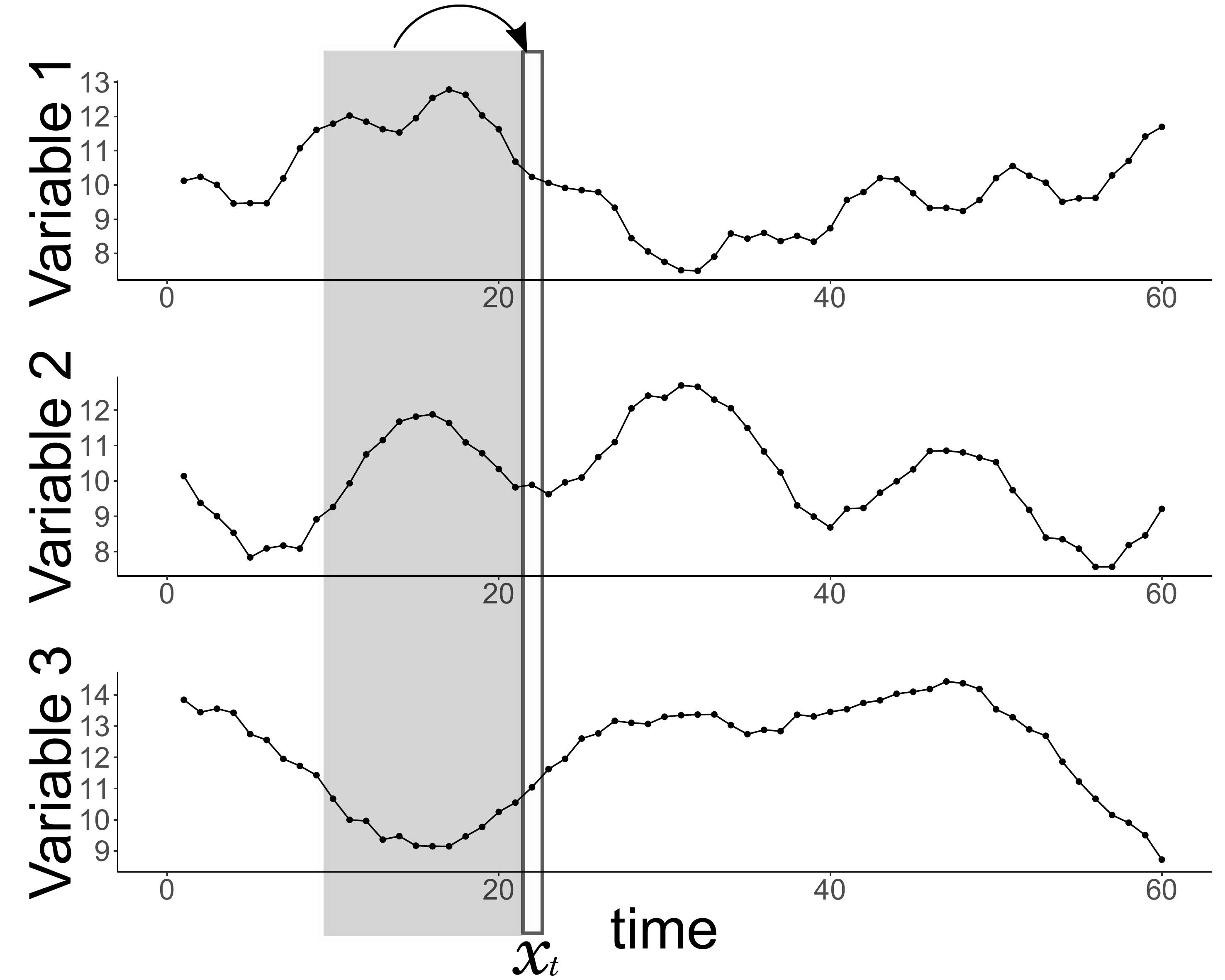}
                 \caption{Prediction models.}
                 \label{fig:predicmod}
             \end{subfigure}
                \caption{Example of the data used in model-based techniques.}
                \label{fig:estimationmulti}
\end{figure}

As in univariate time series, \textit{model-based} techniques can also be used to detect point outliers in multivariate time series. Methods within this group are based on fitting a model that captures the dynamics of the series to obtain expected values in the original input time series. Then, for a predefined threshold $\tau$, outliers are identified if:
\begin{equation}
    ||\boldsymbol{x}_t-\boldsymbol{\hat{x}}_t||>\tau
    \label{eq:multipred}
\end{equation}
where $\boldsymbol{x}_t$ is the actual $k$-dimensional data point, and $\boldsymbol{\hat{x}}_t$ its expected value. Note that this is a generalization of the definition given for the model-based techniques in univariate time series and that the intuition is repeated (refer to equation \eqref{eq:pred} and Table \ref{tab:estiprediuni}). In fact, $\boldsymbol{\hat{x}_t}$ can be obtained using \textit{estimation models}---which use past, current, and future values---or \textit{prediction models}---which only use past values (see Fig. \ref{fig:estimationmulti}). 

Within the \textit{estimation model-based} category, autoencoders are one of the most commonly used methods. Autoencoders are a type of neural network that learns only the most significant features of a training set used as the reference of normality. Since outliers often correspond to non-representative features, autoencoders fail to reconstruct them, providing large errors in \eqref{eq:multipred}. \citet{Sakurada2014} use this method, where the input of the autoencoder is a single multivariate point of the time series. The temporal correlation between the observations within each variable is not considered in this approximation. Therefore, to account for temporal dependencies, \citet{Kieu2018} propose extracting features within overlapping sliding windows (e.g., statistical features or the difference between the Euclidean norms of two consecutive time windows) before applying the autoencoder. \citet{Su2019} suggest a more complex approach based on a Variational AutoEncoder (VAE) with a Gated Recurrent Unit (GRU). The input of the model is a sequence of observations containing $\boldsymbol{x}_t$ and $l$ preceding observations to it. The output is the reconstructed $\boldsymbol{x}_t$ ($\hat{\boldsymbol{x}}_t$). Additionally, they apply extreme value theory \citep{Siffer2017} in the reference of normality to automatically select the threshold value.

Apart from using autoencoders, $\boldsymbol{\hat{x}_t}$ can also be derived from the general trend estimation of multiple co-evolving time series. \citet{Zhou2018, Zhou2019} use a non-parametric model that considers both the temporal correlation between the values within each variable and inter-series relatedness in a unified manner to estimate that trend. In this case, even though the trend of each variable is estimated using the multivariate time series, univariate point outliers are identified within each variable using equation \eqref{eq:pred} instead of $k$-dimensional data points.

The other model-based techniques that can be used to obtain $\boldsymbol{\hat{x}_t}$ in equation \eqref{eq:multipred} are the \textit{prediction model-based} techniques (see Fig. \ref{fig:predicmod}). Techniques within this category also fit a model to a multivariate time series, but the expected values are the predictions for the future made on the basis of past values. For example, the Contextual Hidden Markov Model (CHMM) incrementally captures both the temporal dependencies and the correlations between the variables in a multivariate time series \citep{Zhou2016}. The temporal dependence is modeled by a basic HMM, and the correlation between the variables is included into the model by adding an extra layer to the HMM network. The DeepAnt algorithm \citep{Munir2018} mentioned in Section \ref{sec:univariatets} is also capable of detecting point outliers in multivariate time series using the CNN prediction model. Once the model is learned, the next timestamp is predicted using a window of previous observations as input. 

All of these estimation and prediction model-based techniques can theoretically be employed in a streaming context using sliding windows because no subsequent points to $\boldsymbol{x_t}$ are needed. As with univariate time series, the estimation-based methods need to consider at least the newly arrived point $\boldsymbol{x_t}$ ($k_2=0$ in Table \ref{tab:estiprediuni}), so prediction-based techniques are more appropriate for detecting outliers in a streaming fashion. Additionally, these model-based techniques all use a fixed model, and they do not adapt to changes over time, except for the proposal of \citet{Zhou2016}, which is incrementally updated. 

The \textit{dissimilarity-based} methods will be discussed next. These techniques are based on computing the pairwise dissimilarity between multivariate points or their representations, without the need for fitting a model. Therefore, for a predefined threshold $\tau$, $\boldsymbol{x_t}$ is a point outlier if:
\begin{equation}
    s(\boldsymbol{x_t},\boldsymbol{\hat{x}_t})>\tau
    \label{eq:connectivity_multi}
\end{equation}
where $\boldsymbol{x_t}$ is the actual $k$-dimensional point, $\boldsymbol{\hat{x}_t}$ its expected value, and $s$ measures the dissimilarity between two multivariate points. These methods do not usually use the raw data directly, but instead use different representation methods. For example, \citet{Cheng2008, Cheng2009} represent the data using a graph where nodes are the multivariate points of the series and the edges the similarity values between them computed with the Radial Basis Function (RBF). The idea is to apply a random walk model in the graph to detect the nodes that are dissimilar to the others (i.e., hardly accessible in the graph). By contrast, \citet{Li2009} propose recording the historical similarity and dissimilarity values between the variables in a vector. The aim is to analyze the dissimilarity of consecutive points over time and detect changes using $||.||_1$. 

The last group depicted in Fig. \ref{fig:esqpointMULTI} refers to the \textit{histogramming} approach, where the term \textit{deviant} has also been used in the context of multivariate series (see the definition given in \eqref{eq:histogram}). \citet{Muthukrishnan2004} extend the technique for deviant detection explained in Section \ref{sec:univariatets} to multivariate time series by treating the measurements collected at the same timestamp as a vector. Similar to the univariate case, the method aims to detect the vectors that should be removed so that the compressed representation (histogram) of the remaining data is improved. The authors propose an algorithm for both streaming and non-streaming series to find approximate optimal deviants. 

The main characteristics of the multivariate techniques analyzed for point outlier detection in multivariate time series are depicted in Table \ref{tab:multipoint} in chronological order. Even if most of them find multivariate point outliers, some use the multivariate information to identify point outliers that only affect a single variable (i.e., univariate point outliers). All of the analyzed techniques are non-iterative, and outliers represent events of interest for the researchers. In addition, most of them obtain different results if they are applied to a shuffled version of the time series. Although all these methods can detect outliers in a streaming context, few are incremental or updated as new data arrives. 

\begin{table}[htb!]
    \caption{Summary of the multivariate techniques used in multivariate time series for point outlier detection.}
    \label{tab:multipoint}
    \centering
    \ra{1.3}
    \resizebox{.7\textwidth}{!}{
    \begin{threeparttable}
    \begin{tabular}{@{}lllccl@{}}\toprule
   Paper & Technique& Point & Temporality& Incremental & Term\\
     \midrule
        \citet{Muthukrishnan2004} & Histogramming & Multivariate & \checkmark & \checkmark & D \\ \hline
        \citet{Cheng2008,Cheng2009} & Estimation & Multivariate & \ding{55} & \ding{55} & A \\ \hline
        \citet{Li2009} & Dissimilarity & Univariate &\checkmark& \checkmark & O \\ \hline
        \citet{Sakurada2014} & Estimation& Multivariate & \ding{55} & \ding{55} & A  \\ \hline
        \citet{Zhou2016} & Prediction & Multivariate & \checkmark&\checkmark & N/A'  \\ \hline
        \citet{Kieu2018} & Estimation & Multivariate & \checkmark & \ding{55} & O  \\ \hline
        \citet{Munir2018} & Prediction & Multivariate & \checkmark& \ding{55} & A/O  \\ \hline
        \citet{Zhou2018,Zhou2019} & Estimation & Univariate & \checkmark& \ding{55} & O/N/E \\ \hline
        \citet{Su2019} & Estimation & Multivariate & \checkmark & \ding{55} & A \\
    \bottomrule
    \end{tabular}
    \begin{tablenotes}
            \item D: Deviant; O: Outlier; A: Anomaly; A': Abnormality; E: Event; N: Novelty.
    \end{tablenotes}
  \end{threeparttable}
  }
\end{table}

\section{Subsequence outliers}
\label{sec:subsequenceout}

\begin{figure}[htb!]
    \centering
    \includegraphics[width=4.9cm]{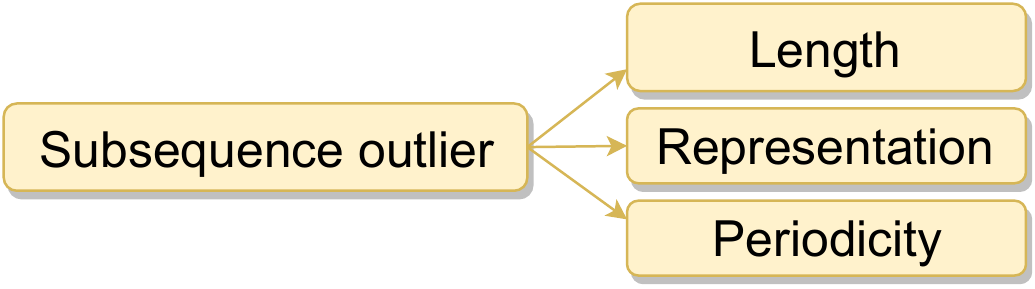}
    \caption{Characteristics related to subsequence outlier detection problems.}
    \label{fig:esqsubs}
\end{figure}

As shown in Fig. \ref{fig:esquema}, subsequence outliers are the second type of outliers that can be detected in time series data. In this case, the aim is to identify a set of consecutive points that jointly behave unusually. To this end, subsequence outlier detection methods need to consider some key aspects, which are shown in Fig. \ref{fig:esqsubs}, and which make the detection of subsequence outliers more challenging than point outlier detection.

To begin with, subsequences are composed of a set of points and are not composed of a single point, so they have a certain length. Typically, methods consider fixed-length subsequences, although some techniques do allow us to detect subsequences of different lengths (variable-length) simultaneously (e.g., \citet{Senin2015, Wang2018}) \footnote{Note that methods that find fixed-length subsequence outliers could also find subsequences of different lengths by applying the method repeatedly for each possible length.}. Methods based on fixed-length subsequences need the user to predefine the length, and they commonly obtain the subsequences using a sliding window over the time series. In contrast, methods that allow finding variable-length subsequences do not prespecify this length. A final aspect to consider regarding the length of the subsequence is that the number of subsequences that the method will consider and analyze depends on the chosen length, thus the number of subsequences will be higher when the length is shorter.

The second characteristic that subsequence outlier detection methods need to consider is the representation of the data. Since the comparison between subsequences is more challenging and costly than the comparison between points, many techniques rely on a representation of the subsequences rather than using the original raw values. Discretization is a widely used representation method and can be obtained using approaches such as equal-frequency binning (e.g., \citet{Keogh2002}), equal-width binning (e.g., \citet{Chen2011, Chen2013, Rasheed2014}), or SAX (e.g., \citet{Senin2015, Wang2018}). These discretization techniques can also be used as a starting point to obtain other representations, such as bitmaps (e.g., \citet{Wei2005, Kumar2005}). A detailed overview of the existing research regarding outlier detection in discrete sequences can be found in \citet{Chandola2012}, which highlights the applicability of those techniques to time series data. Additionally, apart from discretization, raw data has also been used directly to obtain representations based on dictionaries (e.g., \citet{Carrera2016}), exemplars (e.g., \citet{Jones2016}), or connectivity values (e.g., \citet{Ren2017}).

Another issue that has been barely considered in the literature but which makes the detection of subsequence outliers more challenging is that they can be periodic. Periodic subsequence outliers are unusual subsequences that repeat over time (e.g., \citet{Yang2001, Rasheed2014}). Unlike point outliers where periodicity is not relevant, periodic subsequence outliers are important in areas such as fraud detection because it might be interesting to discover certain periodic anomalous transactions over time. 

Finally, as opposed to point outlier detection, where some methods did not take into account the temporal dependencies, subsequences consider the temporality by default. In addition, when analyzing subsequence outliers in a streaming context, three cases can occur: i) a single data point arrives, and an answer (i.e., outlier or non-outlier) must be given for a subsequence containing this new point; ii) a subsequence arrives, and it needs to be marked as an outlier or non-outlier; and iii) a batch of data arrives and subsequence outliers need to be found within it. In either case, the literature provides methods that can give an answer in a streaming fashion using sliding windows. However, most of them keep the model fixed and do not adapt to changes in the streaming sequence. We will focus on these incremental techniques because they are more suitable for processing streaming time series.

\subsection{Univariate time series} \label{sec:unisubsq}

\begin{figure}[htb!]
    \centering
    \includegraphics[width=12cm]{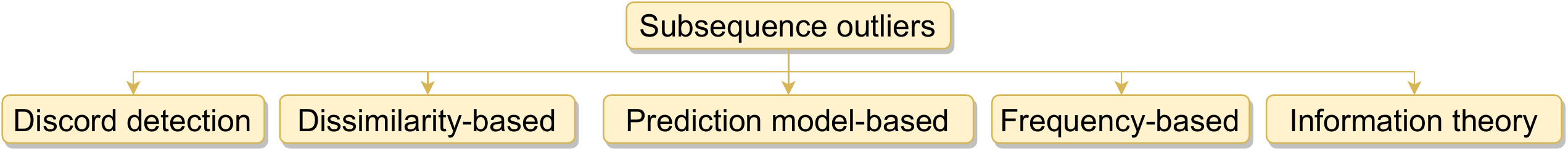}
    \caption{Types of methods for detecting subsequence outliers in univariate time series.}
    \label{fig:defsubseq}
\end{figure}

The detection methods used to detect univariate subsequence outliers in univariate time series will be analyzed in this section. We have grouped these techniques according to the different ideas or definitions on which they are based (see Fig. \ref{fig:defsubseq}).

\begin{figure}[htb!]
     \centering
     \begin{subfigure}[b]{0.38\textwidth}
         \centering
         \includegraphics[width=\textwidth]{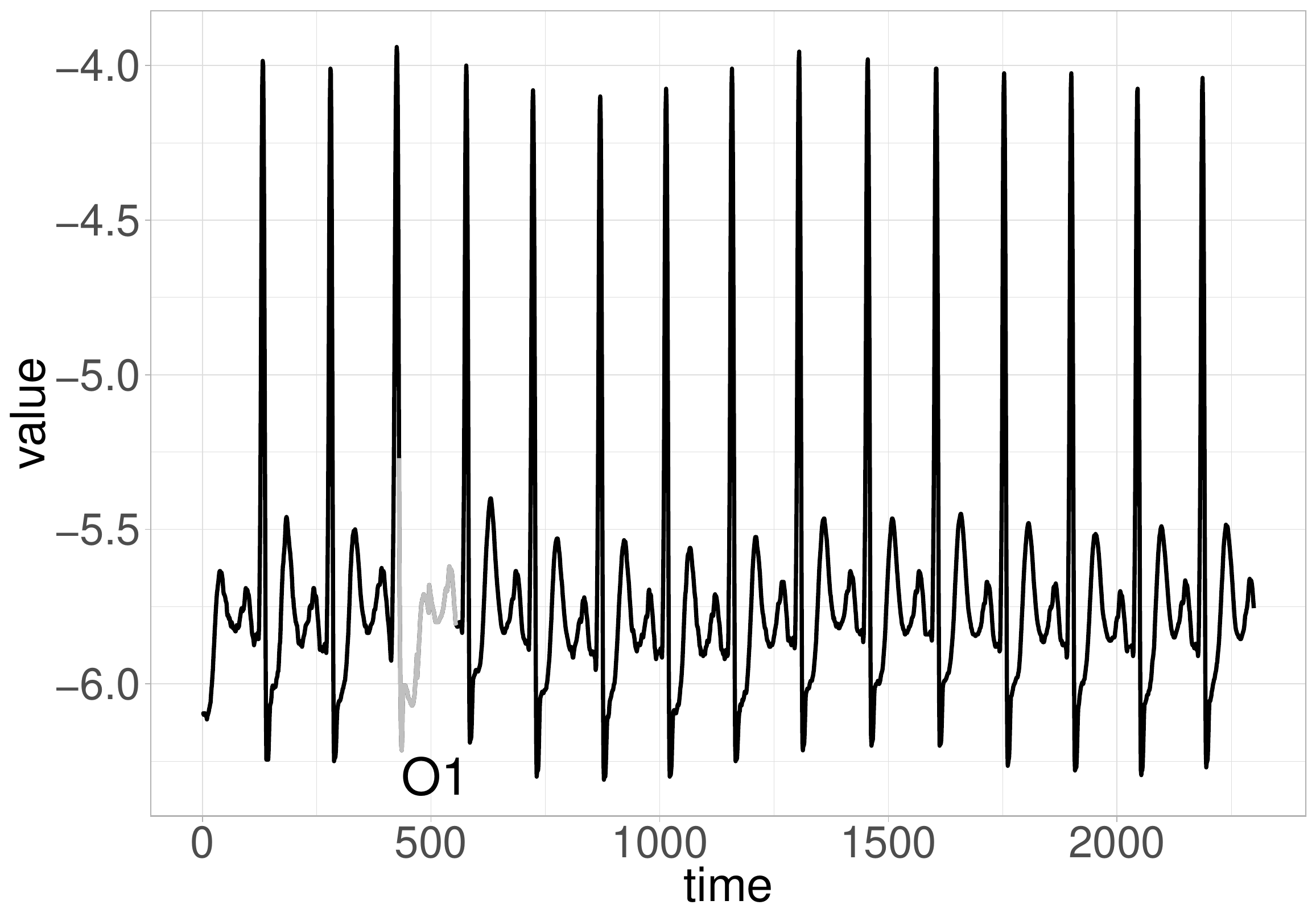}
         \caption{}
         \label{fig:discord1}
     \end{subfigure}
     \hspace*{0.6cm}
     \begin{subfigure}[b]{0.38\textwidth}
         \centering
         \includegraphics[width=\textwidth]{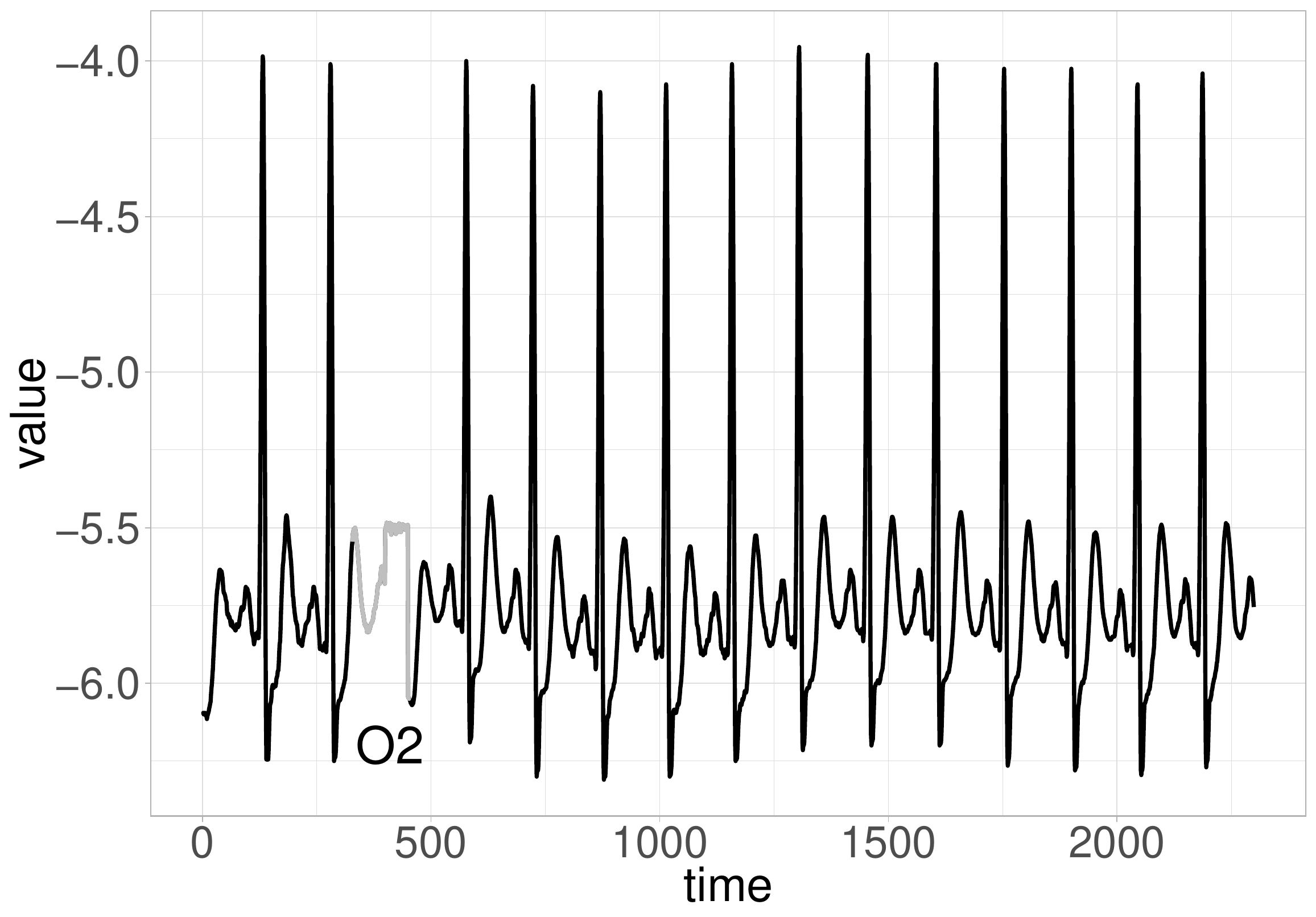}
         \caption{}
         \label{fig:discord2}
     \end{subfigure}
        \caption{Discord examples using \texttt{jmotif} package \citep{Senin2018}.}
        \label{fig:discord}
\end{figure}

The first and most straightforward idea consists of detecting the most unusual subsequence in a time series (denominated time series \textit{discord}) \citep{Keogh2005, Lin2005}, by comparing each subsequence with the others; that is, $D$ is a discord of time series $X$ if
\begin{equation}
    \forall S \in A, \hspace{0.2cm} \underset{D'\in A, D\cap D'=\emptyset}{min} (d(D, D')) > \underset{S'\in A, S\cap S'=\emptyset}{min}(d(S, S'))
    \label{eq:discord}
\end{equation}
where $A$ is the set of all subsequences of $X$ extracted by a sliding window, $D'$ is a subsequence in $A$ that does not overlap with $D$ (non-overlapping subsequences), $S'$ in $A$ does not overlap with $S$ (non-overlapping subsequences), and $d$ is the Euclidean distance between two subsequences of equal length. Typically, discord discovery techniques require the user to specify the length of the discord. 
Two examples are given in Fig. \ref{fig:discord}, in which the most unusual subsequences (O1 and O2) are shown. 

The simplest way of finding discords is using brute-force, which requires a quadratic pairwise comparison. Even so, the process can be sped-up by reordering the search using heuristics and pruning off some fruitless calculations with the HOT-SAX algorithm \citep{Keogh2005, Keogh2007, Lin2005}, which is based on the SAX discrete representation.

Many variants of the HOT-SAX algorithm aim to reduce its time complexity by enhancing both the heuristic reordering and the pruning for discord detection, which may be based on the Haar wavelet transformation and augmented trie \citep{Fu2006, Bu2007}, on bit representation clustering \citep{Li2013}, on bounding boxes \citep{Sanchez2014}, on clustering categorical data \citep{Chau2018}, and on the iSAX representation \citep{Buu2011}. Additionally, \citet{Liu2009} not only is the HOT-SAX algorithm applied in a streaming context using a sliding window but an incremental algorithm has also been suggested to simplify this computation. The method proposed by \citet{Chau2018} is also based on this incremental algorithm.  

Those discord discovery techniques require the user to prespecify the length of the discord, which in many cases may not be known. Consequently, \citet{Senin2015} present two approaches to detect variable-length discords applying grammar-induction procedures in the time series discretized with SAX. Since symbols that are rarely used in grammar rules are non-repetitive and thus most likely to be unusual, subsequence outliers correspond to rare grammar rules, which naturally vary in length. Given that the lengths of the subsequences vary, the Euclidean distance between them is calculated by shrinking the longest subsequence with the Piecewise Aggregate Approximation (PAA) \citep{Keogh2001} to obtain subsequences of the same length. 

The abovementioned discord detection techniques are limited to finding the most unusual subsequence within a time series. However, since they do not have a reference of normality or a threshold, they cannot specify whether it is indeed an outlier or not. Therefore, this decision must be made by the user. Conversely, the other categories in Fig. \ref{fig:defsubseq} consider a criterium of normality and contain specific rules to decide whether or not the identified subsequences are outliers.

\begin{figure}[htb!]
     \centering
     \begin{subfigure}[b]{0.32\textwidth}
         \centering
         \includegraphics[width=\textwidth]{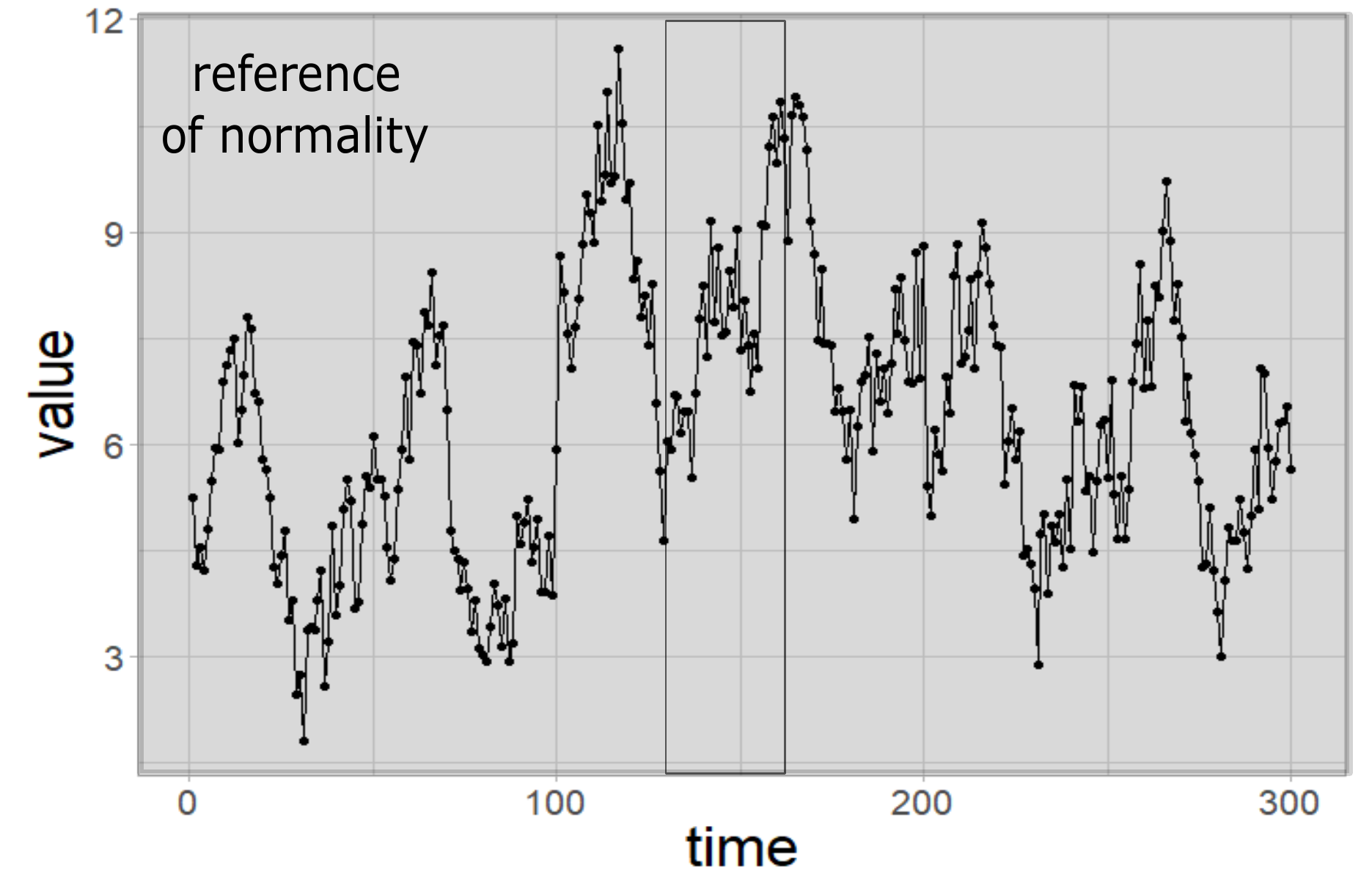}
         \caption{Same time series.}
         \label{fig:samets}
     \end{subfigure}
     \hspace*{0.09cm}
     \begin{subfigure}[b]{0.32\textwidth}
         \centering
         \includegraphics[width=\textwidth]{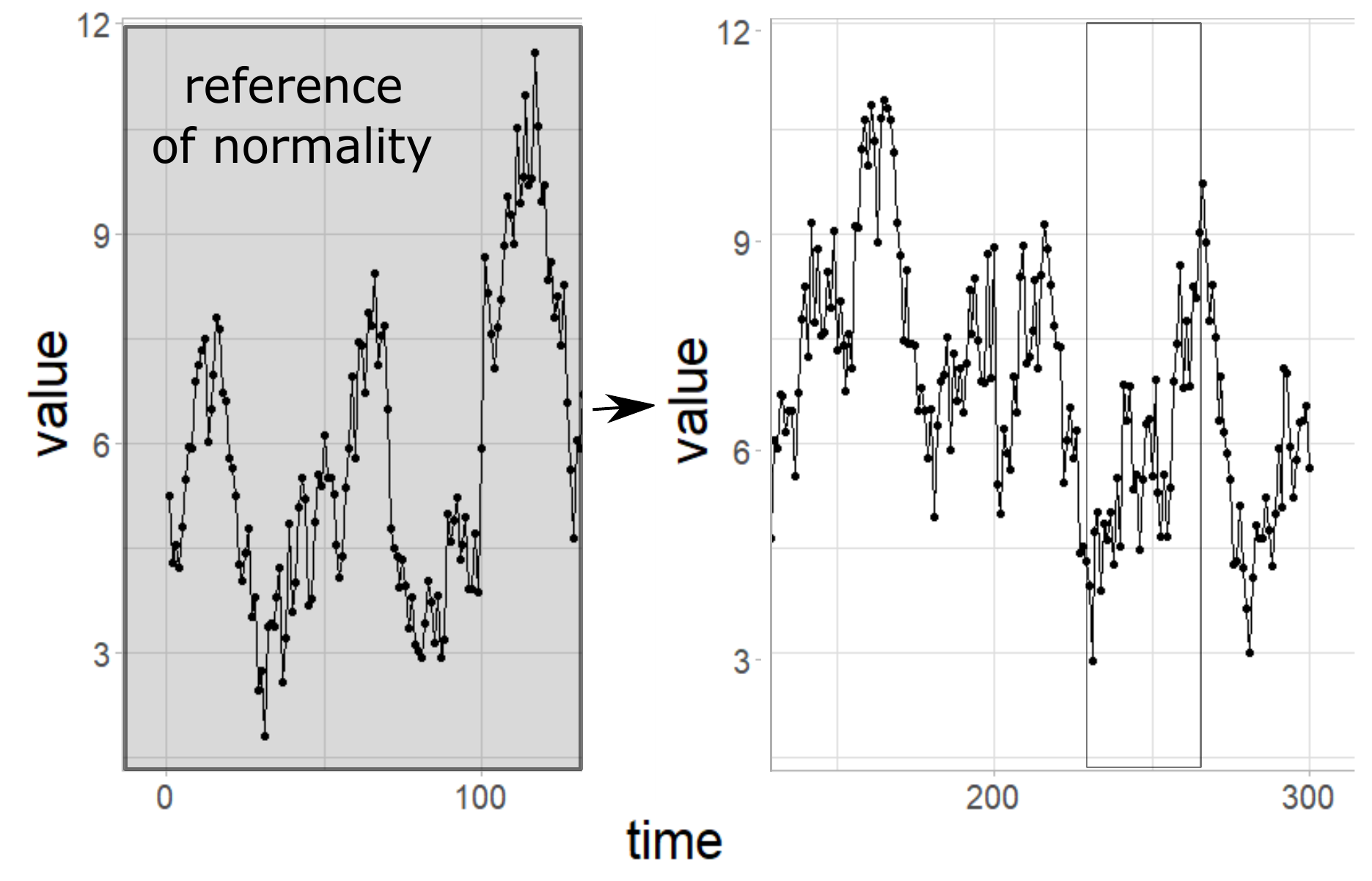}
         \caption{External time series.}
         \label{fig:fixedts}
     \end{subfigure}
     \hspace*{0.09cm}
     \begin{subfigure}[b]{0.32\textwidth}
         \centering
         \includegraphics[width=\textwidth]{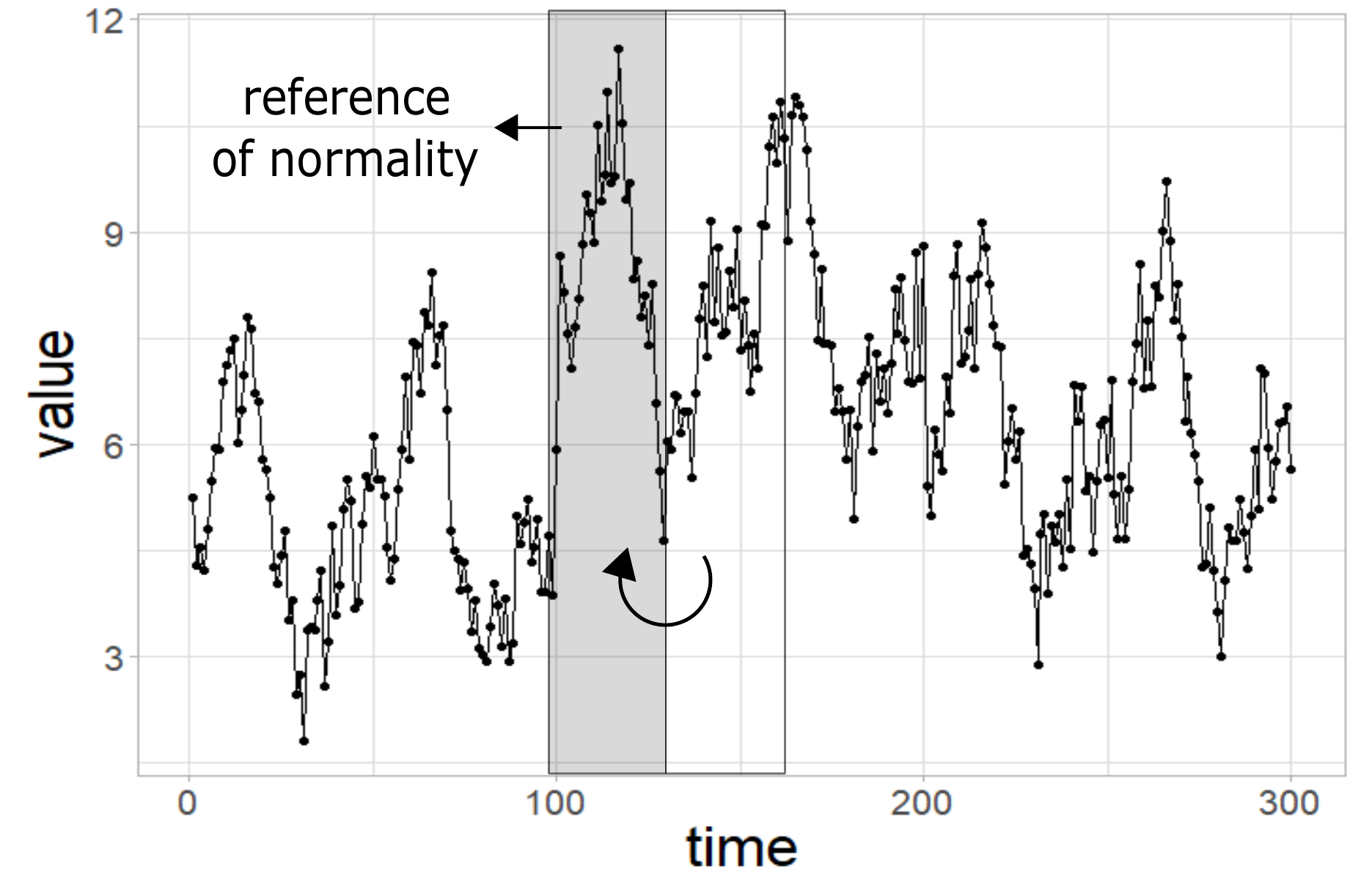}
         \caption{Previous subsequence.}
         \label{fig:previoussubs}
     \end{subfigure}
        \caption{Reference of normality used by similarity-based approaches.}
        \label{fig:normality}
\end{figure}

For example, \textit{dissimilarity-based} methods are based on the direct comparison of subsequences or their representations, using a reference of normality. In this category, the reference of normality, as well as the representations used to describe the subsequences, vary widely, in contrast to the dissimilarity-based techniques analyzed in Section \ref{sec:multitech}. Then, for a predefined threshold $\tau$, subsequence outliers are those that are dissimilar to normal subsequences; that is,
\begin{equation}
    s(S,\hat{S})>\tau
    \label{eq:outcluster}
\end{equation}
where $S$ is the subsequence being analyzed or its representation, $\hat{S}$ is the expected value of $S$ obtained based on the reference of normality, and $s$ measures the dissimilarity between two subsequences. Typically, $S$ is of fixed-length, non-periodic, and extracted via a sliding window. Some dissimilarity-based approaches are described below ordered based on the considered reference of normality (see Fig. \ref{fig:normality}), which will be used to obtain the expected value $\hat{S}$. 

\begin{figure}[htb!]
    \centering
    \includegraphics[width=4.5cm]{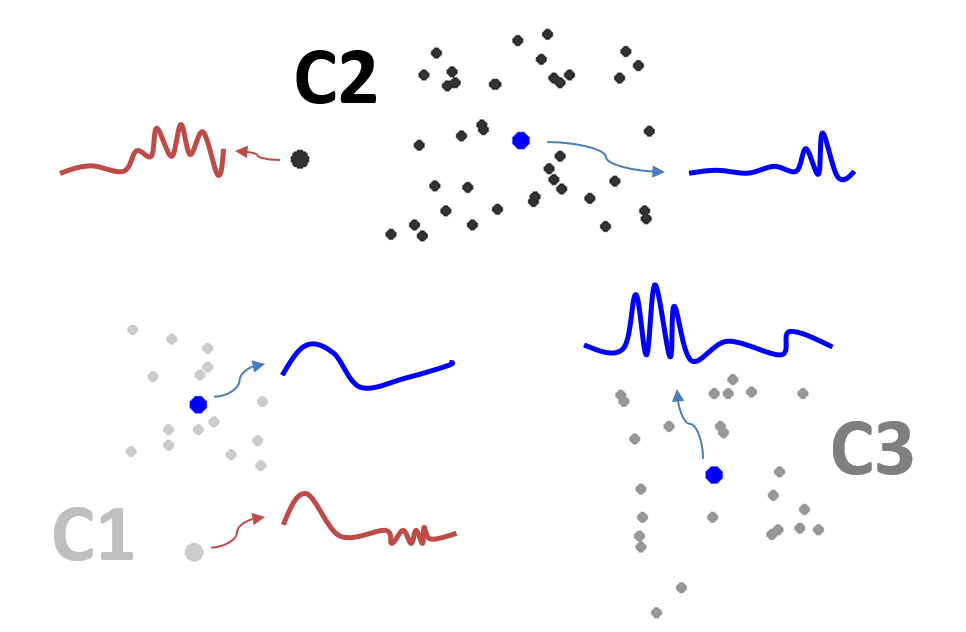}
    \caption{Clustering of the subsequences in a univariate time series. Cluster centroids are highlighted, and $C1$ and $C2$ contain subsequence outliers.}
    \label{fig:cluster}
\end{figure}

\paragraph{Same time series} The most straightforward approach is to consider the same time series object of the analysis as the reference of normality (Fig. \ref{fig:samets}). Clustering techniques commonly use this reference of normality to create clusters by grouping subsequences that are similar to each other and by separating dissimilar subsequences into different clusters (see Fig. \ref{fig:cluster}). Then, $\hat{S}$ in equation \eqref{eq:outcluster} can be defined by the centroid or center of the cluster to which $S$ belongs. For example, \citet{Chen2011} and \citet{Chen2013} cluster discretized subsequences of the same length and flag subsequences that are far from the nearest centroid ($\hat{S}$) as outliers. In this case, the distance used is the Euclidean norm of three specific distances between the discretized subsequences. Alternatively, \citet{Izakian2013} employ Fuzzy C-Means (FCM) clustering on the raw data, allowing each subsequence to belong to more than one cluster. In particular, the authors use the Euclidean distance in \eqref{eq:outcluster}. Note that these two clustering approaches are not applied to streaming time series in the original papers but could be extended using stream clustering algorithms \citep{Silva2013}.

As opposed to the clustering methods that detect fixed-length subsequence outliers, \citet{Wang2018} propose dynamic clustering to identify variable-length outliers in a streaming context. The method works directly with raw data and assumes that the observations arrive in fixed-length batches ($B_i$ is the $i^{th}$ batch). Given $L$ initial batches, grammar-induction is used to find a rule for dividing the incoming batches of data into $M$ non-overlapping variable-length subsequences; that is, $B_i=S_{i1}\cup S_{i2} \cup ...\cup S_{iM}$ (all batches are split in the same points). Clustering is then applied to the set of subsequences $S_{1j},S_{2j},...,S_{Lj}$ for each $j$ separately. A new incoming subsequence is flagged as an outlier using dynamic clustering and observing when it is either far from its closest cluster centroid or belongs to a sparsely populated cluster.

There are other techniques besides clustering that also use the same time series as the reference of normality. For instance, \citet{Ren2017} use the same time series to represent each subsequence with a value that indicates how dissimilar it is with the rest (connectivity value). To this end, a Markov chain random walk model is applied in a graph where each node is a subsequence and each edge the pairwise similarity value between the nodes \citep{Moonesinghe2006}. In particular, the pairwise similarity values are computed based on the Piecewise Aggregate Pattern Representation (PAPR), which is a matrix representation that captures the statistical information covered in the subsequence.

\paragraph{External time series} There are some methods that rely on an external time series as a reference of normality, assuming that it has been generated by the same underlying process but with no outliers. For example, \citet{Jones2016} use an external time series to find a set of exemplars that summarizes all of its subsequences within it and which detects outliers as those that are far from their nearest exemplar ($\hat{S}$). In this case, subsequences are represented by a feature vector of two components that captures both the shape and the stochastic variations of the subsequences in a time series, using smoothing and some statistics (e.g., the mean and the standard deviation). The authors use a weighted Euclidean distance in equation \eqref{eq:outcluster}, taking into account the components of both a subsequence and its nearest exemplar.

In this category, we have also included techniques that use past non-outlier subsequences as the reference of normality. Even if these subsequences belong to the same time series, this set is considered as an external set where outliers are not detected, unlike the previous category. For instance, \citet{Carrera2016} collect the most relevant subsequences within a series of past subsequences in a dictionary. Subsequence outliers are those that cannot be accurately approximated by a linear combination of some subsequences of the dictionary ($\hat{S}$), resulting in a large error in \eqref{eq:outcluster} using the Euclidean distance  \citep{Longoni2018,Carrera2019}. 

\paragraph{Previous subsequence} Finally, there are some techniques that only use the previous adjacent non-overlapping window to the subsequence being analyzed as the reference of normality, which means that they have a much more local perspective than the others. \citet{Wei2005} and \citet{Kumar2005} use this reference of normality to obtain $\hat{S}$. Specifically, the authors represent each subsequence by a bitmap; that is, a matrix in which each cell represents the frequency of a local region within a subsequence (the frequency of the subwords). A subsequence whose local regions differ from the regions of its previous adjacent subsequence ($\hat{S}$) is flagged as an outlier, using the squared Euclidean distance between each pair of elements of the bitmaps. In this case, bitmaps are incrementally updated at each time step.

Returning to the general classification of the subsequence outlier detection techniques, the third group of methods belongs to the \textit{prediction model-based} category in Fig. \ref{fig:defsubseq}, which assume that normality is reflected by a time series composed of past subsequences. These methods intend to build a prediction model that captures the dynamics of the series using past data and thus make predictions of the future. Subsequences that are far from those predictions are flagged as outliers because, although they may resemble subsequences that have appeared previously, they do not follow the dynamics captured by the model: 
\begin{equation}
    \sum_{i=p}^{p+n-1} |x_i - \hat{x}_i|>\tau
    \label{eq:modelsub}
\end{equation}
where $S=x_p,...,x_{p+n-1}$ is the subsequence being analyzed, $\hat{S}$ is its predicted value, and $\tau$ is the selected threshold. Predictions can be made in two different manners: point-wise or subsequence-wise. Models that use point-wise prediction make predictions for as many points as the length of the subsequence iteratively. With this in mind, any method within Section \ref{sec:univariatets} could be used for this purpose. However, since each predicted value is used to predict the subsequent point, the errors accumulate as predictions are made farther into the future. In contrast, the subsequence-wise prediction calculates predictions for whole subsequences at once (not iteratively). Within this category, \citet{Munir2018} use Convolutional Neural Networks (CNN) to predict whole subsequences and detect outliers using a model built with past subsequences. 

The next subsequence outlier detection methods are the \textit{frequency-based}, as shown in Fig. \ref{fig:defsubseq}, which also use one of the reference of normality mentioned in Fig. \ref{fig:normality}. A subsequence $S$ is an outlier if it does not appear as frequently as expected:
\begin{equation}
    |f(S)-\hat{f}(S)|>\tau
    \label{eq:frequ}
\end{equation}
where $f(S)$ is the frequency of occurrence of $S$, $\hat{f}(S)$ its expected frequency, and $\tau$ a predefined threshold. Due to the difficulty of finding two exact real-valued subsequences in a time series when counting the frequencies, these methods all work with a discretized time series. 

A paradigmatic example can be found in \citet{Keogh2002}. Given an external univariate time series as reference, a subsequence extracted via a sliding window in a new univariate time series is an outlier relative to that reference if the frequency of its occurrence in the new series is very different to that expected. \citet{Rasheed2014} also propose an algorithm based on the frequency of the subsequences but the aim is to detect periodic subsequence outliers together with their periodicity. In this case, the reference of normality is the same time series. The intuition behind this method is that a periodic subsequence outlier repeats less frequently in comparison to the more frequent subsequences of same length. The algorithm checks the periodicity based on the algorithm described in \citet{Rasheed2011}. 

As shown in Fig. \ref{fig:defsubseq}, the last group of subsequence outlier detection methods correspond to \textit{information theory} based techniques, which are closely related to the frequency-based methods. In particular, \citet{Yang2001, Yang2004} focus on detecting periodic subsequence outliers in discretized univariate time series using this theory. They assume that a subsequence that occurs frequently is less surprising and thus carries less information than a rare subsequence. Therefore, the aim is to find infrequent but still repetitive subsequences with rare symbols, using the same time series as the reference of normality; that is, \citet{Yang2001, Yang2004} mark $S$ as an outlier if 
\begin{equation}
    I(S)\times f(S)>\tau
    \label{eq:information}
\end{equation}
where $I(S)\geq0$ is the information carried by $S$ and $f(S)\geq1$ is the number of occurrences of $S$. $I(S)$ is computed taking into account the number of times the symbols within $S$ are repeated through the series, so a discretized subsequence $S$ that has symbols that do not commonly appear in the time series has a large $I(S)$. Conversely, if $f(S)$ is large ($S$ occurs frequently), then the information $I(S)$ will be lower, closer to $0$.

A summary of the methods that detect subsequence outliers in univariate time series is presented in Table \ref{tab:subuni}. Most of these methods detect non-periodic outliers of fixed-length, assuming the length is known in advance. Many techniques use a discretized version of the time series to compare real-valued subsequences effectively or to speed up the search process of outliers, this one specifically in the discord discovery category. However, discretizing a time series may cause loss of information. In contrast to the many methods that discover point outliers, the analyzed methods for subsequences are not iterative. In addition, for all of the authors, the outlier represents an event of interest. A reference of normality is also commonly used when detecting subsequence outliers. Finally, it is worth mentioning that some methods may perform better in time series with periodic or repetitive patterns (e.g., clustering, the frequency-based, and the information theory based).

\begin{table}[htb!]
    \caption{Summary of the characteristics of subsequence outlier detection approaches in univariate time series.}
    \label{tab:subuni}
    \centering
    \ra{1.3}
    \resizebox{.65\textwidth}{!}{
    \begin{threeparttable}
    \begin{tabular}{@{}llccccccc@{}}\toprule
    \multirow{2}{*}{Paper} &\multirow{2}{*}{Technique} & \multirow{2}{1cm}{Periodic outliers}  & & \multicolumn{2}{c}{Length} & & \multirow{2}{*}{Discretization}  & \multirow{2}{*}{Term}\\
    \cmidrule{5-6} 
    & &  \phantom{a} & & F &  V & &&  \\\midrule
        \citet{Yang2001, Yang2004} & Inf. theory &  \checkmark & & \checkmark &&& \checkmark&S \\ \hline
        \citet{Keogh2002} & Frequency & \ding{55}  & & \checkmark &&&\checkmark     & S \\ \hline
        \citet{Wei2005} & \multirow{2}{*}{Similarity} & \multirow{2}{*}{\ding{55}}  & & \multirow{2}{*}{\checkmark} &&&\multirow{2}{*}{\checkmark} 
        &  \multirow{2}{*}{A} \\  
        \citet{Kumar2005} &  & & &  &  & \\  \hline
        \citet{Keogh2005,Keogh2007} & \multirow{2}{*}{Discord} & \multirow{2}{*}{ \ding{55}} & & \multirow{2}{*}{\checkmark}&&&\multirow{2}{*}{\checkmark} 
        & \multirow{2}{*}{D} \\  
        \citet{Lin2005} &  & & & && \\  \hline
        \citet{Fu2006} & Discord &  \ding{55} & & \checkmark &  && \checkmark & D/A \\  \hline
        \citet{Bu2007} & Discord &  \ding{55} & & \checkmark &&& \checkmark & D \\  \hline
        \citet{Liu2009} & Discord &  \ding{55} & & \checkmark &&& \checkmark & D \\ \hline 
        \citet{Buu2011} & Discord &  \ding{55} & & \checkmark &&&\checkmark &  D \\  \hline
        \citet{Chen2011} & \multirow{2}{*}{Dissimilarity} & \multirow{2}{*}{ \ding{55}} & & \multirow{2}{*}{\checkmark} & & &\multirow{2}{*}{\checkmark}
        &\multirow{2}{*}{O/A}  \\ 
        \citet{Chen2013} &   & & & &  \\ \hline
        \citet{Li2013} & Discord &  \ding{55} & & \checkmark &&& \checkmark &D\\  \hline
        \citet{Izakian2013} & Dissimilarity &\ding{55} && \checkmark  &&& \ding{55} & A \\ \hline
        \citet{Sanchez2014} & Discord & \ding{55} & & \checkmark &&& \ding{55} &D/A \\  \hline
        \citet{Rasheed2014} & Frequency &  \checkmark & &\checkmark &&&\checkmark & O/S \\ \hline
        \citet{Senin2015} & Discord &  \ding{55} & & & \checkmark&&\checkmark & D/A \\  \hline
        \citet{Carrera2016,Carrera2019} & Dissimilarity &  \ding{55} & & \checkmark & && \ding{55} & O/A \\  \hline
        \citet{Jones2016} & Dissimilarity &  \ding{55} & &\checkmark & &&\ding{55}  &A \\ \hline
        \citet{Ren2017} & Dissimilarity& \ding{55} & & \checkmark & &&\ding{55}  & A \\  \hline
        \citet{Chau2018} & Discord &  \ding{55} & & \checkmark &&&\checkmark &D/O/A \\  \hline
        \citet{Munir2018} & Model& \ding{55} & & \checkmark &&&\ding{55}& D/O/A\\\hline
        \citet{Wang2018} & Dissimilarity &  \ding{55} &  & & \checkmark &&\checkmark & A \\  
    \bottomrule
    \end{tabular}
    \begin{tablenotes}
            \item F: Fixed; V: Variable // S: Surprise; D: Discord; O: Outlier; A: Anomaly.
         \end{tablenotes}
  \end{threeparttable}
    }
\end{table}

\subsection{Multivariate time series} \label{sec:multisubsequence}

This section presents some of the detection techniques that have been used in the literature to detect subsequence outliers in multivariate time series data. The nature of these methods can be either univariate (Section \ref{sec:msubsuni}) or multivariate (Section \ref{sec:msubsmulti}), and they can detect subsequence outliers that affect either one variable (univariate subsequence outliers) or multiple variables, which are all aligned (multivariate subsequence outliers).

\subsubsection{Univariate detection techniques} \label{sec:msubsuni}

Each variable of the multivariate time series can contain subsequence outliers that are independent of other variables. The identification of those subsequences may be carried out by applying the univariate techniques discussed in Section \ref{sec:unisubsq} to each time-dependent variable. In particular, \citet{Jones2014, Jones2016} apply the exemplar-based method to each variable of the multivariate input time series, using an external time series as the reference of normality and without considering the correlation that may exist between variables. Recall that omitting this correlation can lead to a loss of information. 

Intending to simplify the multivariate task but without completely ignoring the correlation between the variables, some methods reduce the dimensionality of the input series before applying a univariate detection technique. For example, \citet{Wang2018} extend their method for univariate subsequences by applying clustering to a simplified series to detect variable-length subsequence outliers (see equation \eqref{eq:outcluster}). The simplified series is obtained by first applying their univariate technique to each of the variables independently; that is, each univariate batch of data is separated into variable-length subsequences, and the obtained subsequences are then clustered as explained in Section~\ref{sec:unisubsq}. With this process, a set of representative univariate subsequences is obtained for each variable. Each new multivariate batch of data is then represented by a vector of distances, $(d_1, d_2, ..., d_l)$,  where $d_j$ represents the Euclidean distance between the $j^{th}$ variable-length subsequence of the new batch and its corresponding representative subsequence. As with their univariate technique, the reference of normality that is considered by this method is the same time series.

The technique proposed by \citet{Hu2019} is also based on reducing the dimensionality of the time series and allows us to detect variable-length discords, while using the same time series as the reference of normality. This is based on the fact that the most unusual subsequences tend to have local regions with significantly different densities (points that are similar) in comparison to the other subsequences in the series. Each point in the new univariate time series describes the density of a local region of the input multivariate time series obtained by a sliding window. This series is also used to obtain the variable-length subsequences. Discords are identified using the Euclidean and Bhattacharyya distances simultaneously.

Table \ref{tab:multi_subseq_univ} provides a summary of the univariate techniques used in multivariate time series for the detection of subsequence outliers together with their characteristics. The detected subsequence outliers affect a single variable if the univariate technique is applied to each variable. Otherwise, if the technique is employed in a reduced time series, then the outliers commonly affect multiple variables because the new series contains multivariate information. It should be noted that although \citet{Wang2018} identify multivariate batch outliers, the variable-length subsequence outliers that are then identified within those batches affect a single variable. Finally, none of the analyzed techniques detect periodic subsequence outliers.

\begin{table}[htb!]
    \caption{Summary of the univariate techniques used in multivariate time series for subsequence outlier detecion.}
    \label{tab:multi_subseq_univ}
    \centering
    \ra{1.3}
    \resizebox{.7\textwidth}{!}{
    \begin{threeparttable}
    \begin{tabular}{@{}llcccclcl@{}}\toprule
    \multirow{2}{*}{Paper} & \multirow{2}{*}{Technique}  &\phantom{a} & \multicolumn{2}{c}{Length} & \phantom{a}& \multirow{2}{*}{Subsequence}& \multirow{2}{*}{Discretization}  &   \multirow{2}{*}{Term} \\
    \cmidrule{4-5}
    & &  & Fixed & Variable &  &   \\   \midrule
    \citet{Jones2014} & Dissimilarity & & \checkmark & & &Univariate &\ding{55} &   A \\\hline
    \citet{Wang2018} & Dim.red. & &  & \checkmark &&Univariate&\checkmark & A \\\hline
    \citet{Hu2019} & Dim.red.  & &  & \checkmark & &Multivariate&\ding{55}&  A/D \\
    \bottomrule
    \end{tabular}
     \begin{tablenotes}
         \item A: Anomaly; D: Discord.
     \end{tablenotes}
  \end{threeparttable}
    }
\end{table}

\subsubsection{Multivariate detection techniques} \label{sec:msubsmulti}

\begin{figure}[htb!]
    \centering
    \includegraphics[width=6.1cm]{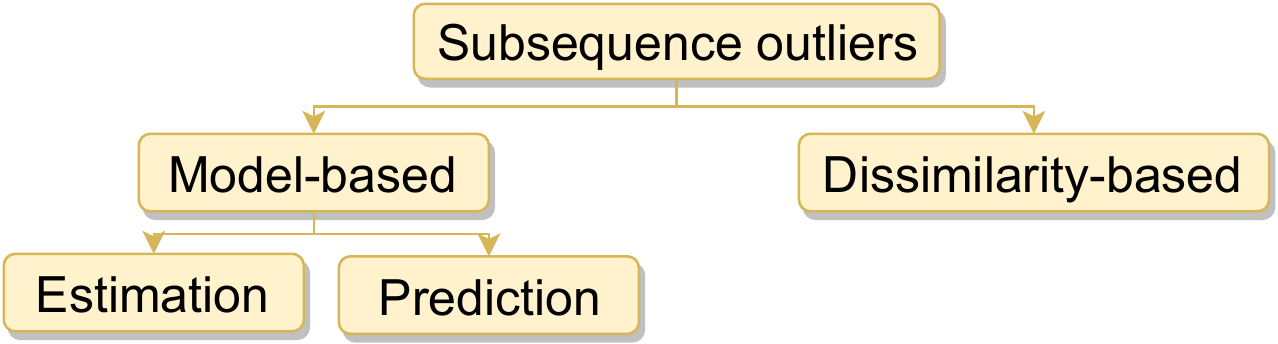}
    \caption{Types of methods for detecting subsequence outliers in multivariate time series.}
    \label{fig:esqsubMULTI}
\end{figure}

The techniques for multivariate subsequence outlier detection that will be reviewed in this section deal with multiple time-dependent variables simultaneously and typically detect temporally aligned outliers that affect multiple variables. As shown in Fig. \ref{fig:esqsubMULTI}, these techniques are divided into two main groups. However, the philosophy behind some of them is repeated because they are an extension of simpler techniques introduced in previous sections. 

The first type of technique to be discussed is the \textit{model-based} technique. As mentioned in previous sections, the aim is to detect the subsequences that are far from their expected value, which can be obtained using an \textit{estimation model} or a \textit{prediction model}. A subsequence $S=\boldsymbol{x_p},...,\boldsymbol{x_{p+n-1}}$ of length $n$ is an outlier if
\begin{equation}
    \sum_{i=p}^{p+n-1} ||\boldsymbol{x_i}-\boldsymbol{\hat{x}_i}||>\tau
\end{equation}
for a predefined threshold $\tau$. 

Within the techniques based on \textit{estimation models}, the approach proposed by \citet{Jones2014} intends to find pairs of related variables using a set of nonlinear functions. Once this is done, a value in a variable can be estimated using observations from another variable (past and future with respect to $\boldsymbol{x_t}$). These functions are learned using an external multivariate time series as the reference of normality (see Fig. \ref{fig:normality}). Estimations and error calculations can then be done for new time series. Within this new series, a subsequence in one variable is estimated using the learned functions and data from another variable. In contrast, the CNN method described by \citet{Munir2018} is a \textit{prediction model-based} technique, which is a direct extension of the method explained in Section \ref{sec:unisubsq} to detect time-aligned outliers that affect multiple variables.

The second group corresponds to the \textit{dissimilarity-based} techniques, a generalization of equation \eqref{eq:outcluster} that finds unusual subsequences in a multivariate time series based on the pairwise dissimilarity calculation between subsequences or their representations. Unlike in the univariate subsequence outlier detection, this type of technique has barely been used in multivariate series. In the only example, \citet{Cheng2008,Cheng2009} extend their method for point outliers (Section \ref{sec:multitech}) to obtain how dissimilar is a node representing a fixed-length subsequence with the others. This dissimilarity value is obtained by applying a random walk model in the graph, and the computation of the pairwise similarity between those nodes is also based on the RBF.

A summary of the multivariate techniques analyzed for subsequence outlier detection is given in Table \ref{tab:multi_subseq}. In all of the cases, the subsequence outliers represent an event of interest. In addition, the techniques find non-periodic outliers of fixed-length without using a discretization technique. 

\begin{table}[htb!]
    \caption{Summary of the multivariate techniques used in multivariate time series for subsequence outlier detecion.}
    \label{tab:multi_subseq}
    \centering
    \ra{1.3}
    \begin{threeparttable}
    \resizebox{0.55\columnwidth}{!}{
    \begin{tabular}{@{}llll@{}}\toprule
    Paper & Technique  & Subsequence &   Term \\
     \midrule
    \citet{Cheng2008,Cheng2009} & Dissimilarity & Multivariate & Anomaly \\\hline
    \citet{Jones2014} & Estimation & Univariate & Anomaly  \\\hline
    \citet{Munir2018} & Prediction & Multivariate & Anomaly / Outlier \\
    \bottomrule
    \end{tabular}
    }
  \end{threeparttable}
\end{table}
\section{Outlier time series}
\label{sec:ts}

The analyzed task has so far been to identify point and subsequence outliers within a time series, either univariate or multivariate. However, there are some cases where it may also be interesting to find entire unusual variables in a multivariate time series. This could lead, for example, to the identification of malicious users or mail servers behaving unusually. Recall that outlier time series can only be detected in multivariate time series and using a multivariate detection technique. Unfortunately, the literature has barely analyzed the detection of this type of outlier. 

Some of the key aspects presented for the subsequence outliers (see Fig. \ref{fig:esqsubs}) can also appear when attempting to detect outlier time series. First, each time-dependent variable in a multivariate time series can have a different length. Second, representation methods such as discretization can also be used to facilitate the comparisons between variables. In addition, all of the outlier time series detection methods can theoretically be applied in a streaming context using sliding windows. However, in contrast to subsequences, the property of temporality is not necessarily considered by these methods.

\begin{figure}[htb!]
    \centering
    \caption{Types of methods for detecting outlier time series in multivariate time series.}
    \includegraphics[width=6cm]{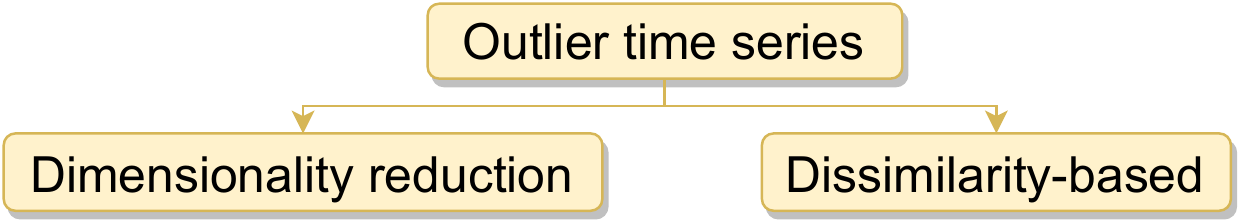}
    \label{fig:esqoutlierts}
\end{figure}

This section will examine the techniques that detect outlier time series, following the diagram given in Fig. \ref{fig:esqoutlierts}. 

\begin{figure}[htb!]
    \centering
    \includegraphics[width=11cm]{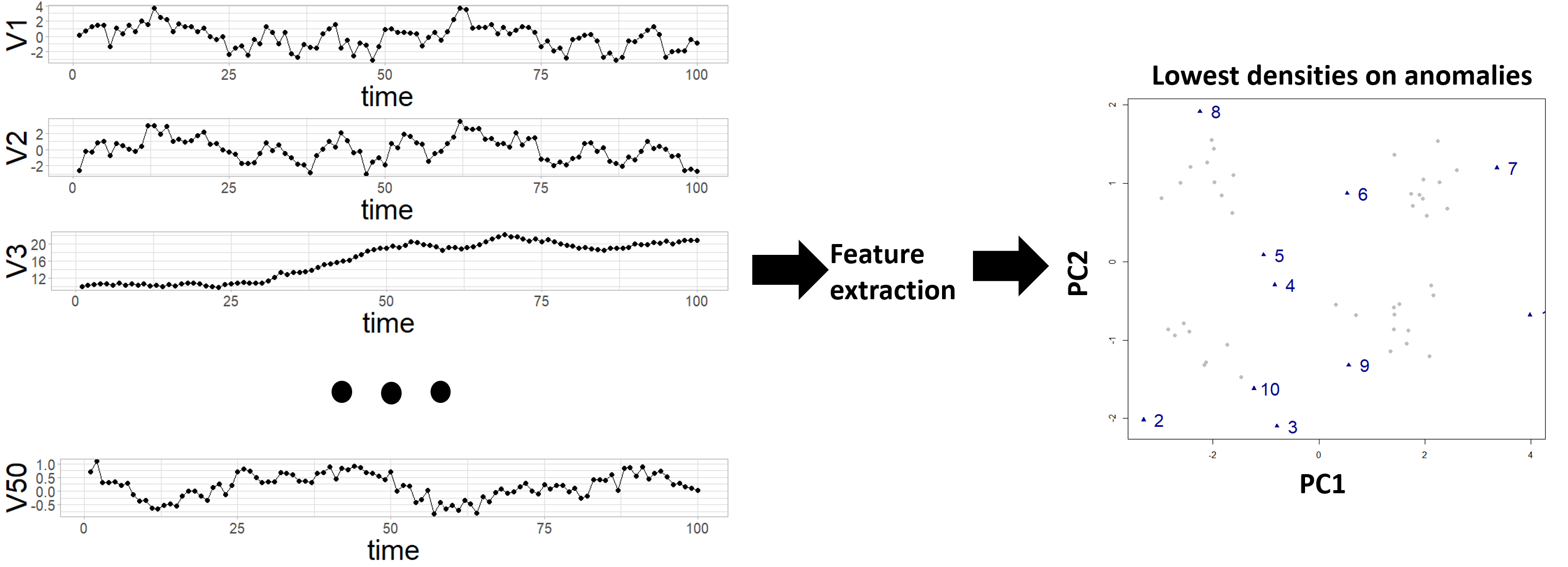}
    \caption{Outlier time series detection in a multivariate time series composed of 50 variables using PCA in the extracted features with the \texttt{anomalous} package in \texttt{R}. }
    \label{fig:outlierts}
\end{figure}

The first type of technique to be discussed is based on \textit{dimensionality reduction}. As mentioned in the previous sections, the aim of these techniques is to reduce the dimensionality of the input multivariate time series into a set of uncorrelated variables. As shown in Fig. \ref{fig:outlierts}, \citet{Hyndman2016} propose reducing the dimensionality by extracting some representative statistical features from each time series (e.g., mean, and first order of autocorrelation) and then applying PCA. Outlier time series are detected by their deviation from the highest density region in the PCA space, which is defined by the first two principal components. \citet{Laptev2015} use clustering in that PCA space to detect outlier time series by measuring the deviation between the cluster centroids and the points that represent the time series.

Instead of applying a detection technique in a lower space, original raw data can also be used directly. Indeed, the \textit{dissimilarity-based} techniques in Fig. \ref{fig:esqoutlierts} directly analyze the pairwise dissimilarity between the time series. The most common approach within this category is clustering. The intuition is similar to that depicted in Fig. \ref{fig:cluster}, but whole time series are considered instead of subsequences. In this case, the reference of normality is the same multivariate time series.

Given that the well-known $k$-means clustering algorithm is not useful for unsynchronized variables because it uses the Euclidean distance, \citet{Rebbapragada2009} propose using Phased $k$-means in unsynchronized periodic multivariate time series. This technique is a modification of $k$-means so that the phase of each time-dependent variable to its closest centroid is adjusted prior to dissimilarity calculation at every iteration. Once a set of representative time series centroids is generated, the outliers are identified using equation \eqref{eq:outcluster} and cross-correlation as the similarity measure between time series. \citet{Benkabou2018} also use clustering but in this case based on the Dynamic Time Warping (DTW) measure, which allows the series to be warped and shifted in time and have different lengths. The method optimizes an objective function by both minimizing the within-cluster distances using the DTW measure and maximizing the negative entropy of some weights that are assigned to each time series. Time series that increase the within-cluster distances to their closest cluster have smaller weights. Specifically, time series with small weights are considered to be outliers.

Other dissimilarity-based techniques are based on shapelets, which are representative time series subsequences \citep{Ye2009}. \citet{Beggel2018} use them to describe the shape of normal variables and detect outlier time-dependent variables in a multivariate time series. The shapelets are learned using an external time series as the reference of normality. The idea is to measure the dissimilarity that each learned shapelet has with a variable using the Euclidean distance. Subsequences of outlier variables are dissimilar to the learned shapelets.

A summary of the techniques used for outlier time series detection is provided in Table \ref{tab:char_outlierts}. All of these techniques intend to find events of interest and none is iterative. Unlike the subsequences, the length of the series is always specified. The approach proposed by \citet{Benkabou2018} is the only one that can deal with time series with variables of different lengths. The works reviewed in this section directly use raw data and consider temporality.

\begin{table}[htb!]
    \caption{Summary of the characteristics of outlier time series detection in multivariate time series.}
    \label{tab:char_outlierts}
    \centering
    \ra{1.3}
    \resizebox{0.73\textwidth}{!}{
    \begin{tabular}{@{}llcl@{}}\toprule
    \multicolumn{1}{c}{Paper} & Technique & Variable-length  & Term \\ \midrule
    \citet{Rebbapragada2009} & Dissimilarity-based & \ding{55} & Anomaly / Outlier \\ \hline
    \citet{Hyndman2016} & Dimensionality reduction & \ding{55} & Anomaly / Outlier / Unusual \\ \hline
    \citet{Benkabou2018} & Dissimilarity-based & \checkmark & Outlier \\\hline
    \citet{Beggel2018} & Dissimilarity-based & \ding{55} & Anomaly \\
    \bottomrule
    \end{tabular}
    }
\end{table}

\section{Publicly available software}
\label{sec:4}

In this section, the publicly available software for outlier detection in time series data according to the techniques mentioned in previous sections is provided. A summary of this software can be found in Table \ref{tab:software}, in which the technical descriptions (\textit{Related research} column) and the link to access the code (\textit{Code} column) are presented. The table is organized based on the outlier type the technique detects. In particular, the only package that includes multiple methods is \texttt{OTSAD} in \texttt{R} \citep{Iturria2020}, which detects point outliers in univariate time series.

\begin{table}[htb!]
    \caption{Summary of the publicly available software in chronological order.}
    \label{tab:software}
    \centering
    \ra{1.3}
    \begin{threeparttable}
    \resizebox{0.75\columnwidth}{!}{
    \begin{tabular}{@{}p{2.2cm}p{1.2cm}p{3.5cm}p{6.6cm}@{}} \toprule
    Name & Language & Related research & Code  \\\midrule
    
    \multicolumn{3}{l}{Point outliers} \\\midrule
    \multicolumn{1}{l}{\hspace{0.1cm}tsoutliers} & R & \citet{Chen1993} & \url{https://cran.r-project.org/web/packages/tsoutliers} \\
    \multicolumn{1}{l}{\hspace{0.1cm}spirit} & Matlab & \citet{Papadimitriou2005} & \url{http://www.cs.cmu.edu/afs/cs/project/spirit-1/www}  \\ 
    \multicolumn{1}{l}{\hspace{0.1cm}STORM} & Java & \citet{Fassetti, Angiulli2010} & \url{https://github.com/Waikato/moa/tree/master/moa/src/main/java/moa/clusterers/outliers/Angiulli} \\ 
    \multicolumn{1}{l}{\hspace{0.1cm}SCREEN} & Java & \citet{Song2015} &  \url{https://github.com/zaqthss/sigmod15-screen}  \\ 
    \multicolumn{1}{l}{\hspace{0.1cm}EGADS} & Java & \citet{Laptev2015} & \url{https://github.com/yahoo/egads} \\
    \multicolumn{1}{l}{\hspace{0.1cm}SCR} & Java & \citet{Zhang2016} &  \url{https://github.com/zaqthss/sigmod16-scr}  \\ 
    \multicolumn{1}{l}{\hspace{0.1cm}libspot} & C++ & \citet{Siffer2017} & \url{https://github.com/asiffer/libspot} \\  
    \multicolumn{1}{l}{\hspace{0.1cm}AnomalyDetection} & R & \citet{Hochenbaum2017} & \url{https://github.com/twitter/AnomalyDetection} \\
    \multicolumn{1}{l}{\hspace{0.1cm}Nupic} & Python & \citet{Ahmad2017} & \url{https://github.com/numenta/nupic} \\
    \multicolumn{1}{l}{\hspace{0.1cm}telemanon} & Python & \citet{Hundman2018} &  \url{https://github.com/khundman/telemanom}  \\ 
    \multicolumn{1}{l}{\hspace{0.1cm}OmniAnomaly} & Python & \citet{Su2019} & \url{https://github.com/smallcowbaby/OmniAnomaly} \\
    \multicolumn{1}{l}{\hspace{0.1cm}OTSAD} & R & \citet{Carter2012,Ishimtsev2017a,Iturria2020} & \url{https://cran.r-project.org/package=otsad} \\ \midrule

    \multicolumn{3}{l}{Subsequence outliers}\\ \midrule
    \multicolumn{1}{l}{\hspace{0.1cm}tsbitmaps} & Python & \citet{Wei2005, Kumar2005} & \url{https://github.com/binhmop/tsbitmaps} \\ 
    \multicolumn{1}{l}{\hspace{0.1cm}jmotif} & R & \citet{Keogh2005, Keogh2007} & \url{https://github.com/jMotif/jmotif-R} \\ 
     &  & \citet{Senin2015, Senin2018b} &  \\ 
    \multicolumn{1}{l}{\hspace{0.1cm}jmotif} & Java & \citet{Keogh2005, Keogh2007} & \url{https://github.com/jMotif/SAX} \\
     &  & \citet{Senin2015} & \\ 
    \multicolumn{1}{l}{\hspace{0.1cm}saxpy} & Python & \citet{Keogh2005, Keogh2007} & \url{https://pypi.org/project/saxpy} \phantom{aaaaaaaaaaaaaaaa} \url{https://github.com/seninp/saxpy}  \\ 
    \multicolumn{1}{l}{\hspace{0.1cm}EBAD} & C & \citet{Jones2016} &  \url{http://www.merl.com/research/license}  \\  
    \multicolumn{1}{l}{\hspace{0.1cm}GrammarViz} & Java & \citet{Senin2015, Senin2018b} &  \url{https://github.com/GrammarViz2/grammarviz2_src}  \\ \midrule

    \multicolumn{3}{l}{Outlier time series}\\ \midrule 
    \multicolumn{1}{l}{\hspace{0.1cm}anomalous} & R & \citet{Hyndman2016} &  \url{http://github.com/robjhyndman/anomalous-acm} \\ 
    \bottomrule
    \end{tabular}
    }
    \end{threeparttable}
\end{table}
\section{Concluding remarks and future work}
\label{sec:6}

In this paper, an organized overview of outlier detection techniques in time series data has been proposed. Moreover, a taxonomy that categorizes these methods depending on the input data type, the outlier type, and the nature of the detection method has been included. This section first discusses some general remarks about the analyzed techniques, and it then introduces the conclusions regarding the axes of the taxonomy. 

As seen in previous sections, a broad terminology has been used to refer to the same concept; that is, the outlier in unlabeled time series data. The term \textit{outlier} has been mainly used in point outlier detection, whereas \textit{discord} or \textit{anomaly} have been more frequently used in subsequence outlier detection. In outlier time series detection, both \textit{outlier} and \textit{anomaly} have been used. Additionally, these terms are related to the objective of the detection such that \textit{outlier} is mostly used when detecting unwanted data, whereas \textit{anomaly} has been when detecting events of interest.

In most of the analyzed works, the concept outlier represents an event of interest; that is, the authors mainly focus on extracting the outlier information considered as useful data rather than on cleaning the useless or unwanted data to improve the data quality for further analysis. Most of those who focus on handling unwanted data use iterative techniques and detect point outliers in univariate time series. These outliers are usually removed or replaced by an expected value. Therefore, it might be interesting to extend this type of methods by developing techniques that improve the data quality of multivariate time series data.

Despite the variety in terminology and purpose, all of the methods that we have reviewed are based on the same idea: detecting those parts in a time series (point, subsequence, or whole time series) that significantly differ from their expected value. Each author uses different means to obtain this expected value and compute how far it is from the actual observation to decide whether or not it is an outlier. 

Although some techniques obtain the expected value based on an external or reference set, caution must be taken because it can itself contain outliers and distort the detection. In fact, they are in the limit of the unsupervised framework because although the time series has no labels, it is usually assumed that all the external or reference data are non-outliers. 

Once the expected value is obtained, a threshold is often needed to decide whether or not we have found an outlier. Given that the results directly vary depending on that selection and few techniques give an automatic way to determine the threshold \citep{Siffer2017, Lu2018, Hundman2018, Su2019}, an interesting future line of research would be to deepen on the dynamic and adaptive selection of thresholds in both univariate and multivariate time series. Indeed, to the best of our knowledge, there are no methods that include this type of threshold in the subsequence and entire time series outlier analysis. 

As a final general remark and before proceeding with the conclusions regarding the axes of the taxonomy, the time elapsed from one observation to the subsequent is also an important aspect to consider. The vast majority of methods assume that the time series are regularly sampled. However, real life often does not provide this type of data, and converting it into such type is not always the best option. Therefore, outlier detection in an irregularly sampled time series is an interesting future direction line. 

Having provided some general conclusions, we will now focus on each of the axes. Starting from the first axis, the most remarkable conclusion is that even if most of the analysis has been performed on univariate input time series data, in recent years special emphasis has been placed on multivariate series. However, some techniques analyze each variable independently, even though they have a multivariate time series as input.

For the second axis, point outlier detection is the most researched problem due to its simplicity in comparison to the rest. Within this category, it is remarkable that some techniques do not consider the temporal information at all. This can lead to not detecting outliers that violate the temporal correlation of the series. Thus, possible future work may include temporal information to techniques that do not consider it (e.g., \citet{Cheng2008, Cheng2009, Siffer2017}) to see if the results improve. Additionally, even if theoretically many techniques can determine if a new data point is an outlier upon arrival, no study has been conducted to analyze whether these methods can really give an immediate response in real time or not. Consequently, an in-depth investigation could be done to analyze the computational cost of outlier detection techniques and to determine whether these methods can be practically used in real-time scenarios. There is also a chance for further incremental algorithm developments to ensure quick responses that adapt to the evolution of the stream.

Subsequence outliers and outlier time series have been handled less frequently. Most of the subsequence outlier detection methods find non-periodic and fixed-length outliers. Indeed, there are no techniques that identify periodic subsequence outliers in multivariate time series. This can be a promising area for further investigation and may be of interest in areas such as cybersecurity to detect periodic network attacks, or in credit-fraud detection to detect periodic thefts. Additionally, within techniques that detect subsequence outliers, care must be taken with the way in which clustering is performed because it has been shown that clustering all of the subsequences extracted from a time series by a sliding window produces meaningless results \citep{Keogh2005a}. A straightforward solution could be to use non-overlapping subsequences, but this slicing may miss relevant information when the time series presents a non-periodical structure. 

Not much work has been carried out on the detection of outlier time series. Other research directions in this line include using different distance measures and creating more effective methods for dealing with time-dependent variables of different lengths. In addition, variables could be represented using other representations than raw data to facilitate comparisons.

Within the dissimilarity-based techniques that detect either subsequence or whole time series outliers (e.g., clustering), the dissimilarity measure that is used influences the results obtained. The Euclidean distance accounts for the majority of all published work mentioned in this review due to its computation efficiency. However, other measures, such as DTW, could lead to an improvement in the detection of outliers because they include temporal information. An interesting future research direction would be to observe how different dissimilarity measures influence the outlier detection algorithms in time series data to see if any of them improve the results of the Euclidean distance. In particular, meta-learning approaches such as that proposed by \citet{Mori2016} could be used as they provide an automatic selection of the most suitable distance measure. 

In addition to these types of outliers, there could be other types of unexplored outliers. For example, it may be interesting to detect outliers that propagate over time within different variables of a multivariate time series; that is, an outlier may begin in a unique variable and then propagate through other variables in later time steps. As far as we know, this problem has not been addressed yet in the literature, or at least it has not been done under the name of outlier/anomaly detection. Therefore, developing an algorithm capable of detecting outliers caused by the same origin or variable but occurring at different time steps in different variables appears to be a promising direction for future research.

Finally, with regard to the nature of the detection method, it should be noted that some univariate techniques (e.g., the distance-based) can easily be generalized to the multivariate case by considering the distance between vectors instead of points. However, complex correlations between the variables in a multivariate time series are not taken into account. This may lead us to not identify observations that look normal in each variable individually but which violate the correlation structure between variables. In addition, when applying a univariate technique separately to each variable in a multivariate time series, the correlation dependencies are ignored. This can be computationally expensive when the number of variables is large. Hence, an extension of the univariate detection techniques applied to multivariate time series should be studied (e.g., \citet{Xu2017, Hundman2018}) so that the correlations between variables representing complex system behaviors are included.

\section{Acknowledgments}
A. Bl\'azquez-Garc\'ia and A. Conde are supported by grant KK/2019-00095 from the Elkartek program under the DIGITAL project of the Basque Government. U. Mori is supported by the Basque Government through the research group grant IT1244-19 and by the Spanish Ministry of Science, Innovation and Universities: TIN2016-78365-R.; J.A. Lozano is supported by the Basque Government through the BERC 2018-2021 program and research group grant IT1244-19 and by the Spanish Ministry of Science, Innovation and Universities: BCAM Severo Ochoa accreditation SEV-2017-0718 and TIN2016-78365-R.

\bibliographystyle{Bibliography/ACM-Reference-Format-Journals}
\bibliography{Bibliography/biblio1}

%%% -*-BibTeX-*-
%%% Do NOT edit. File created by BibTeX with style
%%% ACM-Reference-Format-Journals [18-Jan-2012].

\begin{thebibliography}{00}

%%% ====================================================================
%%% NOTE TO THE USER: you can override these defaults by providing
%%% customized versions of any of these macros before the \bibliography
%%% command.  Each of them MUST provide its own final punctuation,
%%% except for \shownote{}, \showDOI{}, and \showURL{}.  The latter two
%%% do not use final punctuation, in order to avoid confusing it with
%%% the Web address.
%%%
%%% To suppress output of a particular field, define its macro to expand
%%% to an empty string, or better, \unskip, like this:
%%%
%%% \newcommand{\showDOI}[1]{\unskip}   % LaTeX syntax
%%%
%%% \def \showDOI #1{\unskip}           % plain TeX syntax
%%%
%%% ====================================================================

\ifx \showCODEN    \undefined \def \showCODEN     #1{\unskip}     \fi
\ifx \showDOI      \undefined \def \showDOI       #1{{\tt DOI:}\penalty0{#1}\ }
  \fi
\ifx \showISBNx    \undefined \def \showISBNx     #1{\unskip}     \fi
\ifx \showISBNxiii \undefined \def \showISBNxiii  #1{\unskip}     \fi
\ifx \showISSN     \undefined \def \showISSN      #1{\unskip}     \fi
\ifx \showLCCN     \undefined \def \showLCCN      #1{\unskip}     \fi
\ifx \shownote     \undefined \def \shownote      #1{#1}          \fi
\ifx \showarticletitle \undefined \def \showarticletitle #1{#1}   \fi
\ifx \showURL      \undefined \def \showURL       #1{#1}          \fi

\bibitem[\protect\citeauthoryear{Aggarwal}{Aggarwal}{2016}]%
        {Aggarwal2016}
{C. Aggarwal}. 2016.
\newblock {\em {Outlier Analysis}\/} (2 ed.).
\newblock Springer.
\newblock


\bibitem[\protect\citeauthoryear{Aguinis, Gottfredson, and Joo}{Aguinis
  et~al\mbox{.}}{2013}]%
        {Aguinis2013}
{H. Aguinis}, {R.~K. Gottfredson}, {and} {H. Joo}. 2013.
\newblock \showarticletitle{{Best-Practice Recommendations for Defining,
  Identifying, and Handling Outliers}}.
\newblock {\em Organizational Research Methods\/} {16}, 2 (2013), 270--301.
\newblock


\bibitem[\protect\citeauthoryear{Ahmad, Lavin, Purdy, and Agha}{Ahmad
  et~al\mbox{.}}{2017}]%
        {Ahmad2017}
{S. Ahmad}, {A. Lavin}, {S. Purdy}, {and} {Z. Agha}. 2017.
\newblock \showarticletitle{{Unsupervised real-time anomaly detection for
  streaming data}}.
\newblock {\em Neurocomputing\/}  {262} (2017), 134--147.
\newblock


\bibitem[\protect\citeauthoryear{Akouemo and Povinelli}{Akouemo and
  Povinelli}{2014}]%
        {Akouemo2014}
{H.~N. Akouemo} {and} {R.~J. Povinelli}. 2014.
\newblock \showarticletitle{{Time series outlier detection and imputation}}.
\newblock {\em 2014 IEEE PES General Meeting | Conference Exposition\/} (2014),
  1--5.
\newblock


\bibitem[\protect\citeauthoryear{Akouemo and Povinelli}{Akouemo and
  Povinelli}{2016}]%
        {Akouemo2016}
{H.~N. Akouemo} {and} {R.~J. Povinelli}. 2016.
\newblock \showarticletitle{{Probabilistic anomaly detection in natural gas
  time series data}}.
\newblock {\em Int. J. Forecast.\/} {32}, 3 (2016), 948--956.
\newblock


\bibitem[\protect\citeauthoryear{Akouemo and Povinelli}{Akouemo and
  Povinelli}{2017}]%
        {Akouemo2017}
{H.~N. Akouemo} {and} {R.~J. Povinelli}. 2017.
\newblock \showarticletitle{{Data Improving in Time Series Using ARX and ANN
  Models}}.
\newblock {\em IEEE Trans. Power Syst.\/} {32}, 5 (2017), 3352--3359.
\newblock


\bibitem[\protect\citeauthoryear{Angiulli and Fassetti}{Angiulli and
  Fassetti}{2007}]%
        {Fassetti}
{F. Angiulli} {and} {F. Fassetti}. 2007.
\newblock \showarticletitle{{Detecting Distance-Based Outliers in Streams of
  Data}}. In {\em Proceedings of the 16th ACM Conference on Information and
  Knowledge Management (CIKM '07)}. ACM, Lisbon, Portugal, 811--820.
\newblock


\bibitem[\protect\citeauthoryear{Angiulli and Fassetti}{Angiulli and
  Fassetti}{2010}]%
        {Angiulli2010}
{F. Angiulli} {and} {F. Fassetti}. 2010.
\newblock \showarticletitle{{Distance-based outlier queries in data streams:
  The novel task and algorithms}}.
\newblock {\em Data Min. Knowl. Discov.\/} {20}, 2 (2010), 290--324.
\newblock


\bibitem[\protect\citeauthoryear{Baragona and Battaglia}{Baragona and
  Battaglia}{2007}]%
        {Baragona2007}
{R. Baragona} {and} {F. Battaglia}. 2007.
\newblock \showarticletitle{{Outliers detection in multivariate time series by
  independent component analysis}}.
\newblock {\em Neural Computation\/} {19}, 7 (2007), 1962--1984.
\newblock


\bibitem[\protect\citeauthoryear{Basu and Meckesheimer}{Basu and
  Meckesheimer}{2007}]%
        {Basu2007}
{S. Basu} {and} {M. Meckesheimer}. 2007.
\newblock \showarticletitle{{Automatic outlier detection for time series: An
  application to sensor data}}.
\newblock {\em Knowl. Inf. Syst.\/} {11}, 2 (2007), 137--154.
\newblock


\bibitem[\protect\citeauthoryear{Beggel, Kausler, Schiegg, Pfeiffer, and
  Bischl}{Beggel et~al\mbox{.}}{2019}]%
        {Beggel2018}
{L. Beggel}, {B.~X. Kausler}, {M. Schiegg}, {M. Pfeiffer}, {and} {B. Bischl}.
  2019.
\newblock \showarticletitle{{Time series anomaly detection based on shapelet
  learning}}.
\newblock {\em Comput. Stat.\/} {34}, 3 (2019), 945--976.
\newblock


\bibitem[\protect\citeauthoryear{Benkabou, Benabdeslem, and Canitia}{Benkabou
  et~al\mbox{.}}{2018}]%
        {Benkabou2018}
{S.~E. Benkabou}, {K. Benabdeslem}, {and} {B. Canitia}. 2018.
\newblock \showarticletitle{{Unsupervised outlier detection for time series by
  entropy and dynamic time warping}}.
\newblock {\em Knowledge and Information Systems\/} {54}, 2 (2018), 463--486.
\newblock


\bibitem[\protect\citeauthoryear{Bu, Leung, Fu, Keogh, Pei, and Meshkin}{Bu
  et~al\mbox{.}}{2007}]%
        {Bu2007}
{Y. Bu}, {T.~W. Leung}, {A.~W.~C. Fu}, {E. Keogh}, {J. Pei}, {and} {S.
  Meshkin}. 2007.
\newblock \showarticletitle{{Wat: Finding top-k discords in time series
  database}}. In {\em Proceedings of the 7th SIAM International Conference on
  Data Mining}. SIAM, Minneapolis, Minnesota, USA, 449--454.
\newblock


\bibitem[\protect\citeauthoryear{Buu and Anh}{Buu and Anh}{2011}]%
        {Buu2011}
{H.~T.~Q. Buu} {and} {D.~T. Anh}. 2011.
\newblock \showarticletitle{{Time series discord discovery based on iSAX
  symbolic representation}}. In {\em Proceedings of the 3rd International
  Conference on Knowledge and Systems Engineering}. IEEE, Hanoi, Vietnam,
  11--18.
\newblock


\bibitem[\protect\citeauthoryear{Carre{\~{n}}o, Inza, and Lozano}{Carre{\~{n}}o
  et~al\mbox{.}}{2019}]%
        {Carreno2019}
{A. Carre{\~{n}}o}, {I. Inza}, {and} {J.~A. Lozano}. 2019.
\newblock \showarticletitle{{Analyzing rare event, anomaly, novelty and outlier
  detection terms under the supervised classification framework}}.
\newblock {\em Artificial Intelligence Review\/} (2019).
\newblock


\bibitem[\protect\citeauthoryear{Carrera, Rossi, Fragneto, and
  Boracchi}{Carrera et~al\mbox{.}}{2019}]%
        {Carrera2019}
{D. Carrera}, {B. Rossi}, {P. Fragneto}, {and} {G. Boracchi}. 2019.
\newblock \showarticletitle{{Online anomaly detection for long-term ECG
  monitoring using wearable devices}}.
\newblock {\em Pattern Recognition\/}  {88} (2019), 482--492.
\newblock


\bibitem[\protect\citeauthoryear{Carrera, Rossi, Zambon, Fragneto, and
  Boracchi}{Carrera et~al\mbox{.}}{2016}]%
        {Carrera2016}
{D. Carrera}, {B. Rossi}, {D. Zambon}, {P. Fragneto}, {and} {G. Boracchi}.
  2016.
\newblock \showarticletitle{{ECG monitoring in wearable devices by sparse
  models}}. In {\em Proceedings of the European Conference on Machine Learning
  and Principles and Practice of Knowledge Discovery in Databases}. Springer,
  Riva del Garda, Italy, 145--160.
\newblock


\bibitem[\protect\citeauthoryear{Carter and Streilein}{Carter and
  Streilein}{2012}]%
        {Carter2012}
{K.~M. Carter} {and} {W.~W. Streilein}. 2012.
\newblock \showarticletitle{{Probabilistic reasoning for streaming anomaly
  detection}}. In {\em IEEE Statistical Signal Processing Workshop (SSP)},
  Vol.~1. IEEE, Ann Arbor, MI, USA, 1--4.
\newblock


\bibitem[\protect\citeauthoryear{Chandola, Banerjee, and Kumar}{Chandola
  et~al\mbox{.}}{2007}]%
        {Chandola2007}
{V. Chandola}, {A. Banerjee}, {and} {V. Kumar}. 2007.
\newblock {Outlier Detection: A Survey}.  (2007).
\newblock


\bibitem[\protect\citeauthoryear{Chandola, Banerjee, and Kumar}{Chandola
  et~al\mbox{.}}{2009}]%
        {Chandola2009}
{V. Chandola}, {A. Banerjee}, {and} {V. Kumar}. 2009.
\newblock \showarticletitle{{Anomaly detection: A survey}}.
\newblock {\em ACM Comput. Surv.\/} {41}, 3 (2009), 1--72.
\newblock


\bibitem[\protect\citeauthoryear{Chandola, Banerjee, and Kumar}{Chandola
  et~al\mbox{.}}{2012}]%
        {Chandola2012}
{V. Chandola}, {A. Banerjee}, {and} {V. Kumar}. 2012.
\newblock \showarticletitle{{Anomaly Detection for Discrete Sequences: A
  Survey}}.
\newblock {\em IEEE Trans. Knowl. Data Eng.\/} {24}, 5 (2012), 823 -- 839.
\newblock


\bibitem[\protect\citeauthoryear{Chau, Duc, and Anh}{Chau
  et~al\mbox{.}}{2018}]%
        {Chau2018}
{P.~M. Chau}, {B.~M. Duc}, {and} {D.~T. Anh}. 2018.
\newblock \showarticletitle{{Discord Discovery in Streaming Time Series based
  on an Improved HOT SAX Algorithm}}. In {\em Proceedings of the 9th
  International Symposium on Information and Communication Technology}. ACM,
  Danang City, Vietnam, 24--30.
\newblock


\bibitem[\protect\citeauthoryear{Chen and Cook}{Chen and Cook}{2011}]%
        {Chen2011}
{C. Chen} {and} {D.~J. Cook}. 2011.
\newblock \showarticletitle{{Energy Outlier Detection in Smart Environments}}.
  In {\em Workshops at the 25th AAAI Conference on Artificial Intelligence}.
  AAAI Press, San Francisco, California, USA, 9--14.
\newblock


\bibitem[\protect\citeauthoryear{Chen, Cook, and Crandall}{Chen
  et~al\mbox{.}}{2013}]%
        {Chen2013}
{C. Chen}, {D.~J. Cook}, {and} {A.~S. Crandall}. 2013.
\newblock \showarticletitle{{The user side of sustainability: Modeling behavior
  and energy usage in the home}}.
\newblock {\em Pervasive and Mobile Computing\/} {9}, 1 (2013), 161--175.
\newblock


\bibitem[\protect\citeauthoryear{Chen and Liu}{Chen and Liu}{1993}]%
        {Chen1993}
{C. Chen} {and} {L.~M. Liu}. 1993.
\newblock \showarticletitle{{Joint Estimation of Model Parameters and Outlier
  Effects in Time Series}}.
\newblock {\it J. Amer. Statist. Assoc.} {88}, 421 (1993), 284--297.
\newblock


\bibitem[\protect\citeauthoryear{Chen, Li, Lau, Cao, and Wang}{Chen
  et~al\mbox{.}}{2010}]%
        {Chen2010}
{J. Chen}, {W. Li}, {A. Lau}, {J. Cao}, {and} {K. Wang}. 2010.
\newblock \showarticletitle{{Automated load curve data cleansing in power
  systems}}.
\newblock {\em IEEE Trans. Smart Grid\/} {1}, 2 (2010), 213--221.
\newblock


\bibitem[\protect\citeauthoryear{Cheng, Tan, Potter, and Klooster}{Cheng
  et~al\mbox{.}}{2008}]%
        {Cheng2008}
{H. Cheng}, {P.~N. Tan}, {C. Potter}, {and} {S. Klooster}. 2008.
\newblock \showarticletitle{{A robust graph-based algorithm for detection and
  characterization of anomalies in noisy multivariate time series}}. In {\em
  Workshop Proceedings of the 8th IEEE International Conference on Data Mining
  (ICDM '08)}. IEEE, Pisa, Italy, 349--358.
\newblock


\bibitem[\protect\citeauthoryear{Cheng, Tan, Potter, and Klooster}{Cheng
  et~al\mbox{.}}{2009}]%
        {Cheng2009}
{H. Cheng}, {P.~N. Tan}, {C. Potter}, {and} {S. Klooster}. 2009.
\newblock \showarticletitle{{Detection and Characterization of Anomalies in
  Multivariate Time Series}}. In {\em Proceedings of the 2009 SIAM
  International Conference on Data Mining}. SIAM, Sparks, Nevada, USA,
  413--424.
\newblock


\bibitem[\protect\citeauthoryear{Dani, Jollois, Nadif, and Freixo}{Dani
  et~al\mbox{.}}{2015}]%
        {Dani2015}
{M.~C. Dani}, {F.~X. Jollois}, {M. Nadif}, {and} {C. Freixo}. 2015.
\newblock \showarticletitle{{Adaptive Threshold for Anomaly Detection Using
  Time Series Segmentation}}. In {\em Proceedings of the 22nd International
  Conference on Neural Information Processing (ICONIP '15)}. Springer, Cham,
  Istanbul, Turkey, 82--89.
\newblock


\bibitem[\protect\citeauthoryear{Esling and Agon}{Esling and Agon}{2012}]%
        {Esling2012}
{P. Esling} {and} {C. Agon}. 2012.
\newblock \showarticletitle{{Time-series data mining}}.
\newblock {\em ACM Comput. Surv.\/} {45}, 1 (2012), 1--34.
\newblock


\bibitem[\protect\citeauthoryear{Fox}{Fox}{1972}]%
        {Fox1972}
{A.~J. Fox}. 1972.
\newblock \showarticletitle{{Outliers in Time Series}}.
\newblock {\em Journal of the Royal Statistical Society: Series B
  (Methodological)\/} {34}, 3 (1972), 350--363.
\newblock


\bibitem[\protect\citeauthoryear{Fu, Leung, Keogh, and Lin}{Fu
  et~al\mbox{.}}{2006}]%
        {Fu2006}
{A.~W.~C. Fu}, {O.~T.~W. Leung}, {E. Keogh}, {and} {J. Lin}. 2006.
\newblock \showarticletitle{{Finding Time Series Discords Based on Haar
  Transform}}. In {\em Proceedings of the 2nd International Conference on
  Advanced Data Mining and Applications (ADMA '06)}. Springer, Berlin,
  Heidelberg, Xi'an, China, 31--41.
\newblock


\bibitem[\protect\citeauthoryear{Fu}{Fu}{2011}]%
        {Fu2011}
{T.~C. Fu}. 2011.
\newblock \showarticletitle{{A review on time series data mining}}.
\newblock {\em Engineering Applications of Artificial Intelligence\/} {24}, 1
  (2011), 164--181.
\newblock


\bibitem[\protect\citeauthoryear{Galeano, Pe{\~{n}}a, and Tsay}{Galeano
  et~al\mbox{.}}{2006}]%
        {Galeano2006}
{P. Galeano}, {D. Pe{\~{n}}a}, {and} {R.~S. Tsay}. 2006.
\newblock \showarticletitle{{Outlier detection in multivariate time series by
  projection pursuit}}.
\newblock {\it J. Amer. Statist. Assoc.} {101}, 474 (2006), 654--669.
\newblock


\bibitem[\protect\citeauthoryear{Gama, Zliobaite, Bifet, Pechenizkiy, and
  Bouchachia}{Gama et~al\mbox{.}}{2014}]%
        {Gama2014}
{J. Gama}, {I. Zliobaite}, {A. Bifet}, {M. Pechenizkiy}, {and} {A. Bouchachia}.
  2014.
\newblock \showarticletitle{{A Survey on Concept Drift Adaptation}}.
\newblock {\em ACM Comput. Surv.\/} {46}, 4 (2014), 1--35.
\newblock


\bibitem[\protect\citeauthoryear{Gupta, Gao, Aggarwal, and Han}{Gupta
  et~al\mbox{.}}{2014a}]%
        {Gupta2014}
{M. Gupta}, {J. Gao}, {C. Aggarwal}, {and} {J. Han}. 2014a.
\newblock {\em {Outlier Detection for Temporal Data}}.
\newblock Morgan {\&} Claypool Publishers. 1--129 pages.
\newblock


\bibitem[\protect\citeauthoryear{Gupta, Gao, Aggarwal, and Han}{Gupta
  et~al\mbox{.}}{2014b}]%
        {Gupta2013a}
{M. Gupta}, {J. Gao}, {C. Aggarwal}, {and} {J. Han}. 2014b.
\newblock \showarticletitle{{Outlier Detection for Temporal Data: A Survey}}.
\newblock {\em IEEE Trans. Knowl. Data Eng.\/} {26}, 9 (2014), 2250--2267.
\newblock


\bibitem[\protect\citeauthoryear{Hawkins}{Hawkins}{1980}]%
        {Hawkins1980}
{D.~M. Hawkins}. 1980.
\newblock {\em {Identification of outliers}}.
\newblock Springer Netherlands, New York.
\newblock


\bibitem[\protect\citeauthoryear{Hill and Minsker}{Hill and Minsker}{2010}]%
        {Hill2010}
{D.~J. Hill} {and} {B.~S. Minsker}. 2010.
\newblock \showarticletitle{{Anomaly detection in streaming environmental
  sensor data: A data-driven modeling approach}}.
\newblock {\em Environmental Modelling and Software\/} {25}, 9 (2010),
  1014--1022.
\newblock


\bibitem[\protect\citeauthoryear{Hochenbaum, Vallis, and Kejariwal}{Hochenbaum
  et~al\mbox{.}}{2017}]%
        {Hochenbaum2017}
{J. Hochenbaum}, {O.~S. Vallis}, {and} {A. Kejariwal}. 2017.
\newblock \showarticletitle{{Automatic Anomaly Detection in the Cloud Via
  Statistical Learning}}.
\newblock {\em arXiv preprint arXiv:1704.07706\/} (2017).
\newblock


\bibitem[\protect\citeauthoryear{Hodge and Austin}{Hodge and Austin}{2004}]%
        {Hodge2004}
{V. Hodge} {and} {J. Austin}. 2004.
\newblock \showarticletitle{{A Survey of Outlier Detection Methodologies}}.
\newblock {\em Artif. Intell. Rev.\/} {22}, 2 (2004), 85--126.
\newblock


\bibitem[\protect\citeauthoryear{Hu, Feng, Ji, Yan, and Zhou}{Hu
  et~al\mbox{.}}{2019}]%
        {Hu2019}
{M. Hu}, {X. Feng}, {Z. Ji}, {K. Yan}, {and} {S. Zhou}. 2019.
\newblock \showarticletitle{{A novel computational approach for discord search
  with local recurrence rates in multivariate time series}}.
\newblock {\em Information Sciences\/}  {477} (2019), 220--233.
\newblock


\bibitem[\protect\citeauthoryear{Hundman, Constantinou, Laporte, Colwell, and
  Soderstrom}{Hundman et~al\mbox{.}}{2018}]%
        {Hundman2018}
{K. Hundman}, {V. Constantinou}, {C. Laporte}, {I. Colwell}, {and} {T.
  Soderstrom}. 2018.
\newblock \showarticletitle{{Detecting Spacecraft Anomalies Using LSTMs and
  Nonparametric Dynamic Thresholding}}. In {\em Proceedings of the ACM SIGKDD
  International Conference on Knowledge Discovery and Data Mining}. ACM,
  London, UK, 387--395.
\newblock


\bibitem[\protect\citeauthoryear{Hyndman, Wang, and Laptev}{Hyndman
  et~al\mbox{.}}{2015}]%
        {Hyndman2016}
{R.~J. Hyndman}, {E. Wang}, {and} {N. Laptev}. 2015.
\newblock \showarticletitle{{Large-Scale Unusual Time Series Detection}}. In
  {\em Proceedings of the 15th IEEE International Conference on Data Mining
  Workshop (ICDMW '15)}. IEEE, Atlantic City, NJ, USA, 1616--1619.
\newblock


\bibitem[\protect\citeauthoryear{Ishimtsev, Bernstein, Burnaev, and
  Nazarov}{Ishimtsev et~al\mbox{.}}{2017}]%
        {Ishimtsev2017a}
{V. Ishimtsev}, {A. Bernstein}, {E. Burnaev}, {and} {I. Nazarov}. 2017.
\newblock \showarticletitle{{Conformal k-NN Anomaly Detector for Univariate
  Data Streams}}. In {\em Proceedings of the Sixth Workshop on Conformal and
  Probabilistic Prediction and Applications}, Vol.~60. PMLR, Stockholm, Sweden,
  213--227.
\newblock


\bibitem[\protect\citeauthoryear{Iturria, Carrasco, Charramendieta, Conde, and
  Herrera}{Iturria et~al\mbox{.}}{2020}]%
        {Iturria2020}
{A. Iturria}, {J. Carrasco}, {S. Charramendieta}, {A. Conde}, {and} {F.
  Herrera}. 2020.
\newblock \showarticletitle{{otsad: A package for online time-series anomaly
  detectors}}.
\newblock {\em Neurocomputing\/}  {374} (2020), 49--53.
\newblock


\bibitem[\protect\citeauthoryear{Izakian and Pedrycz}{Izakian and
  Pedrycz}{2013}]%
        {Izakian2013}
{H. Izakian} {and} {W. Pedrycz}. 2013.
\newblock \showarticletitle{{Anomaly detection in time series data using a
  fuzzy c-means clustering}}. In {\em Proceedings of the 2013 Joint IFSA World
  Congress and NAFIPS Annual Meeting}. IEEE, Edmonton, Alberta, Canada,
  1513--1518.
\newblock


\bibitem[\protect\citeauthoryear{Jagadish, Koudas, and Muthukrishnan}{Jagadish
  et~al\mbox{.}}{1999}]%
        {Dalgleish1999}
{H.V. Jagadish}, {N. Koudas}, {and} {S. Muthukrishnan}. 1999.
\newblock \showarticletitle{{Mining Deviants in a Time Series Database}}. In
  {\em Proceedings of the 25th International Conference on Very Large Data
  Bases}. Morgan Kaufmann Publishers Inc., Edinburgh, Scotland, UK, 102--113.
\newblock


\bibitem[\protect\citeauthoryear{Jones, Nikovski, Imamura, and Hirata}{Jones
  et~al\mbox{.}}{2014}]%
        {Jones2014}
{M. Jones}, {D. Nikovski}, {M. Imamura}, {and} {T. Hirata}. 2014.
\newblock \showarticletitle{{Anomaly Detection in Real-Valued Multidimensional
  Time Series}}. In {\em International Conference on
  Bigdata/Socialcom/Cybersecurity}. Citeseer, Stanford, CA, USA.
\newblock


\bibitem[\protect\citeauthoryear{Jones, Nikovski, Imamura, and Hirata}{Jones
  et~al\mbox{.}}{2016}]%
        {Jones2016}
{M. Jones}, {D. Nikovski}, {M. Imamura}, {and} {T. Hirata}. 2016.
\newblock \showarticletitle{{Exemplar learning for extremely efficient anomaly
  detection in real-valued time series}}.
\newblock {\em Data Min. Knowl. Discov.\/} {30}, 6 (2016), 1427--1454.
\newblock


\bibitem[\protect\citeauthoryear{Keogh, Chakrabarti, Pazzani, and
  Mehrotra}{Keogh et~al\mbox{.}}{2001}]%
        {Keogh2001}
{E. Keogh}, {K. Chakrabarti}, {M. Pazzani}, {and} {S. Mehrotra}. 2001.
\newblock \showarticletitle{{Dimensionality Reduction for Fast Similarity
  Search in Large Time Series Databases}}.
\newblock {\em Knowledge and Information Systems\/} {3}, 3 (2001), 263--286.
\newblock


\bibitem[\protect\citeauthoryear{Keogh and Lin}{Keogh and Lin}{2005}]%
        {Keogh2005a}
{E. Keogh} {and} {J. Lin}. 2005.
\newblock \showarticletitle{{Clustering of time-series subsequences is
  meaningless: Implications for previous and future research}}.
\newblock {\em Knowledge and Information Systems\/} {8}, 2 (2005), 154--177.
\newblock


\bibitem[\protect\citeauthoryear{Keogh, Lin, and Fu}{Keogh
  et~al\mbox{.}}{2005}]%
        {Keogh2005}
{E. Keogh}, {J. Lin}, {and} {A. Fu}. 2005.
\newblock \showarticletitle{{HOT SAX: Efficiently Finding the Most Unusual Time
  Series Subsequence}}. In {\em Proceedings of the 5th IEEE International
  Conference on Data Mining (ICDM '05)}. IEEE, Houston, TX, USA, 226--233.
\newblock


\bibitem[\protect\citeauthoryear{Keogh, Lin, Lee, and van Herle}{Keogh
  et~al\mbox{.}}{2007}]%
        {Keogh2007}
{E. Keogh}, {J. Lin}, {S.~H. Lee}, {and} {H. van Herle}. 2007.
\newblock \showarticletitle{{Finding the most unusual time series subsequence:
  Algorithms and applications}}.
\newblock {\em Knowledge and Information Systems\/} {11}, 1 (2007), 1--27.
\newblock


\bibitem[\protect\citeauthoryear{Keogh, Lonardi, and Chiu}{Keogh
  et~al\mbox{.}}{2002}]%
        {Keogh2002}
{E. Keogh}, {S. Lonardi}, {and} {B.~Y.~C. Chiu}. 2002.
\newblock \showarticletitle{{Finding surprising patterns in a time series
  database in linear time and space}}. In {\em Proceedings of the 8th ACM
  SIGKDD International Conference on Knowledge Discovery and Data Mining (KDD
  '02)}. ACM, Edmonton, Alberta, Canada, 550--556.
\newblock


\bibitem[\protect\citeauthoryear{Kieu, Yang, and Jensen}{Kieu
  et~al\mbox{.}}{2018}]%
        {Kieu2018}
{T. Kieu}, {B. Yang}, {and} {C.~S. Jensen}. 2018.
\newblock \showarticletitle{{Outlier detection for multidimensional time series
  using deep neural networks}}. In {\em Proceedings of the 19th IEEE
  International Conference on Mobile Data Management}. IEEE, Aalborg, Denmark,
  125--134.
\newblock


\bibitem[\protect\citeauthoryear{Kumar, Lolla, Keogh, Lonardi, and
  Ratanamahatana}{Kumar et~al\mbox{.}}{2005}]%
        {Kumar2005}
{N. Kumar}, {N. Lolla}, {E. Keogh}, {S. Lonardi}, {and} {C.~A. Ratanamahatana}.
  2005.
\newblock \showarticletitle{{Time-series Bitmaps: A Practical Visualization
  Tool for working with Large Time Series Databases}}. In {\em Proceedings of
  the 5th SIAM International Conference on Data Mining}. SIAM, Newport Beach,
  CA, USA, 531--535.
\newblock


\bibitem[\protect\citeauthoryear{Laptev, Amizadeh, and Flint}{Laptev
  et~al\mbox{.}}{2015}]%
        {Laptev2015}
{N. Laptev}, {S. Amizadeh}, {and} {I. Flint}. 2015.
\newblock \showarticletitle{{Generic and Scalable Framework for Automated
  Time-series Anomaly Detection}}. In {\em Proceedings of the 21th ACM SIGKDD
  International Conference on Knowledge Discovery and Data Mining (KDD '15)}.
  ACM, Sydney, NSW, Australia, 1939--1947.
\newblock


\bibitem[\protect\citeauthoryear{Li, Br{\"{a}}ysy, Jiang, Wu, and Wang}{Li
  et~al\mbox{.}}{2013}]%
        {Li2013}
{G. Li}, {O. Br{\"{a}}ysy}, {L. Jiang}, {Z. Wu}, {and} {Y. Wang}. 2013.
\newblock \showarticletitle{{Finding time series discord based on bit
  representation clustering}}.
\newblock {\em Knowledge-Based Systems\/}  {54} (2013), 243--254.
\newblock


\bibitem[\protect\citeauthoryear{Li, Li, Han, and Lee}{Li
  et~al\mbox{.}}{2009}]%
        {Li2009}
{X. Li}, {Z. Li}, {J. Han}, {and} {J.~G. Lee}. 2009.
\newblock \showarticletitle{{Temporal outlier detection in vehicle traffic
  data}}. In {\em 2009 IEEE 25th International Conference on Data Engineering}.
  IEEE, Shanghai, China, 1319--1322.
\newblock


\bibitem[\protect\citeauthoryear{Lin, Keogh, Fu, and {Van Herle}}{Lin
  et~al\mbox{.}}{2005}]%
        {Lin2005}
{J. Lin}, {E. Keogh}, {A. Fu}, {and} {H. {Van Herle}}. 2005.
\newblock \showarticletitle{{Approximations to Magic: Finding Unusual Medical
  Time Series}}. In {\em Proceedings of the 18th IEEE Symposium on
  Computer-Based Medical Systems}. IEEE, Dublin, Ireland, 329--334.
\newblock


\bibitem[\protect\citeauthoryear{Liu, Chen, Wang, and Yin}{Liu
  et~al\mbox{.}}{2009}]%
        {Liu2009}
{Y. Liu}, {X. Chen}, {F. Wang}, {and} {J. Yin}. 2009.
\newblock \showarticletitle{{Efficient detection of discords for time series
  stream}}. In {\em Proceedings of the Joint International Conferences on
  Advances in Data and Web Management}. Springer, Berlin, Heidelberg, Suzhou,
  China, 629--634.
\newblock


\bibitem[\protect\citeauthoryear{Longoni, Carrera, Rossi, Fragneto, Pessione,
  and Boracchi}{Longoni et~al\mbox{.}}{2018}]%
        {Longoni2018}
{M. Longoni}, {D. Carrera}, {B. Rossi}, {P. Fragneto}, {M. Pessione}, {and} {G.
  Boracchi}. 2018.
\newblock \showarticletitle{{A wearable device for online and long-term ECG
  monitoring}}. In {\em Proceedings of the 27th International Joint Conference
  on Artificial Intelligence (IJCAI'18)}. International Joint Conferences on
  Artificial Intelligence, Stockholm, Sweden, 5838--5840.
\newblock


\bibitem[\protect\citeauthoryear{Losing, Hammer, and Wersing}{Losing
  et~al\mbox{.}}{2018}]%
        {Losing2018}
{V. Losing}, {B. Hammer}, {and} {H. Wersing}. 2018.
\newblock \showarticletitle{{Incremental on-line learning: A review and
  comparison of state of the art algorithms}}.
\newblock {\em Neurocomputing\/}  {275} (2018), 1261--1274.
\newblock


\bibitem[\protect\citeauthoryear{Lu, Liu, Fei, and Guan}{Lu
  et~al\mbox{.}}{2018}]%
        {Lu2018}
{H. Lu}, {Y. Liu}, {Z. Fei}, {and} {C. Guan}. 2018.
\newblock \showarticletitle{{An outlier detection algorithm based on
  cross-correlation analysis for time series dataset}}.
\newblock {\em IEEE Access\/}  {6} (2018), 53593--53610.
\newblock


\bibitem[\protect\citeauthoryear{Mehrang, Helander, Pavel, Chieh, and
  Korhonen}{Mehrang et~al\mbox{.}}{2015}]%
        {Mehrang2015}
{S. Mehrang}, {E. Helander}, {M. Pavel}, {A. Chieh}, {and} {I. Korhonen}. 2015.
\newblock \showarticletitle{{Outlier detection in weight time series of
  connected scales}}. In {\em Proceedings of the 2015 IEEE International
  Conference on Bioinformatics and Biomedicine (BIBM '15)}. IEEE, Washington,
  DC, USA, 1489--1496.
\newblock


\bibitem[\protect\citeauthoryear{Moonesinghe and Tan}{Moonesinghe and
  Tan}{2006}]%
        {Moonesinghe2006}
{H.D.K. Moonesinghe} {and} {P.~N. Tan}. 2006.
\newblock \showarticletitle{{Outlier Detection Using Random Walks}}. In {\em
  Proceedings of the 18th IEEE International Conference on Tools with
  Artificial Intelligence (ICTAI '06)}. IEEE, Arlington, VA, USA, 532--539.
\newblock


\bibitem[\protect\citeauthoryear{Mori, Mendiburu, and Lozano}{Mori
  et~al\mbox{.}}{2016}]%
        {Mori2016}
{U. Mori}, {A. Mendiburu}, {and} {J.~A. Lozano}. 2016.
\newblock \showarticletitle{{Similarity Measure Selection for Clustering Time
  Series Databases}}.
\newblock {\em IEEE Trans. Knowl. Data Eng.\/} {28}, 1 (2016), 181--195.
\newblock


\bibitem[\protect\citeauthoryear{Munir, Siddiqui, Dengel, and Ahmed}{Munir
  et~al\mbox{.}}{2019}]%
        {Munir2018}
{M. Munir}, {S.~A. Siddiqui}, {A. Dengel}, {and} {S. Ahmed}. 2019.
\newblock \showarticletitle{{DeepAnT: A Deep Learning Approach for Unsupervised
  Anomaly Detection in Time Series}}.
\newblock {\em IEEE Access\/}  {7} (2019), 1991--2005.
\newblock


\bibitem[\protect\citeauthoryear{Muthukrishnan, Shah, and Vitter}{Muthukrishnan
  et~al\mbox{.}}{2004}]%
        {Muthukrishnan2004}
{S. Muthukrishnan}, {R. Shah}, {and} {J.~S. Vitter}. 2004.
\newblock \showarticletitle{{Mining Deviants in Time Series Data Streams}}. In
  {\em Proceedings of the 16th International Conference on Scientific and
  Statistical Database Management}. IEEE, Santorini Island, Greece, 41--50.
\newblock


\bibitem[\protect\citeauthoryear{Papadimitriou, Sun, and
  Faloutsos}{Papadimitriou et~al\mbox{.}}{2005}]%
        {Papadimitriou2005}
{S. Papadimitriou}, {J. Sun}, {and} {C. Faloutsos}. 2005.
\newblock \showarticletitle{{Streaming Pattern Discovery in Multiple
  Time-Series}}. In {\em Proceedings of the 31st International Conference on
  Very Large Data Bases (VLDB 2005)}. ACM, Trondheim, Norway, 697--708.
\newblock


\bibitem[\protect\citeauthoryear{Rasheed and Alhajj}{Rasheed and
  Alhajj}{2014}]%
        {Rasheed2014}
{F. Rasheed} {and} {R. Alhajj}. 2014.
\newblock \showarticletitle{{A framework for periodic outlier pattern detection
  in time-series sequences}}.
\newblock {\em IEEE Transactions on Cybernetics\/} {44}, 5 (2014), 569--582.
\newblock


\bibitem[\protect\citeauthoryear{Rasheed, Alshalalfa, and Alhajj}{Rasheed
  et~al\mbox{.}}{2011}]%
        {Rasheed2011}
{F. Rasheed}, {M. Alshalalfa}, {and} {R. Alhajj}. 2011.
\newblock \showarticletitle{{Efficient periodicity mining in time series
  databases using suffix trees}}.
\newblock {\em IEEE Trans. Knowl. Data Eng.\/} {23}, 1 (2011), 79--94.
\newblock


\bibitem[\protect\citeauthoryear{Ratanamahatana, Lin, Gunopulos, Keogh,
  Vlachos, and Das}{Ratanamahatana et~al\mbox{.}}{2010}]%
        {Ratanamahatana2010}
{C.~A. Ratanamahatana}, {J. Lin}, {D. Gunopulos}, {E. Keogh}, {M. Vlachos},
  {and} {G. Das}. 2010.
\newblock \showarticletitle{{Mining Time Series Data}}.
\newblock In {\em Data Mining and Knowledge Discovery Handbook} (2nd ed.),
  {Oded Maimon} {and} {Lior Rokach} (Eds.). Springer US, Boston, MA,
  1049--1077.
\newblock


\bibitem[\protect\citeauthoryear{Rebbapragada, Protopapas, Brodley, and
  Alcock}{Rebbapragada et~al\mbox{.}}{2009}]%
        {Rebbapragada2009}
{U. Rebbapragada}, {P. Protopapas}, {C.~E. Brodley}, {and} {C. Alcock}. 2009.
\newblock \showarticletitle{{Finding anomalous periodic time series: An
  application to catalogs of periodic variable stars}}.
\newblock {\em Machine Learning\/} {74}, 3 (2009), 281--313.
\newblock


\bibitem[\protect\citeauthoryear{Reddy, Ordway-West, Lee, Dugan, Whitney,
  Kahana, Ford, Muedsam, Henslee, and Rao}{Reddy et~al\mbox{.}}{2017}]%
        {Reddy2017}
{A. Reddy}, {M. Ordway-West}, {M. Lee}, {M. Dugan}, {J. Whitney}, {R. Kahana},
  {B. Ford}, {J. Muedsam}, {A. Henslee}, {and} {M. Rao}. 2017.
\newblock \showarticletitle{{Using Gaussian Mixture Models to Detect Outliers
  in Seasonal Univariate Network Traffic}}. In {\em 2017 IEEE Security and
  Privacy Workshops (SPW)}. IEEE, San Jose, CA, USA, 229--234.
\newblock


\bibitem[\protect\citeauthoryear{Ren, Liu, Li, and Pedrycz}{Ren
  et~al\mbox{.}}{2017}]%
        {Ren2017}
{H. Ren}, {M. Liu}, {Z. Li}, {and} {W. Pedrycz}. 2017.
\newblock \showarticletitle{{A Piecewise Aggregate pattern representation
  approach for anomaly detection in time series}}.
\newblock {\em Knowledge-Based Systems\/}  {135} (2017), 29--39.
\newblock


\bibitem[\protect\citeauthoryear{Sakurada and Yairi}{Sakurada and
  Yairi}{2014}]%
        {Sakurada2014}
{M. Sakurada} {and} {T. Yairi}. 2014.
\newblock \showarticletitle{{Anomaly Detection Using Autoencoders with
  Nonlinear Dimensionality Reduction}}. In {\em Proceedings of the MLSDA 2014
  2nd Workshop on Machine Learning for Sensory Data Analysis}. ACM, Gold Coast,
  Australia, 4--12.
\newblock


\bibitem[\protect\citeauthoryear{Sanchez and Bustos}{Sanchez and
  Bustos}{2014}]%
        {Sanchez2014}
{H. Sanchez} {and} {B. Bustos}. 2014.
\newblock \showarticletitle{{Anomaly Detection in Streaming Time Series Based
  on Bounding Boxes}}. In {\em Proceedings of the 7th International Conference
  on Similarity Search and Applications (SISAP '14)}. Springer, Cham, Los
  Cabos, Mexico, 201--213.
\newblock


\bibitem[\protect\citeauthoryear{Senin}{Senin}{2018}]%
        {Senin2018}
{P. Senin}. 2018.
\newblock {jmotif: Time Series Analysis Toolkit Based on Symbolic Aggregate
  Dicretization, i.e. SAX}.
\newblock   (2018).
\newblock
\showURL{%
\url{https://cran.r-project.org/package=jmotif}}


\bibitem[\protect\citeauthoryear{Senin, Lin, Wang, Oates, Gandhi, Boedihardjo,
  Chen, and Frankenstein}{Senin et~al\mbox{.}}{2015}]%
        {Senin2015}
{P. Senin}, {J. Lin}, {X. Wang}, {T. Oates}, {S. Gandhi}, {A.~P. Boedihardjo},
  {C. Chen}, {and} {S. Frankenstein}. 2015.
\newblock \showarticletitle{{Time series anomaly discovery with grammar-based
  compression}}. In {\em Proceedings of 18th International Conference on
  Extending Database Technology (EDBT 2015)}. OpenProceedings.org, Brussels,
  Belgium, 481--492.
\newblock


\bibitem[\protect\citeauthoryear{Senin, Lin, Wang, Oates, Gandhi, Boedihardjo,
  Chen, and Frankenstein}{Senin et~al\mbox{.}}{2018}]%
        {Senin2018b}
{P. Senin}, {J. Lin}, {X. Wang}, {T. Oates}, {S. Gandhi}, {A.~P. Boedihardjo},
  {C. Chen}, {and} {S. Frankenstein}. 2018.
\newblock \showarticletitle{{GrammarViz 3.0: Interactive Discovery of
  Variable-Length Time Series Patterns}}.
\newblock {\em ACM Trans. Knowl. Discov. Data\/} {12}, 1 (2018), 1--28.
\newblock


\bibitem[\protect\citeauthoryear{Shahriar, Smith, Rahman, Freeman, Hills,
  Rawnsley, Henry, and Bishop-Hurley}{Shahriar et~al\mbox{.}}{2016}]%
        {Shahriar2016}
{M.~S. Shahriar}, {D. Smith}, {A. Rahman}, {M. Freeman}, {J. Hills}, {R.
  Rawnsley}, {D. Henry}, {and} {G. Bishop-Hurley}. 2016.
\newblock \showarticletitle{{Detecting heat events in dairy cows using
  accelerometers and unsupervised learning}}.
\newblock {\em Comput. Electron. Agric.\/}  {128} (2016), 20--26.
\newblock


\bibitem[\protect\citeauthoryear{Siffer, Fouque, Termier, and
  Largou{\"{e}}t}{Siffer et~al\mbox{.}}{2017}]%
        {Siffer2017}
{A. Siffer}, {P.~A. Fouque}, {A. Termier}, {and} {C. Largou{\"{e}}t}. 2017.
\newblock \showarticletitle{{Anomaly Detection in Streams with Extreme Value
  Theory}}. In {\em Proceedings of the 23rd ACM SIGKDD International Conference
  on Knowledge Discovery and Data Mining (KDD '17)}. ACM, Halifax, NS, Canada,
  1067--1075.
\newblock


\bibitem[\protect\citeauthoryear{Silva, Faria, Barros, Hruschka, {De Carvalho},
  and Gama}{Silva et~al\mbox{.}}{2013}]%
        {Silva2013}
{J.~A. Silva}, {E.~R. Faria}, {R.~C. Barros}, {E.~R. Hruschka}, {A.~C. {De
  Carvalho}}, {and} {J. Gama}. 2013.
\newblock \showarticletitle{{Data stream clustering: A survey}}.
\newblock {\it Comput. Surveys} {46}, 1 (2013), 1--37.
\newblock


\bibitem[\protect\citeauthoryear{Song, Zhang, Wang, and Yu}{Song
  et~al\mbox{.}}{2015}]%
        {Song2015}
{S. Song}, {A. Zhang}, {J. Wang}, {and} {P.~S. Yu}. 2015.
\newblock \showarticletitle{{SCREEN: Stream Data Cleaning under Speed
  Constraints}}. In {\em Proceedings of the 2015 ACM SIGMOD International
  Conference on Management of Data}. ACM, Melbourne, Victoria, Australia,
  827--841.
\newblock


\bibitem[\protect\citeauthoryear{Su, Zhao, Niu, Liu, Sun, and Pei}{Su
  et~al\mbox{.}}{2019}]%
        {Su2019}
{Y. Su}, {Y. Zhao}, {C. Niu}, {R. Liu}, {W. Sun}, {and} {D. Pei}. 2019.
\newblock \showarticletitle{{Robust Anomaly Detection for Multivariate Time
  Series through Stochastic Recurrent Neural Network}}. In {\em Proceedings of
  the 25th ACM SIGKDD International Conference on Knowledge Discovery and Data
  Mining (KDD '19)}. 2828--2837.
\newblock


\bibitem[\protect\citeauthoryear{Tsay}{Tsay}{1988}]%
        {Tsay1988}
{R.~S. Tsay}. 1988.
\newblock \showarticletitle{{Outliers, Level Shifts, and Variance Changes in
  Time Series}}.
\newblock {\em Journal of Forecasting\/}  {7} (1988), 1--20.
\newblock


\bibitem[\protect\citeauthoryear{Tsay, Pe{\~{n}}a, and Pankratz}{Tsay
  et~al\mbox{.}}{2000}]%
        {Tsay2000}
{R.~S. Tsay}, {D. Pe{\~{n}}a}, {and} {A.~E. Pankratz}. 2000.
\newblock \showarticletitle{{Outliers in multivariate time series}}.
\newblock {\em Biometrika\/} {87}, 4 (2000), 789--804.
\newblock


\bibitem[\protect\citeauthoryear{Tsymbal}{Tsymbal}{2004}]%
        {Tsymbal2004}
{A. Tsymbal}. 2004.
\newblock {\em {The problem of concept drift: definitions and related work}}.
\newblock {T}echnical {R}eport. Department of Computer Science, Trinity
  College: Dublin.
\newblock


\bibitem[\protect\citeauthoryear{Wang, Lin, Patel, and Braun}{Wang
  et~al\mbox{.}}{2018}]%
        {Wang2018}
{X. Wang}, {J. Lin}, {N. Patel}, {and} {M. Braun}. 2018.
\newblock \showarticletitle{{Exact variable-length anomaly detection algorithm
  for univariate and multivariate time series}}.
\newblock {\em Data Min. Knowl. Discov.\/} {32}, 6 (2018), 1806--1844.
\newblock


\bibitem[\protect\citeauthoryear{Wei, Kumar, Lolla, Keogh, Lonardi, and
  Ann}{Wei et~al\mbox{.}}{2005}]%
        {Wei2005}
{L. Wei}, {N. Kumar}, {V. Lolla}, {E. Keogh}, {S. Lonardi}, {and} {C. Ann}.
  2005.
\newblock \showarticletitle{{Assumption-free anomaly detection in time
  series}}. In {\em Proceedings of the 17th International Conference on
  Scientific and Statistical Database Management (SSDBM 2005)}. Lawrence
  Berkeley Laboratory, Santa Barbara, CA, USA, 237--240.
\newblock


\bibitem[\protect\citeauthoryear{Xu, Liu, and Yao}{Xu et~al\mbox{.}}{2019}]%
        {Xu2019}
{X. Xu}, {H. Liu}, {and} {M. Yao}. 2019.
\newblock \showarticletitle{{Recent Progress of Anomaly Detection}}.
\newblock {\em Complexity\/}  {2019} (2019), 1--11.
\newblock


\bibitem[\protect\citeauthoryear{Xu, Kersting, and von Ritter}{Xu
  et~al\mbox{.}}{2016}]%
        {Xu2016}
{Z. Xu}, {K. Kersting}, {and} {L. von Ritter}. 2016.
\newblock \showarticletitle{{Adaptive Streaming Anomaly Analysis}}. In {\em
  Proceedings of NIPS 2016 Workshop on Artificial Intelligence for Data
  Science}. Barcelona, Spain.
\newblock


\bibitem[\protect\citeauthoryear{Xu, Kersting, and von Ritter}{Xu
  et~al\mbox{.}}{2017}]%
        {Xu2017}
{Z. Xu}, {K. Kersting}, {and} {L. von Ritter}. 2017.
\newblock \showarticletitle{{Stochastic Online Anomaly Analysis for Streaming
  Time Series}}. In {\em Proceedings of the 26th International Joint Conference
  on Artificial Intelligence (IJCAI '17)}. International Joint Conferences on
  Artificial Intelligence, Melbourne, Australia, 3189--3195.
\newblock


\bibitem[\protect\citeauthoryear{Yang, Wang, and Yu}{Yang
  et~al\mbox{.}}{2001}]%
        {Yang2001}
{J. Yang}, {W. Wang}, {and} {P.~S. Yu}. 2001.
\newblock \showarticletitle{{InfoMiner: Mining Surprising Periodic Patterns}}.
  In {\em Proceedings of the 7th ACM SIGKDD International Conference on
  Knowledge Discovery and Data Mining (KDD '01)}. ACM, San Francisco, CA, USA,
  395--400.
\newblock


\bibitem[\protect\citeauthoryear{Yang, Wang, and Yu}{Yang
  et~al\mbox{.}}{2004}]%
        {Yang2004}
{J. Yang}, {W. Wang}, {and} {P.~S. Yu}. 2004.
\newblock \showarticletitle{{Mining surprising periodic patterns}}.
\newblock {\em Data Mining and Knowledge Discovery\/}  {9} (2004), 189--216.
\newblock


\bibitem[\protect\citeauthoryear{Ye and Keogh}{Ye and Keogh}{2009}]%
        {Ye2009}
{L. Ye} {and} {E. Keogh}. 2009.
\newblock \showarticletitle{{Time series shapelets: a new primitive for data
  mining}}. In {\em Proceedings of the 15th ACM SIGKDD International Conference
  on Knowledge Discovery and Data Mining (KDD '09)}. ACM, Paris, France,
  947--956.
\newblock


\bibitem[\protect\citeauthoryear{Zhang, Song, and Wang}{Zhang
  et~al\mbox{.}}{2016}]%
        {Zhang2016}
{A. Zhang}, {S. Song}, {and} {J. Wang}. 2016.
\newblock \showarticletitle{{Sequential Data Cleaning: A Statistical
  Approach}}. In {\em Proceedings of the 2016 International Conference on
  Management of Data (SIGMOD '16)}. ACM, San Francisco, CA, USA, 909--924.
\newblock


\bibitem[\protect\citeauthoryear{Zhang, Hamm, Meratnia, Stein, van~de Voort,
  and Havinga}{Zhang et~al\mbox{.}}{2012}]%
        {Zhang2012}
{Y. Zhang}, {N.~A.~S. Hamm}, {N. Meratnia}, {A. Stein}, {M. van~de Voort},
  {and} {P.~J.~M. Havinga}. 2012.
\newblock \showarticletitle{{Statistics-based outlier detection for wireless
  sensor networks}}.
\newblock {\em Int. J. Geogr. Inf. Sci.\/} {26}, 8 (2012), 1373--1392.
\newblock


\bibitem[\protect\citeauthoryear{Zhou, Arghandeh, and Spanos}{Zhou
  et~al\mbox{.}}{2016}]%
        {Zhou2016}
{Y. Zhou}, {R. Arghandeh}, {and} {C.~J. Spanos}. 2016.
\newblock \showarticletitle{{Online learning of Contextual Hidden Markov Models
  for temporal-spatial data analysis}}. In {\em 2016 IEEE 55th Conference on
  Decision and Control (CDC)}. IEEE, Las Vegas, NV, USA, 6335--6341.
\newblock


\bibitem[\protect\citeauthoryear{Zhou, Arghandeh, Zou, and Spanos}{Zhou
  et~al\mbox{.}}{2019}]%
        {Zhou2019}
{Y. Zhou}, {R. Arghandeh}, {H. Zou}, {and} {C.~J. Spanos}. 2019.
\newblock \showarticletitle{{Nonparametric Event Detection in Multiple Time
  Series for Power Distribution Networks}}.
\newblock {\em IEEE Transactions on Industrial Electronics\/} {66}, 2 (2019),
  1619--1628.
\newblock


\bibitem[\protect\citeauthoryear{Zhou, Qin, Xu, Sadiq, and Yu}{Zhou
  et~al\mbox{.}}{2018a}]%
        {Zhou2018a}
{Y. Zhou}, {R. Qin}, {H. Xu}, {S. Sadiq}, {and} {Y. Yu}. 2018a.
\newblock \showarticletitle{{A Data Quality Control Method for Seafloor
  Observatories: The Application of Observed Time Series Data in the East China
  Sea}}.
\newblock {\em Sensors\/} {18}, 8 (2018), 2628.
\newblock


\bibitem[\protect\citeauthoryear{Zhou, Zou, Arghandeh, Gu, and Spanos}{Zhou
  et~al\mbox{.}}{2018b}]%
        {Zhou2018}
{Y. Zhou}, {H. Zou}, {R. Arghandeh}, {W. Gu}, {and} {C.~J. Spanos}. 2018b.
\newblock \showarticletitle{{Non-parametric Outliers Detection in Multiple Time
  Series A Case Study: Power Grid Data Analysis}}. In {\em 32nd AAAI Conference
  on Artificial Intelligence}. AAAI Press, New Orleans, Louisiana, USA,
  4605--4612.
\newblock


\end{thebibliography}

\end{document}